\documentclass[11pt,twoside]{article}
\usepackage[dvipsnames]{xcolor}
\usepackage{graphicx,graphics}
\usepackage{pdfpages}
\usepackage{hyperref}
\usepackage{url,doi} 
\usepackage[margin=1.0in]{geometry}
\usepackage{fancyhdr}
\usepackage[numbers]{natbib}
\usepackage{overpic}
\usepackage{textcomp}
\usepackage{tabularx}
\newcolumntype{Y}{>{\raggedright\arraybackslash}X}
\usepackage{xcolor}
\usepackage{etoolbox} %
\usepackage{wrapfig}
\usepackage{array}
\usepackage{colortbl}
\usepackage{subfig}
\usepackage{xspace}
\usepackage{amsthm}
\usepackage{placeins}
\usepackage{enumitem} %
\usepackage{lipsum} %
\usepackage{helvet} %
\usepackage{courier}
\usepackage{parskip} %
\usepackage{amssymb,amsfonts,amsmath} %
\usepackage{bm} %
\usepackage{braket}  %
\usepackage{supertabular}
\usepackage[capitalize,nameinlink]{cleveref} %
\crefname{figure}{Figure}{Figures}
\usepackage{tikz}
\usetikzlibrary{calc,positioning}

\usepackage{xparse}

\usepackage{bookmark}
\usepackage{hyperref}
\hypersetup{
  naturalnames=false,
  hypertexnames=false,
  breaklinks,
  colorlinks = true,
  allcolors = blue!50!black,
}

\hypersetup{ 	
  pdfsubject = {},
  pdftitle = {Randomized Algorithms for Scientific Computing (RASC)},
  pdfauthor = {}
}

\usepackage{lmodern}
\usepackage{listings}
\usepackage[most]{tcolorbox}
\DeclareTCBListing{macrobox}{O{} G{} }{#1,title={#2},
  listing options={style=tcblatex,commentstyle=\color{red!70!black}}
}
\DeclareGraphicsExtensions{.pdf, .png, .gif}

\definecolor{newblue}{rgb}{.38,.41,.76}
\definecolor{newblue2}{rgb}{.805,.852,.9375}
\definecolor{dkblue}{rgb}{.08,.11,.56}
\definecolor{grey}{rgb}{.9,.9,.9}

\usepackage[font={small,it,color=dkblue}]{caption}

\newcommand{\headrulecolor}[1]{\patchcmd{\headrule}{\hrule}{\color{#1}\hrule}{}{}}
\newcommand{\footrulecolor}[1]{\patchcmd{\footrule}{\hrule}{\color{#1}\hrule}{}{}}
\headrulecolor{newblue}
\footrulecolor{newblue}
\NewDocumentCommand{\sublead}{m}{%
  \begin{flushright}\footnotesize
    Subsection lead: #1
  \end{flushright}%
}

\NewDocumentCommand{\subleads}{m}{%
  \begin{flushright}\footnotesize
    Subsection leads: #1
  \end{flushright}%
}

\newtheoremstyle{mystyle}%
        {3ex}%
        {0ex}%
        {}%
        {0em}%
        {\bfseries}%
        {:}%
        { }%
        {}%
\theoremstyle{mystyle}

\usepackage{tcolorbox}
\usepackage{varwidth}
\tcbuselibrary{skins}
\tcbuselibrary{most}

\newtcbtheorem[]{rec}{Recommendation}%
{enhanced,frame empty,interior empty,colframe=ForestGreen!50!white,
coltitle=ForestGreen!50!black,fonttitle=\bfseries,colbacktitle=ForestGreen!15!white,
borderline={0.5mm}{0mm}{ForestGreen!15!white},
borderline={0.5mm}{0mm}{ForestGreen!50!white,dashed},
attach boxed title to top left={yshift=-2mm},
boxed title style={boxrule=0.4pt},varwidth boxed title}{rec}
\crefname{tcb@cnt@rec}{Recommendation}{Recommendations}

\newtcbtheorem[]{theme}{Theme}%
{enhanced,frame empty,interior empty,colframe=YellowOrange!50!white,
coltitle=YellowOrange!50!black,fonttitle=\bfseries,colbacktitle=YellowOrange!15!white,
borderline={0.5mm}{0mm}{YellowOrange!15!white},
borderline={0.5mm}{0mm}{YellowOrange!50!white,dashed},
attach boxed title to top center={yshift=-2mm},
boxed title style={boxrule=0.4pt},varwidth boxed title}{thm}
\crefname{tcb@cnt@theme}{Theme}{Themes}

\newtcolorbox{callout}[1][]{enhanced,
  before skip=3mm,after skip=4mm,
  boxrule=0.4pt,left=5mm,right=2mm,top=1mm,bottom=1mm,
  colback=orange!30,
  colframe=orange!20!black,
  sharp corners,
  underlay={%
    \path[fill=yellow!20!black,draw=none] (interior.south west) rectangle node[white]{\Large\bfseries~} ([xshift=3mm]interior.north west);
  },
  drop fuzzy shadow,#1}

\usepackage{booktabs, longtable}
\newcolumntype{x}[1]{>{\raggedright}p{#1}}

\AtBeginEnvironment{longtable}{%
    \setlist[itemize]{nosep,     %
                      topsep     = 0pt       ,
                      partopsep  = 0pt       ,
                      leftmargin = *         ,
                      label      = $\bullet$ ,
                      before     = \vspace{-\baselineskip},
                      after      = \vspace{-0.5\baselineskip}
                        }
                           }%

\begin{document}

\pagestyle{empty}
\phantomsection
\pdfbookmark[1]{Cover}{cover}
\includepdf{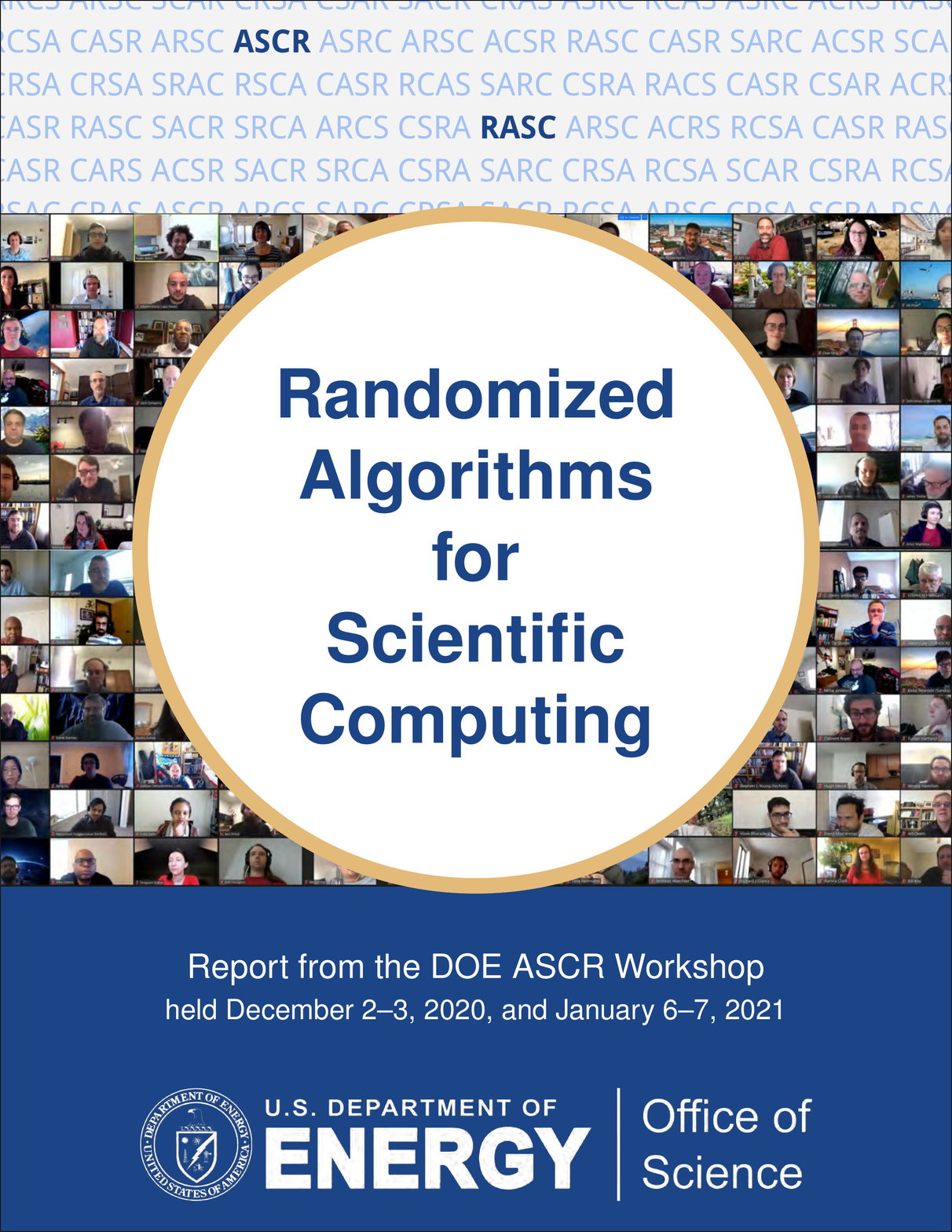}

\clearpage
\phantomsection
\pdfbookmark[1]{Inside Cover}{insidecover}
% ---- Inserted File ----
%

%

\noindent{\textbf{Cover}}\\
Cover art adapted from group photo of attendees of the virtual workshop.

\vspace*{1.5in}
\vfill

Aydin Buluc, Tamara G.\ Kolda, Stefan M.\ Wild, Mihai Anitescu, Anthony DeGennaro, John Jakeman, Chandrika Kamath, Ramakrishnan Kannan, Miles E.\ Lopes, Per-Gunnar Martinsson, Kary Myers, Jelani Nelson, Juan M. Restrepo, C.\ Seshadhri, Draguna Vrabie, Brendt Wohlberg, Stephen J. Wright, Chao Yang, Peter Zwart.
``Randomized Algorithms for Scientific Computing (RASC).'' United States Department of Energy, Advanced Scientific Computing Research, 2021. \href{https://doi.org/10.2172/1807223}{doi:10.2172/1807223}. 

\vspace{0.5in} 

\noindent{\textbf{Availability of This Report}}\\
This report is available, at no cost, at \url{https://www.osti.gov/servlets/purl/1807223}.

\vspace{0.25in} 
\noindent{\textbf{Disclaimer}}\\ 
This document was prepared as an account of a workshop sponsored by the 
U.S.\ Department of Energy. Neither the United States Government nor any 
agency thereof, nor any of their employees or officers, makes any warranty, 
express or implied, or assumes any legal liability or responsibility for the 
accuracy, completeness, or usefulness of any information, apparatus, product, or 
process disclosed, or represents that its use would not infringe privately owned 
rights. Reference herein to any specific commercial product, process, or service 
by trade name, trademark, manufacturer, or otherwise, does not necessarily 
constitute or imply its endorsement, recommendation, or favoring by the United 
States Government or any agency thereof. The views and opinions of document 
authors expressed herein do not necessarily state or reflect those of the United 
States Government or any agency thereof and shall not be used for
advertising or product endorsement purposes.

% --- End Inserted File ---

\pagenumbering{roman}
\setcounter{page}{0}

\clearpage
\pagestyle{fancy} %
\phantomsection
\pdfbookmark[1]{Title Page}{titlepage}
% ---- Inserted File ----
\begin{center}
~
\bigskip

{\bf
  {\Large Randomized Algorithms for Scientific Computing (RASC)}%
}\\

\bigskip

Report from the DOE ASCR Workshop\\
held December 2--3, 2020, and January 6--7, 2021 \\

\bigskip
\bigskip

Prepared for \\
{\em U.S.~Department of Energy\\
Office of Science\\
Advanced Scientific Computing Research Program}

\bigskip
\bigskip

{\bf RASC Report Authors}\\
Ayd\i n Bulu\c{c} (co-chair), Lawrence Berkeley National Laboratory\\
Tamara G.\ Kolda (co-chair), Sandia National Laboratories \\
Stefan M.\ Wild (co-chair), Argonne National Laboratory \\[5mm]
Mihai Anitescu, Argonne National Laboratory \\
Anthony DeGennaro, Brookhaven National Laboratory \\
John Jakeman, Sandia National Laboratories \\
Chandrika Kamath, Lawrence Livermore National Laboratory \\
Ramakrishnan (Ramki) Kannan, Oak Ridge National Laboratory \\
Miles E.\@ Lopes, University of California, Davis \\
Per-Gunnar Martinsson, University of Texas, Austin \\
Kary Myers, Los Alamos National Laboratory \\
Jelani Nelson, University of California, Berkeley \\
Juan M. Restrepo, Oak Ridge National Laboratory \\
C.~Seshadhri, University of California, Santa Cruz \\
Draguna Vrabie, Pacific Northwest National Laboratory \\
Brendt Wohlberg, Los Alamos National Laboratory \\
Stephen J.\@ Wright, University of Wisconsin, Madison \\
Chao Yang, Lawrence Berkeley National Laboratory \\
Peter Zwart, Lawrence Berkeley National Laboratory

\bigskip

{\bf DOE/ASCR Point of Contact}\\
Steven Lee \\

\bigskip
\bigskip
\bigskip

\end{center}

\clearpage

\begin{center}
  {\bf Additional Report Authors}
\end{center}

This report was a community effort, collecting inputs in a variety formats.
We have taken many of the ideas in this report from the thesis statements
collected from registrants of ``Part 2'' of the workshop.
We further solicited input from community members with specific expertise.
Much of the input came from discussions at the workshop, so we must credit
the entire set of participants listed in full in Appendix~\ref{app:participants}.
Although we cannot do justice to the origination of every idea contained herein,
we  highlight those whose contributions were
explicitly solicited or whose submissions were used nearly verbatim.

{\bf Solicited and thesis statement inputs from}\\
Rick Archibald, Oak Ridge National Laboratory\\
David Barajas-Solano, Pacific Northwest National Laboratory\\
Andrew Barker, Lawrence Livermore National Laboratory\\
Charles Bouman, Purdue University \\
Moses Charikar, Stanford University \\
Jong Choi, Oak Ridge National Laboratory\\
Aurora Clark, Washington State University \\
Victor DeCaria, Oak Ridge National Laboratory\\
Zichao Wendy Di, Argonne National Laboratory\\
Jack Dongarra, University of Tennessee, Knoxville\\
Jed Duersch, Sandia National Laboratories\\
Ethan Epperly, California Institute of Technology \\
Benjamin Erichson, University of California, Berkeley \\
Maryam Fazel, University of Washington, Seattle \\
Andrew Glaws, National Renewable Energy Laboratory\\
Carlo Graziani, Argonne National Laboratory\\
Cory Hauck, Oak Ridge National Laboratory \\
Paul Hovland, Argonne National Laboratory\\
William Kay, Oak Ridge National Laboratory\\ 
Nathan Lemons, Los Alamos National Laboratory \\
Ying Wai Li, Los Alamos National Laboratory\\
Dmitriy Morozov, Lawrence Berkeley National Laboratory\\ 
Cynthia Phillips, Sandia National Laboratories \\
Eric Phipps, Sandia National Laboratories \\
Benjamin Priest, Lawrence Livermore National Laboratory\\
Ruslan Shaydulin, Argonne National Laboratory\\
Katarzyna Swirydowicz, Pacific Northwest National Laboratory\\
Ray Tuminaro, Sandia National Laboratories
%
%
%
%
% --- End Inserted File ---

\clearpage
\phantomsection
\pdfbookmark[1]{Contents}{contents}
\tableofcontents

\clearpage
\phantomsection
\addcontentsline{toc}{section}{Executive Summary}
% ---- Inserted File ----
\section*{Executive Summary}

Randomized algorithms have propelled advances in artificial
intelligence (AI) and represent a foundational research
area in advancing AI for Science.
Future advancements in DOE Office of Science priority areas such as
climate science, astrophysics, fusion,  advanced
materials, combustion, and quantum computing all require
randomized algorithms for
surmounting challenges of complexity, robustness,
and scalability.

Advances in data collection and numerical simulation have changed
the dynamics of scientific research and motivate the need for randomized algorithms.
For instance, advances in imaging technologies such as X-ray ptychography, electron microscopy, electron energy loss spectroscopy, or adaptive optics lattice light-sheet microscopy collect hyperspectral imaging and  scattering data in terabytes, at breakneck speed enabled by  state-of-the-art detectors. The data collection is exceptionally fast  compared with its analysis.
Likewise, advances in high-performance architectures have made exascale computing a
reality and changed the economies of scientific computing in the process.
Floating-point operations that create data are essentially free in comparison with data movement.
Thus far, most approaches have focused on creating faster hardware.
Ironically, this faster hardware has exacerbated the problem by making
data still easier to create.
Under such an onslaught, scientists often resort
to heuristic deterministic sampling schemes (e.g., low-precision arithmetic, sampling every $n$th element) and 
sacrifice potentially valuable accuracy.

\begin{figure}[h]
  \centering
  \begin{tikzpicture}[node font=\footnotesize]
    \node[anchor=east,label=data] (matrix) at (0,0) {\includegraphics[trim=0 0 250 0,clip,height=40mm]{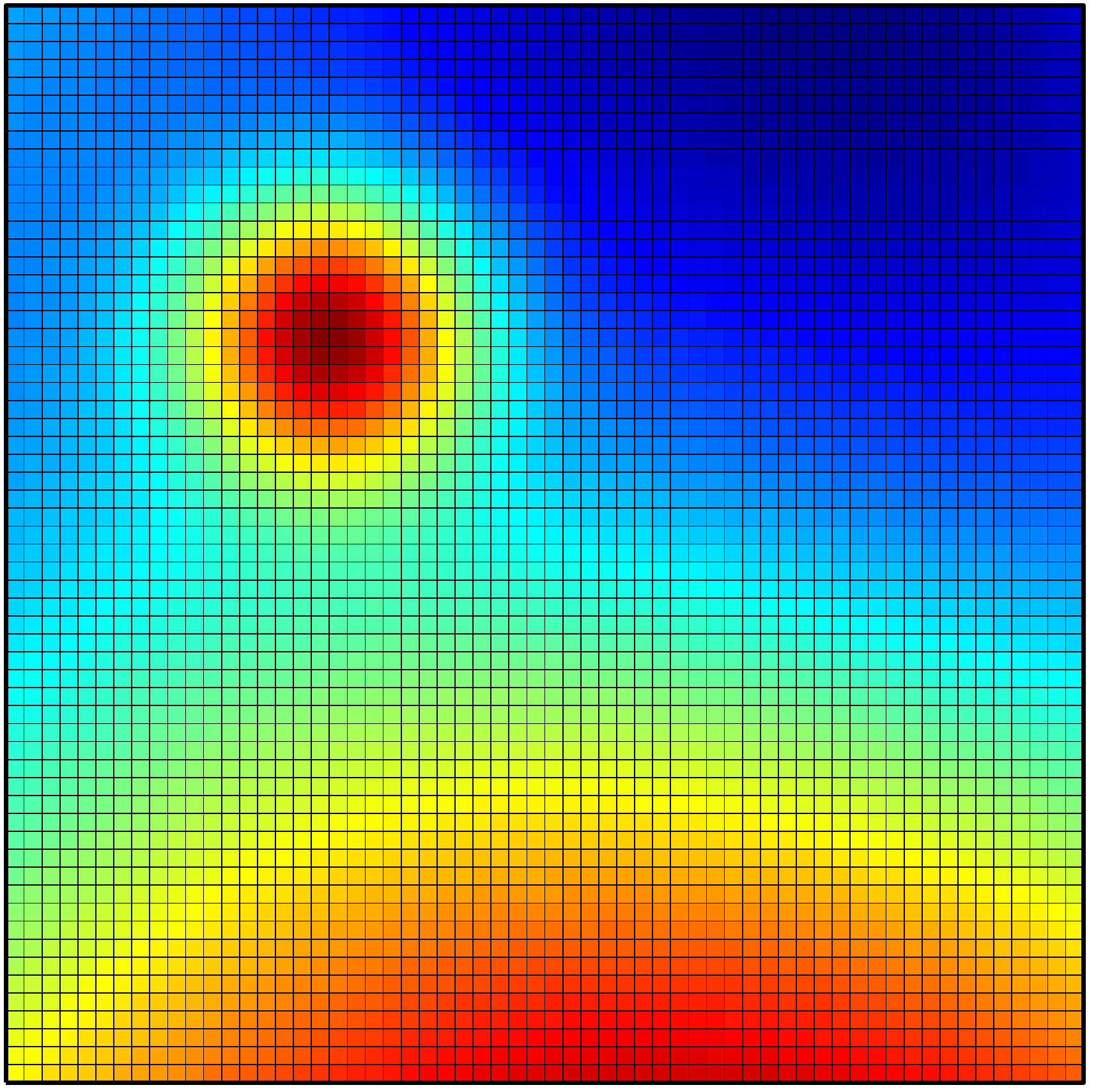}};
    \node[anchor=west,label=sketch] (sketch) at (1,0) {\includegraphics[trim=650 0 0 0,clip,height=40mm]{randomsketch.pdf}};
    \draw[thick,->] (0,0) -- (1,0);
  \end{tikzpicture}
  \caption{Randomized sketching can dramatically reduce the size of massive data sets.}
  \label{fig:sketch}
\end{figure}

Dramatically better results can be achieved via randomized algorithms,
reducing the data size as much as or more than naive deterministic subsampling can achieve, while
retaining the high accuracy of computing on the full data set.
By randomized algorithms we mean those algorithms that employ some form of
randomness in internal algorithmic decisions to accelerate time to
solution, increase scalability, or improve reliability. Examples
include matrix sketching for
solving large-scale least-squares problems (see Figure~\ref{fig:sketch}) and stochastic
gradient descent for training machine learning models.
We are not recommending heuristic methods but rather randomized algorithms that have certificates of correctness and probabilistic guarantees
of optimality and near-optimality.
Such approaches can be useful beyond acceleration, for example, in understanding how to avoid measure zero worst-case scenarios that plague methods such as  QR matrix factorization.

Randomized algorithms have a long history. The Markov chain Monte Carlo method was central to the earliest computing efforts of DOE's precursor (the Atomic Energy Commission). By the 1990s, randomized algorithms were 
deployed in regimes such as randomized routing in Internet protocols, the well-known quicksort algorithm,
and polynomial factoring for cryptography.
Starting in the mid-1990s, random forests and other ensemble classifiers have shown how randomization in machine learning
improve the bias-variance tradeoff.
In the early 2000s, compressed sensing, based on random matrices and
sketching of signals, dramatically changed signal processing.
The 2010s saw a flurry of novel randomization results in linear algebra and optimization,
accelerated by the pursuit of problems of growing scale in AI
and bringing non-asymptotic guarantees for many problems.

The accelerating evolution of randomized algorithms and
the unrelenting tsunami of data from experiments, observations, and simulations
have combined to motivate research in randomized algorithms
focused on  problems specific to DOE.
The new results in artificial intelligence and elsewhere are just the tip of the iceberg in foundational research,
not to mention specialization of the methods for distinctive applications.
Deploying randomized algorithms to advance AI for Science within DOE requires new skill sets,
new analysis, and new software, expanding with
each new application. To that end, the DOE convened a workshop to
discuss such issues.

This report summarizes the outcomes of that  workshop, ``Randomized
Algorithms for Scientific Computing (RASC),'' held virtually across
four days in December 2020 and January 2021. %
Participants were invited to provide input, which 
formed the basis of the workshop discussions as well as this report, compiled by
the workshop writing committee.
The report contains a summary of drivers and motivation
for pursuing this line of inquiry, a discussion of possible research
directions (summarized in Table~\ref{tab:RDs}), and themes and recommendations
summarized in the next two pages.

\vspace{0.5in}
\begin{table}[h!]
\begin{center}
    \caption{A sampling of possible research directions in randomized algorithms for scientific computing}
    \label{tab:RDs}
\begin{tikzpicture}
  \node[align=flush center, node font=\footnotesize,draw,rounded corners, inner xsep=10mm, inner ysep=5mm] (list) {%
    Analysis of randomized algorithms for production conditions \\
    Randomized algorithms cost and error models for emerging hardware \\
    Incorporation of sketching for solving subproblems \\
    Specialized randomization for structured problems \\ 
    Overcoming of parallel computational bottlenecks with probabilistic estimates \\
    Randomized optimization for DOE applications \\
    Computationally efficient sampling\\
    Stratified and topologically aware sampling \\
    Scientifically informed sampling \\
    Reproducibility \\
    Randomized  algorithms  for  solving  well-defined  problems  on  networks \\
    Universal sketching and sampling on discrete data \\
    Randomized algorithms for machine learning on networks \\
    Randomized algorithms for combinatorial and discrete optimization \\
    Randomized algorithms for discrete problems that are not networks \\
    Going beyond worst-case error analysis \\
    Bridging of computational and statistical perspectives \\
    Integration of randomized algorithms into coupled workflows \\
    Mergeable summaries\\
    In situ and real-time data analysis\\
    Privacy \\
    Composable and interoperable randomized abstractions for computation \\
    Use cases for randomized communication\\
    Broker abstractions for randomized input/output 
  };
\end{tikzpicture}  
\end{center}
\end{table}
%
%
%

%
%

% ---- Inserted File ----
%
\newsavebox{\thmCapacity}
\sbox{\thmCapacity}{%
\begin{theme}{Randomized algorithms essential to future computational capacity}{capacity}
  The rate of growth in the computational capacity of integrated circuits
  is expected to slow while data collection is expected to grow
  exponentially, making randomized algorithms---which depend on
  sketching, sampling, and streaming computations---essential to the
  future of computational science and AI for Science.
\end{theme}  
}

\newsavebox{\thmNovel}
\sbox{\thmNovel}{%
\begin{theme}{Novel approaches by reframing long-standing challenges}{novel}
  The potential for randomized algorithms goes beyond keeping up with
  the onslaught of data: it involves opening the door to novel approaches to
  long-standing challenges. These include scenarios
  where some uncertainty is unavoidable, such as in 
  real-time control and experimental steering; 
  design under uncertainty; and mitigating stochastic failures in novel materials, software stacks, or grid infrastructure.
\end{theme}
}

\newsavebox{\thmHW}
\sbox{\thmHW}{%
\begin{theme}{Randomness intrinsic to next-generation hardware}{randhw}
  Computing efficiencies can be realized by purposely allowing random imprecision in computations. Imprecision is inherent in emerging architectures such as quantum and neuromorphic computers.
  Randomized algorithms are a natural fit for these environments, and future computing systems will
  benefit from the co-design of randomized algorithms alongside hardware that favors certain instantiations of randomness.
\end{theme}
}

\newsavebox{\thmBarrier}
\sbox{\thmBarrier}{%
\begin{theme}{Technical hurdles requiring theoretical and practical advances}{barrier}
  Crafting sophisticated approaches that break the ``curse of dimensionality''
  via sublinear sampling, sketching, and online algorithms requires sophisticated analysis,
  which has  been tackled thus far only in a small subset of scientific computing problems.
  Foundational research in theory and algorithms needs to be multiplied many times over in order to cover
  the breadth of DOE applications.
\end{theme}
}

\newsavebox{\thmPsych}
\sbox{\thmPsych}{%
  \begin{theme}{Reconciliation of randomness with user expectations}{psych}
    Users are conditioned to certain expectations, such as viewing machine precision
    as sacrosanct, even when fundamental uncertainties  make such
    precision ludicrous.
  New metrics for success 
  can expand opportunities for scientific breakthroughs by %
  accounting for  tradeoffs among speed, energy consumption, accuracy,
  reliability, 
  and communication.
\end{theme}
}

\newsavebox{\thmExpertise}
\sbox{\thmExpertise}{%
\begin{theme}{Need for expanded expertise in statistics and other areas}{expertise}
  Establishing randomized algorithms in scientific computing
  necessitates integrating statistics, theoretical computer science, 
  data science,
  signal processing, and emerging hardware expertise alongside the
  traditional domains of applied mathematics, computer science, and
  engineering and science domain expertise.
\end{theme}
}

\newsavebox{\recfoundations}
\sbox{\recfoundations}{%
\begin{rec}{Theoretical foundations}{foundations}
  Foundational research in the theory of randomized algorithms
  to (among other issues) understand existing methods, tighten
  theoretical bounds, and tackle problems of propagating theory into
  coupled environments. The output of this research will be
  theorems and proofs to uncover new techniques and guarantees and to address new problem settings.
\end{rec}  
}

\newsavebox{\recalgs}
\sbox{\recalgs}{%
\begin{rec}{Algorithmic foundations}{algs}  
  Foundational development of sophisticated algorithms that leverage the theoretical underpinnings in practice, identifying and mending any gaps in theory, and establishing performance for idealized and simulated
  problems. The output here will be advances in algorithm analysis and understanding, prototype software, and
  reproducible experiments.
\end{rec}
}

\newsavebox{\recapps}
\sbox{\recapps}{%
\begin{rec}{Application integration}{apps}  
  Deployment in scientific applications in concert with domain
  experts. This will often require extending existing theory and
  algorithms to the special cases of relevance for each application,
  as well as application-informed sampling
  designs. The output here will be software alongside benchmarks and best practices for specific
  applications, focused on enabling novel scientific and engineering
  advances.
\end{rec}
}

\newsavebox{\rechw}
\sbox{\rechw}{%
\begin{rec}{Performance on next-generation hardware}{hw}  
  Adaptation of randomized algorithms to 
  take advantage of best-in-class computing hardware,
  from current architectures to quantum, neuromorphic, and other emerging platforms.
  The output here will be high-performance open-source software
  for next-generation computing hardware,
  including enabling efficient utilization of
  nondeterministic hardware and maximizing performance of deterministic hardware.
\end{rec}  
}

\newsavebox{\recoutreach}
\sbox{\recoutreach}{%
\begin{rec}{Outreach}{outreach}
  Outreach to a broader community
  to facilitate engagement outside the traditional computational
  science community, including experts in statistics, applied probability,
  signal processing, and emerging hardware.
  The output of this effort will be community-building workshops
  and research efforts with topically diverse teams that break new frontiers.
\end{rec}}

\newsavebox{\recstandardization}
\sbox{\recstandardization}{
\begin{rec}{Workflow standardization}{standardization}
  Standardization  of workflow, including debugging and test frameworks
  for methods with only probabilistic guarantees, software frameworks
  that both integrate randomized algorithms and provide new
  primitives for sampling and sketching, and modular frameworks for
  incorporating the methods into large-scale codes and deploying to
  new architectures.
  The output here will be community best practices and reduced barriers to contributing to scientific advances.
\end{rec}
}

% --- End Inserted File ---
%

\clearpage
\subsection*{Overarching Themes}

\usebox{\thmCapacity}

\usebox{\thmNovel}

\usebox{\thmHW}

\usebox{\thmBarrier}

\usebox{\thmPsych}

\usebox{\thmExpertise}

\clearpage
\subsection*{Recommendations}
Based on community-wide input and workshop discussions, this report recommends pursuing a research program comprising six complementary thrusts.
\usebox{\recfoundations}

\usebox{\recalgs}

\usebox{\recapps}

\usebox{\rechw}

\usebox{\recoutreach}

\usebox{\recstandardization}

%
%

%
%

%
%

%
%
%
%
% --- End Inserted File ---

\clearpage
\pagenumbering{arabic}
\setcounter{page}{1}
% ---- Inserted File ----
\section{Introduction}
\label{sec:introduction}

Randomized algorithms have propelled advances in artificial
intelligence (AI) and represent a foundational research
area in advancing AI for Science.
Future advancements in DOE Office of Science priority areas such as
climate science, astrophysics, fusion,  advanced
materials, combustion, and quantum computing all require
randomized algorithms for
surmounting challenges of complexity, robustness,
and scalability.

Advances in data collection and numerical simulation have changed
the dynamics of scientific research  and motivate the need for randomized algorithms. 
For instance, advances in imaging technologies such as X-ray ptychography, electron microscopy, electron energy loss spectroscopy, and adaptive optics lattice light-sheet microscopy collect hyperspectral imaging and  scattering data in terabytes at breakneck speed enabled by  state-of-the-art detectors. The data collection is exceptionally fast  compared with its analysis.
Likewise, advances in high-performance architectures have made exascale computing a
reality and changed the economies of scientific computing in the process.
Floating-point operations that create data are essentially free in comparison with data movement.
Thus far, most approaches have focused on creating faster hardware.
Ironically, this faster hardware has exacerbated the problem by making
data still easier to create.
Under such an onslaught, scientists often resort
to heuristic deterministic sampling schemes (e.g., low-precision arithmetic, sampling every $n$th element) and 
sacrifice potentially valuable accuracy.

Dramatically better results can be achieved via randomized algorithms,
reducing the data size as much as or more than naive deterministic subsampling while
retaining the high accuracy of computing on the full data set.
By randomized algorithms we mean those algorithms that employ some form of
randomness in internal algorithmic decisions to accelerate time to
solution, increase scalability, or improve reliability. Examples
include matrix sketching for
solving large-scale least-squares problems and stochastic
gradient descent for training machine learning models.
We are not recommending heuristic methods but rather randomized algorithms that have certificates of correctness and probabilistic guarantees
of optimality and near-optimality.
Such approaches can be useful beyond acceleration, for example, in understanding how to avoid measure zero
worst-case scenarios that plague methods such as the QR matrix factorization.

Randomized algorithms have a storied history in computing.
Monte Carlo methods were at the forefront of early Atomic Energy Commission (AEC) developments by Enrico Fermi, Nicholas Metropolis, and Stanislaw Ulam \cite{metropolis1987beginning,doi:10.1080/01621459.1949.10483310} and inspired John von Neumann to consider early automated generation of pseudorandom numbers to avoid latency costs of relying on state-of-the-art tables \cite{MR-1418-RC}. Ulam's line of inquiry was rooted in solitaire card games but aimed at practical efficiency \cite{eckhardt1987stan}:
\begin{quote}
After spending a lot of time trying to estimate them by pure combinatorial calculations, I wondered whether a more practical method than abstract thinking might not be to lay it out say one hundred times and simply observe and count the number of successful plays. 
\end{quote}
In the 1950s, Arianna Rosenbluth programmed the first Markov chain Monte Carlo implementation, which was for an equation-of-state computation on AEC's groundbreaking MANIAC I (Mathematical Analyzer Numerical Integrator and Automatic Computer Model I) computer \cite{Metropolis1953}. In subsequent years, the consideration of systems at equilibrium and study of game theory have resulted in many randomized algorithms for resolving mixed strategies for Nash equilibria \cite{Nash1951}. 
By the 1990s, randomized algorithms were 
deployed in regimes such as randomized routing in Internet protocols, the well-known quicksort algorithm,
and polynomial factoring for cryptography \cite{NRC-1992,Karp_1991}. 
In the mid-1990s, methods such as random forests and other randomized ensemble classifiers improved accuracy
in machine learning, demonstrating that ensembles built from independent random observations
can yield superior generalization \cite{breiman1996bagging,Breiman01}.
In the early 2000s, compressed sensing, based on random matrices and
sketching of signals, dramatically changed signal processing \cite{Davenport2013}.
The National Academies' Mathematical Sciences in 2025 report~\cite{math2025} stated:
\begin{quote}
  It revealed a protocol for acquiring information,
  all kinds of information, in the most efficient way. This research
  addresses a colossal paradox in contemporary science, in that many protocols
  acquire massive amounts of data and then discard much of it, without
  much or any loss of information, through a subsequent compression stage,
  which is usually necessary for storage, transmission, or processing purposes. 
\end{quote}
The work showed that the traditional Shannon bounds of information
theory can be overturned whenever the underlying signal has structure and that randomized algorithms are key to this development.

The 2010s saw a flurry of novel results in linear algebra and optimization,
accelerated by problems of increasing scale in artificial intelligence.

The accelerating evolution of randomized algorithms and
the unrelenting tsunami of data from experiments, observations, and simulations
have combined to motivate research in randomized algorithms
focused on  problems specific to DOE.
The new results in AI and elsewhere are just the tip of the iceberg in foundational research,
not to mention specialization of the methods for distinctive applications.
Deploying randomized algorithms to advance AI for Science within DOE requires new skill sets,
new analysis, and new software, expanding with
each new application. To that end,  DOE convened a workshop to
discuss such issues.

This report summarizes the outcomes of that workshop,  ``Randomized
Algorithms for Scientific Computing (RASC),'' held virtually across
four days in December 2020 and January 2021.%
\footnote{In contrast to past workshops, such as Scientific Machine Learning \cite{Baker2019}
and AI for Science \cite{AI-for-Science},
this workshop was held virtually because of the COVID-19 pandemic,
which prevented  travel in the winter of 2020/2021.
The virtual format
had the benefit of enabling much broader engagement than past
in-person workshops.}
The first two days of the workshop, the ``boot camp,''
focused on highly interactive technical presentations from experts
and had 453 participants.
The second part of the workshop, held one month later,
focused on community input and had 204 fully engaged participants.
Participants in both parts were invited to provide inputs
during, in between, and after the sessions. These inputs
have formed the basis of this report, which was compiled by
the workshop writing committee.

The report is organized as follows. \Cref{sec:drivers} describes the need for a colossal leap
in computational capacity, across the board,
motivated by ever-larger and more heterogeneous data collection,
larger-scale and higher-resolution simulations,
bounding uncertainty in high-dimensional inverse problems,
higher-complexity real-time control scenarios,
inherent randomness in emerging computational hardware itself,
and scientific applications of AI.
Ideas for foundational and applied research in randomized algorithms that address
these needs are described in \cref{sec:research}. These ideas  range from linear and nonlinear systems, to algorithms for discrete and combinatorial
problems,
to random sampling strategies and streaming computations, to software abstractions.
Much attention is focused on providing a combination of theoretical robustness (i.e., assurances of
correctness for model problems), efficiency, practicality, and relevance for problems
of interest to DOE.
We conclude with high-level themes and recommendations in \cref{sec:them-recomm},
not least of which is the need for reconciling user expectations with
the results of randomized algorithms and the need to engage broader
expertise (e.g., from statistics) than has historically been needed
in the ASCR program.
The appendices give further details of the workshop:
see \cref{app:workshop-agenda} 
for the workshop agenda,  \cref{app:participants} for a full list of participants, and \cref{app:acks} for acknowledgments.

%

%

%
%
%
%
%
%
%

%
%
%
%
%
%
%

%

%
%

%
%
%

%
%

%
%

%
%

%
%

%
%
%

%
%
%

%
%

%
%

%

%
%
%
%
% --- End Inserted File ---

\clearpage
% ---- Inserted File ----
\section{Application Needs and Drivers}
\label{sec:drivers}

Over the next decade, DOE anticipates scientific breakthroughs that depend on
modeling more complex chemical interactions yielding higher-capacity batteries,
computing on emerging hardware platforms such as quantum computers with inherent randomness,
analyzing petabytes of data per day from experimental facilities to understand the subatomic structure of biomolecules, and
discovering rare and previously undetected isotopes. 
Achieving these science breakthroughs requires a colossal leap in DOE's capacity for
massive data analysis and exascale simulation science.

Algorithmic advances have always been the key to  such progress, but
the scale of the challenge over the next decade will oblige DOE to forge into the domain of
\emph{randomized algorithms}.
Supporting this new effort will mean recruiting experts outside  the traditional areas of 
computational science and engineering, high-performance computing, and mathematical modeling
by attracting and involving experts in
applied probability, statistics, signal processing, and theoretical computer science.
Advances in randomized algorithms have been accelerating in the past decade,
but there is a high barrier to integration into DOE science.
This is in large part because DOE has \emph{unique needs that require domain-specific approaches}.
In this section we highlight specific DOE applications,
the challenges of the coming decade,
and the potential of randomized algorithms to overcome these hurdles.

% ---- Inserted File ----
\subsection{Massive Data from Experiments, Observations, and Simulations}
\label{sec:facilities}
\sublead{C.~Kamath}

\begin{figure}
  \centering
  \includegraphics[width=4in]{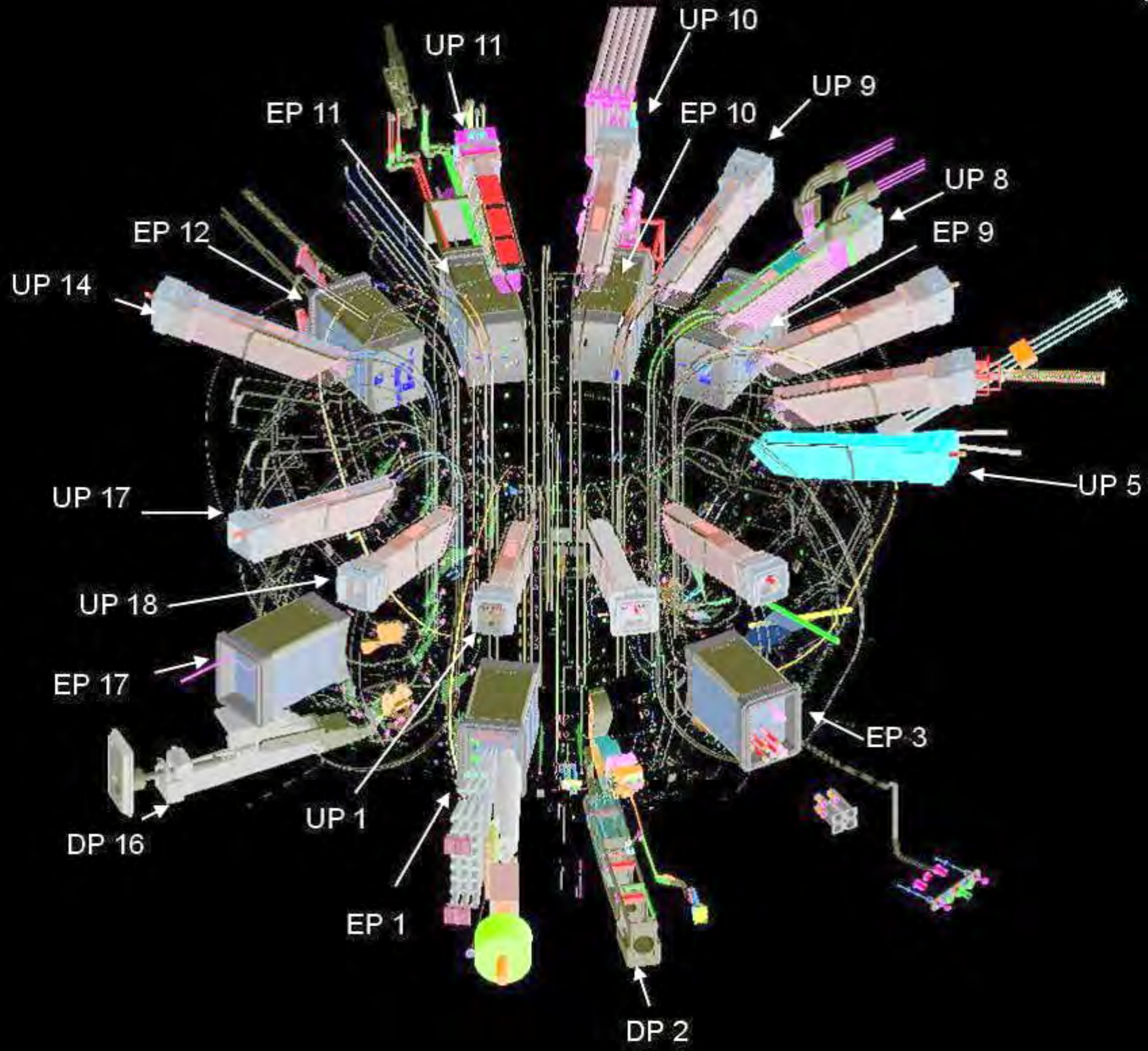}
  \caption{Schematic of \href{https://www/iter.org/newsline/-/3360}{ITER diagnostics}~\cite{iterdiagwebpage} illustrating some of the instruments that will generate two petabytes of data per day at the \href{https://www.iter.org/newsline/-/3534}{ITER Scientific Data Centre}~\cite{iterdatawebpage}. Randomized algorithms offer a potential solution to the challenge of processing of these massive, complex data sets.}
  \label{fig:iter_diagnostics}
\end{figure}

The DOE Office of Science operates several national science user facilities, including accelerators, colliders, supercomputers, light sources, and neutron sources \cite{ASCR-EOD-Workshop-Report:2015}. Spanning many different disciplines, these facilities generate massive amounts of complex scientific data through experiments, observations, and simulations. For example, by the year 2035, ITER, the world's largest fusion experiment~\cite{iterwebpage}, will produce two petabytes of data every day, with 60 instruments measuring 101 parameters (Figure~\ref{fig:iter_diagnostics}) during each experiment or ``shot.'' The data will be processed subject to a wide range of time and computing constraints, such as analysis in near-real time, during a shot, between shots, and overnight, as well as remote analysis and campaign-wide long-term analysis~\cite{choi2020:fusion}. 

\begin{figure}
  \centering
  \subfloat[Visualization of STEM interacting with a silicon atom in a graphene hole.]{\includegraphics[scale = 1.5]{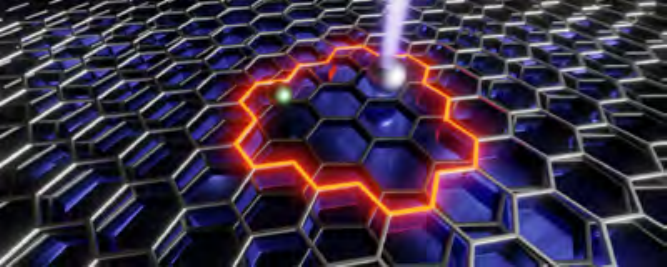} } \\
\subfloat[Reconstruction of potential on a torus expanded coordinate system.]{\includegraphics[scale = 0.165]{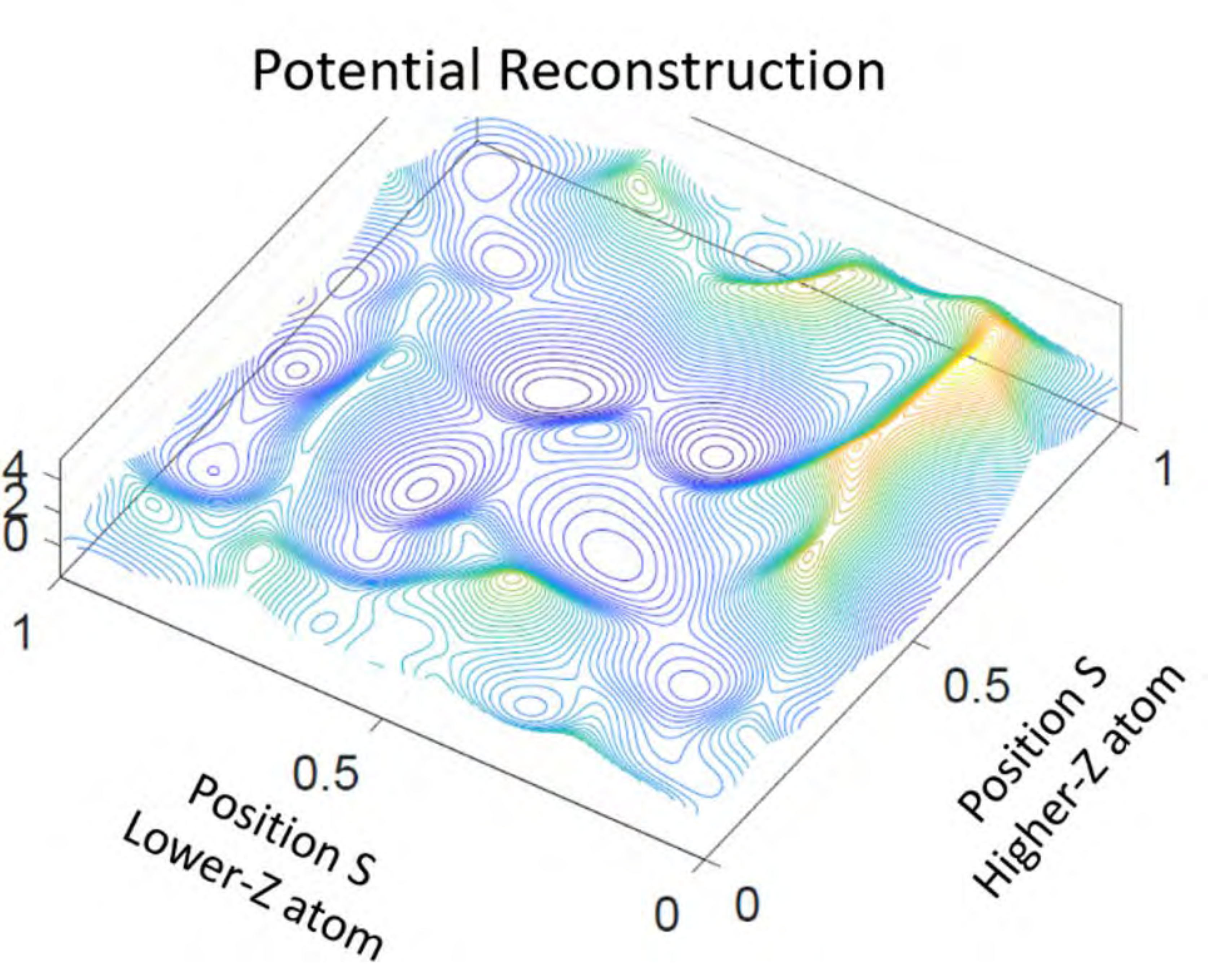} } \quad
\subfloat[%
Tracking of atomic position using 
auxiliary particle and backward doubly stochastic differential equation filters.
]{\includegraphics[scale = 0.44]{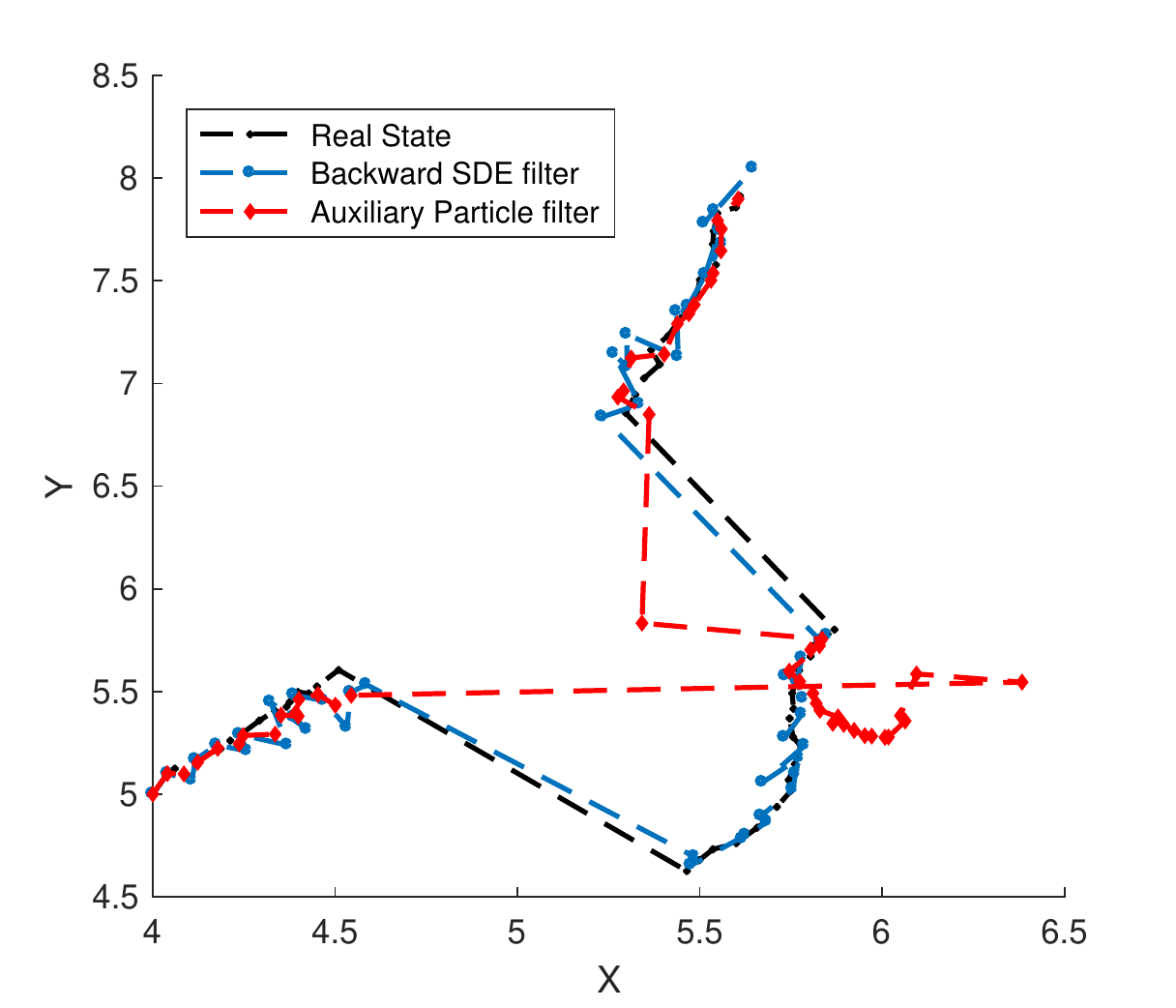} }
  \caption{A scanning transmission electron microscope is capable of measuring and modifying the location of a silicon atom in a graphene hole. Current mathematical methods cannot reconstruct the energetic landscape from the sparse and noisy measurements from the microscope or track the atoms accurately.  However, newly developed randomized algorithms are providing fundamental tools to achieve the goal of real-time atomic control~\cite{DYCK2021116508,bao2020levy}}
  \label{fig:atomic_forge}
\end{figure}

These constraints, as well as the volume and complexity of the data, require new advances in data-processing techniques. Similar requirements are echoed by scientists as they prepare their simulations for the exascale era~\cite{DOEexacrosscut2018}. The nanoscale facilities at DOE also provide challenges as scientists aim to control matter at the atomic scale through the Atomic Forge~\cite{Kalinin-Atom}. Manipulating atoms by using a scanning transmission electron microscope involves real-time monitoring, feedback, and beam control. Scalable randomized algorithms will be essential for achieving success in this new field of atom-by-atom fabrication (Figure~\ref{fig:atomic_forge}).

\begin{callout}
In order to fully realize the benefits of the science enabled by  DOE facilities, the techniques currently used for processing the data must be enhanced to keep pace with the ever-increasing rate, size, and complexity of the data.  
\end{callout}

In order to fully realize the benefits of the science enabled by these DOE facilities, the techniques currently used for processing the data must be enhanced to keep pace with the ever-increasing rate, size, and complexity of the data. For simulation data generated on exascale systems, these techniques include compression, in situ analysis, and computational steering, while experimental and observational data require robust, real-time techniques to identify outliers, to fit machine learning models, or to plan subsequent experiments. In contrast to simulations that can be paused to reduce the data overload, experimental and observational data  often must processed as it arrives in an unrelenting stream, or  else risk losing parts of the data.

Randomized algorithms offer a solution to this challenge of processing massive data sets in near-real time by introducing concepts of sparsity and randomness into the algorithms currently in use. However, several technical issues must be addressed before such algorithms are broadly accepted and used by scientists. Most important is a need to understand the uncertainties in the reliability and  reproducibility of the results from these algorithms, especially given their ``random'' nature. In experiments where not all the data being generated can be stored, the critical information to be saved must be correctly identified, even though the randomized algorithms  sample only a subset of the data stream; and predicting whether a sample is useful or not can be difficult. Randomized algorithms must also be able to process data at different levels of precision, as well as data stored in different data structures, such as images and time series, including structured and unstructured data from simulations. 

Addressing some of these roadblocks to the use of randomized algorithms in processing massive data sets requires longer-term research, but  several near-term opportunities exist. Two areas where randomized algorithms could prove  useful are data reduction through various compression techniques and acceleration of data analysis. These algorithms could also be more accurate than periodic sampling in some problems and more efficient than the use of dense linear algebra, lead to better communication efficiency in high-performance computing, and allow more sophisticated analysis to be performed in real time. In problems involving both simulations and experiments/observations, randomized algorithms could enable data assimilation of larger data sets, improve data-driven simulations by allowing faster use of experimental data, perform better parameter estimation by exploring larger domains, and open new venues for improvement as ideas for use of these algorithms in simulations are transferred to experiments/observations and vice versa. The insight and compression capabilities provided by randomized algorithms could also be used for improved long-term storage of data. The specific areas where additional research is required to accomplish these improvements are outlined in \cref{sec:research}.

Technical issues are not the only roadblocks to the use of randomized algorithms in processing massive data sets. From a societal viewpoint, these algorithms are currently not well understood enough for domain scientists to be comfortable incorporating them into their work. Often lacking is an  awareness of opportunities where such algorithms may make processing massive data sets tractable, as well as  a lack of robust software that is scalable to massive data sets. Addressing both the technical and societal concerns would help scientific communities processing data from experiments, observations, and simulations to accept and benefit from randomized algorithms. If successful, this would result in reduced time to science, more effective science, and a greater return on the investment DOE has made in its facilities.

%
%
%
%
% --- End Inserted File ---
% ---- Inserted File ----
\subsection{Forward Models}
\label{sec:forward-models}

\sublead{C.~Yang}

\begin{figure}
  \centering
  \includegraphics[width=0.6\textwidth]{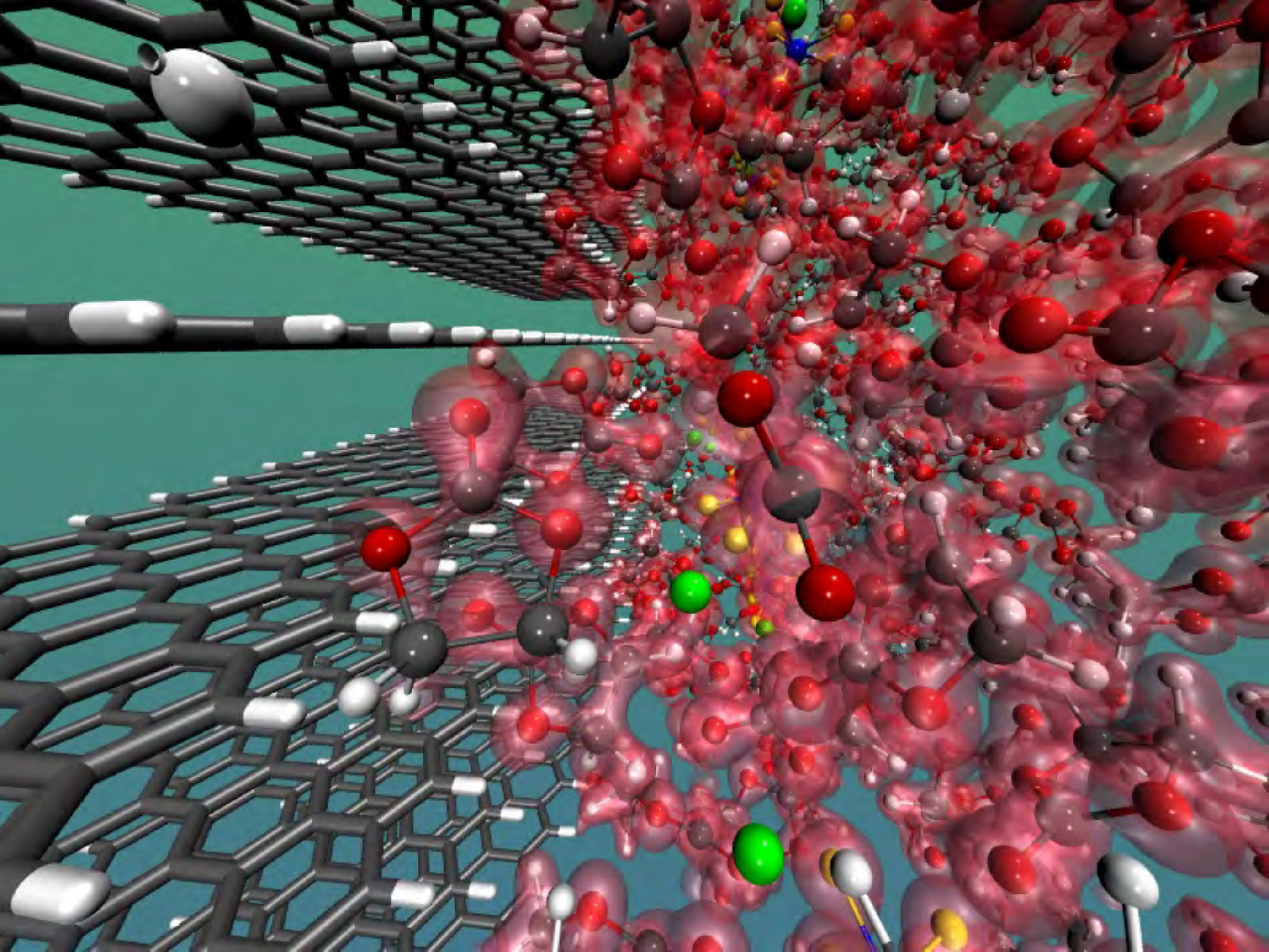}
  \caption{Snapshot of an ab initio molecular dynamics simulation of the solid-electrolyte interphase (SEI) of a lithium-ion battery. A nonlinear eigenvalue with millions of degrees of freedom needs to be solved at each time step. Thousands of time steps are required to generate a trajectory from which properties of the SEI can be characterized. Randomized trace estimation may be used to reduce the computational complexity in each time step from $O(n^3)$ to $O(n)$ and significantly shorten the simulation time.     (provided by J. Pask; a similar figure appeared in~\cite{BES2015EXA})}
  \label{fig:battery_sei_aimd}
\end{figure}

\begin{figure}
  \centering
  \includegraphics[width=0.8\textwidth]{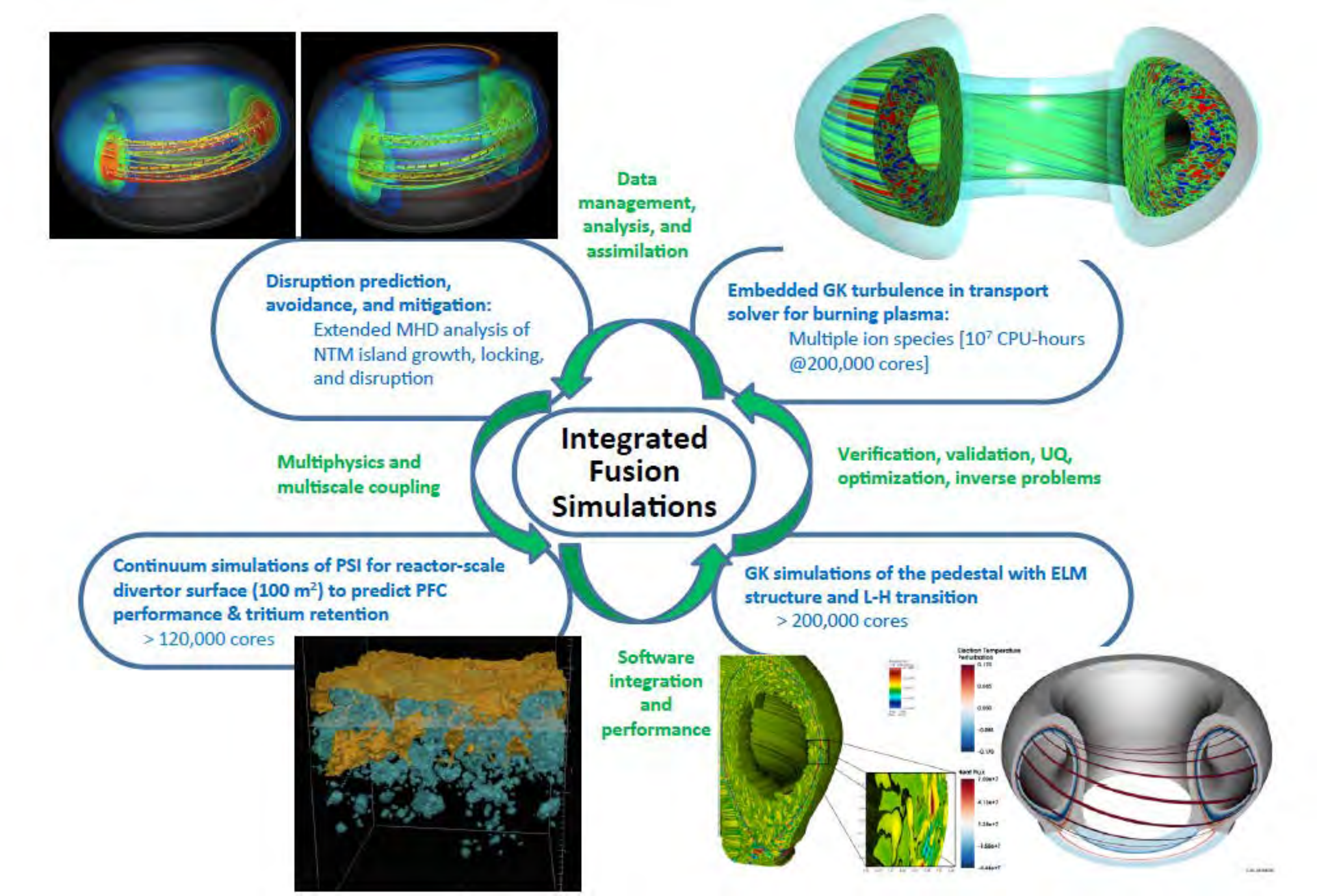}
  \caption{A whole-device model of a tokamak when using the highest-fidelity physics models available for core transport, the pedestal and scrape-off layer, plasma--materials interactions at the wall, and global MHD stability. Simulating such a model is extremely challenging and requires a tremendous amount of computational resources. Randomized projection methods can be used to significantly reduce the computational cost. (from~\cite{fusion_report})}
  \label{fig:WDM_fusion}
\end{figure}

\begin{figure}
  \centering
  \includegraphics[width=0.8\textwidth]{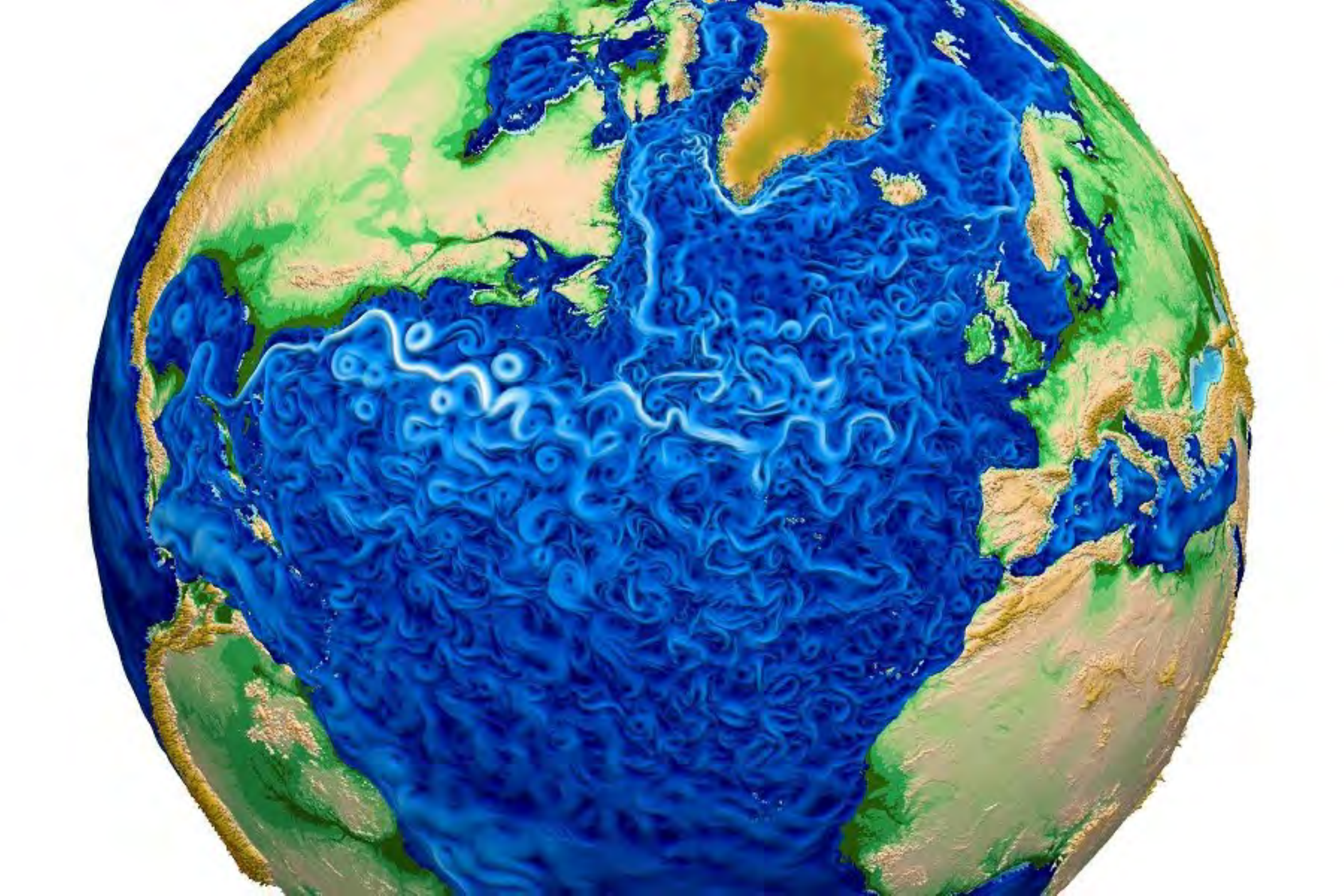}
  \caption{DOE’s Energy Exascale Earth System Model(E3SM) simulation showing global eddy activity. Performing this simulation, which combines atmosphere, ocean, and land models, plus sea- and land-ice models, requires DOE exascale computing resources. Randomized algorithms can significantly accelerate the computation and are well suited for exascale platforms. (from \url{e3sm.org} and the article \url{https://ascr-discovery.org/2020/11/climate-on-a-new-scale/})}
  \label{fig:simearth}
 \end{figure}

As we gain a deeper understanding of a wide range of physical phenomena, forward models become more complex. 
A scientific inquiry often begins with a hypothesis in the form of a forward model that describes what we believe to be the fundamental laws of  nature and how different physical processes interact with each other.  Mathematically, these models are represented by algebraic, integral, or differential equations.
They contain many degrees of freedom to account for the multiscale and multiphysics nature of the underlying physical processes. 
For example, to model the interaction of fluids and structures, we need to include velocity, pressure for the fluid, and displacement of the structure. To simulate a photovoltaic system, we need take into account electron excitation, charge separation, and transport processes, as well as interface problems at a device level. To understand electrochemistry in a Lithium-ion battery, we need to simulate the dynamic interface between the electrode and electrolytes during the charging and discharging cycles (Figure~\ref{fig:battery_sei_aimd}). To perform a whole-device modeling of a tokamak fusion reactor, we need to combine the simulation of core transport, plasma materials interaction at the wall, and global MHD stability analysis (Figure~\ref{fig:WDM_fusion}). To model the fully coupled Earth system, we need to consider the interactions among the atmospheric, terrestrial/subsurface, and ocean cryosphere components (Figure~\ref{fig:simearth}).

Such complexity challenges our ability to perform computer simulations with sufficient resolution to validate our hypotheses by comparing with scientific observations and experimental results. 
A high-fidelity simulation to predict extreme events in a climate model at a spatial resolution of 1 kilometer yields an extremely large number of coupled algebraic, integral, and differential equations with the number of variables, $n$, in the billions. 
The complexity of existing numerical
methods for solving these problems is often $O(n^p)$ for some integer power $p>1$, and the number of degrees of freedom $n$ can be millions or billions. For example, the complexity of a density-functional-theory-based electronic structure calculation for weakly correlated quantum many-body systems is $O(n^3)$, with $n$ as large as a million. More accurate models for strongly correlated systems such as the coupled cluster model may require $O(n^7)$ floating-point operations, with $n$ in the thousands.
The computational bottleneck is often in the solution of large-scale linear algebra problems.
Because of the nonlinearity of many forward models, these equations need to be solved repeatedly in an iterative procedure. Even with the availability of the exascale computing resources, performing these types of simulations is extremely challenging .

\begin{callout}
A high-fidelity simulation to predict extreme events in a climate model at a spatial resolution of 1 kilometer yields an extremely large number of coupled algebraic, integral, and differential equations, with the number of variables, $n$, in the billions. 
\end{callout}

Furthermore, because of model uncertainties, such as fluctuation
and noise, multiple simulations need to be performed in order to obtain ensemble averages.


\subsubsection{State of the Art}

Randomized algorithms have proven  effective at overcoming some of the challenges discussed above.
In particular, randomized projection methods have been used to reduce the dimension of some problems by projecting linear and nonlinear operators in the model onto a randomly generated
low-dimensional subspace and solving a smaller problem in this subspace~\cite{RAMOR19,allakutz2019}. Although projection methods have been used in a variety of applications, the traditional approach often requires a judicious construction (e.g., through a truncated singular value decomposition or the application of a Krylov subspace method) of a desired subspace that captures the main characteristics of the solution to the original problem.  This step can be costly. For problems that exhibit fast singular value decay, randomized projection works equally well but at much lower cost.
Fast sketching strategies based on structured random maps can potentially accelerate computations dramatically.

In addition to being used as an efficient technique for dimension reduction, randomized algorithms have played an important role in reducing the complexity of 
linear solvers for discretized elliptic partial differential equations (PDEs) and Helmholtz equations~\cite{hss2011}.  By taking advantage of the low-rank nature of the long-range interaction in the Green's function, randomized algorithms allow us to construct compact representations of approximate factors of linear operators~\cite{martinsson_tropp_2020,Halko2011,Woodruff14}, preconditioners for iterative solvers~\cite{RA4LS,2016_ghysels_randomized_multifrontal,2020_keyes_hierarchical_hierarchical}, or direct solvers that construct data sparse representations of the inverse of the coefficient matrix for many linear systems  \cite{doi:10.1137/12087116X,2019_martinsson_fast_direct_solvers}.
They are also used in  constructing low-rank approximations to tensors, for example, the two-electron integral tensors that appear in Hartree--Fock or hybrid functional density-functional-theory-based electronic structure calculations~\cite{isdf,isdftddft,isdfthc}.

\begin{callout}
In addition to being used as an efficient technique for dimension reduction, randomized algorithms have played an important role in reducing the complexity of linear solvers for discretized elliptic PDEs and Helmholtz equations.
\end{callout}

Randomized algorithms have also been used to compute physical observables such as
energy density through randomized trace estimation~\cite{Hutch90,RATrace11,Ubaru17}. This technique has been used successfully in ground- and excited-state electronic structure calculations for molecules and solids~\cite{RADFT19,RAGW17}. The use of this type of randomized algorithm often leads to linear complexity scaling with respect to the number of atoms, which is a major improvement compared with methods that require solving a large-scale eigenvalue problem with $O(n^3)$ complexity.

For high-dimension problems, Monte Carlo methods have been used extensively to generate random samples of desired quantities to be averaged over (as an approximation to a high-dimensional integral)~\cite{QMC12,MCTS84,MCTurb}.

The random samples must be generated according to an underlying distribution function that may be unknown. A widely used technique to achieve this goal is the Metropolis--Hasting  algorithm, also known as the Markov chain Monte Carlo. This algorithm has been successfully used in kinetic models to study rare events and chemical reactions in large-scale  molecular systems~\cite{kmcvoter}, quantum many-body models to approximate the ground state energies of strongly correlated systems~\cite{QMC12,QMC86}, and turbulent flow models~\cite{MCTurb} used to study atmosphere-ocean dynamics, combustion, and other engineering applications.

\subsubsection{Opportunities}

Although randomized algorithms have been developed and used in several applications,  many more applications  can potentially benefit. This is particularly true in a multifidelity or multiscale framework in which we need to couple simulations across multiple spatial and temporal scales, for example, a wind farm modeled at $\mathcal{O}(10m)$ resolution coupled with climate simulations run at $\mathcal{O}(10km)$. Randomized algorithms might be used to provide avenues for enhancing the information shared in such a coupling and open the door to questions related to uncertainty.  One example is the use of stochastic subgrid process models, that  is, computational models that provide source terms from processes with spatial and time scales below mesh resolution, to represent missing information (introduced by grid-level filtering) probabilistically, rather
than deterministically, by sampling subgrid source term from appropriate distributions conditioned nonlocally on grid-level evolution.  Probabilistic models learn such distributions from direct numerical simulation  data and deploy learned samplers in large-scale simulations.

\begin{callout}
Although randomized algorithms have been developed and used in several applications,  many more applications  can potentially benefit from randomized algorithms. This is particularly true in a multifidelity or multiscale framework in which we need to couple simulations across multiple spatial and temporal scales.
\end{callout}

In general, sampling methods play an important role in the convergence of randomized algorithms. A good sampling method can lead to a significant reduction in variance and, consequently, a reduced number of samples required to achieve high accuracy.  Although importance sampling and umbrella sampling methods have been developed in several applications in quantum and statistical mechanics, room for improvement still remains. These techniques can potentially be applied to a much broader class of problems.

Despite the tremendous success randomized linear algebra has enjoyed in accelerating key matrix computation kernels of many simulations, much is yet to be done to extend these techniques to multilinear algebra (tensor) and nonlinear problems.

In order to achieve both high accuracy and efficiency, a hybrid method may be desirable in which a low-complexity randomized algorithm is used to provide a sufficiently good approximation that can be refined or postprocessed by a deterministic method.

Integrating randomized algorithms in the existing deterministic algorithms-based simulation pipeline would require a careful examination of practical issues including data structure and parallelization.  Randomized or hybrid randomized and deterministic algorithms have the potential to reduce the complexity of many of the most demanding simulation problems to $\mathcal{O}(n^2)$ or $\mathcal{O}(n)$. They can be much more scalable than existing methods and are suitable for exascale machines.

With the help of randomized algorithms we can continue to push the envelope of large-scale simulation in many scientific disciplines and significantly improve the fidelity of models needed to provide accurate descriptions of a broad range of natural phenomena and engineered processes. Randomized algorithms are particularly powerful in tackling high-dimensional problems that are not amenable to conventional deterministic approaches. They
are game changers for solving some of the most challenging computational problems in many applications.

%
%
%
%
%
%
%
%
%

%
%
%
%
% --- End Inserted File ---
% ---- Inserted File ----
\subsection{Inverse Problems}
\label{sec:inverse-problems}
\sublead{B.~Wohlberg}

The analysis of experimental measurements plays a fundamental role in basic science and mission areas within  DOE. In many cases the quantities of interest cannot be directly measured and must instead be inferred from related quantities that are amenable to measurement. This inference is typically posed as an \emph{inverse problem}, where the corresponding \emph{forward problem} describes the physics of the measurement process that maps the quantities of interest to the measurements.

\subsubsection{Large-Scale Inverse Problems for Science Data}

A classical example of an inverse problem is X-ray computed tomography (CT), in which a map of the internal density of an object is inferred from a sequence of radiographs taken from different view directions.

\begin{figure}[htbp]
  \centering
  \includegraphics[width=0.95\textwidth]{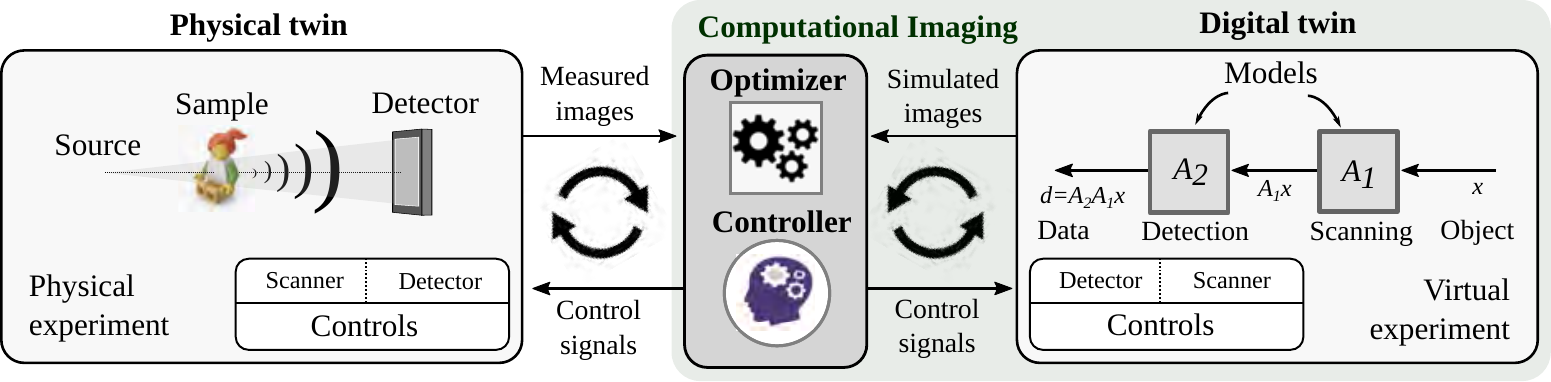}
  \caption{Illustration of the main components of a computational imaging system, including a sensing system and an inverse problem solver consisting of a ``digital twin'' or ``forward model'' of the imaging system and an optimizer. (courtesy of D. G\"ursoy, Argonne National Laboratory.)}
  \label{fig:comp_image}
\end{figure}

In addition to CT, numerous other imaging techniques involving  inverse problems (\cref{fig:comp_image}) are relevant to DOE applications including materials science, parameter estimation for complex models related to global climate change, oil and gas exploration,  groundwater hydrology and geochemistry, and  calibration of computational cosmology models. Many inverse problems arise in experiments at DOE imaging facilities, which can produce  large volumes of data (e.g., up to 10 GB/s at the current Linac Coherent Light Source, with 100 GB/s predicted for next-generation instruments~\cite[Sec. 15.1.4]{ASCR-EOD-Workshop-Report:2015}).
Examples of such problems include 
CT~\cite{hidayetolu2019memxct} and CT reconstruction from a set of coherent imaging experiments~\cite{barutcu2020simultaneous}. Such data sets are rapidly growing in size as imaging technology improves and new experimental facilities are constructed, leading to an urgent need for improved computational capabilities in this area. 

The primary challenges to be addressed are the following.
\begin{itemize}
    \item Many of these problems are of a sufficient scale that they can  be solved only by using advanced high-performance computing resources (e.g., see ~\cite[p. 158]{ASCR-EOD-Workshop-Report:2015}) that are in limited supply. While computing power is important, the  most significant constraint is usually the need to keep the entire reconstruction and measured data set in working memory. Online or streaming algorithms that avoid this need would allow large-scale problems to be solved on a much broader range of computing hardware.
    \item DOE  imaging  facilities are heavily oversubscribed, with the result that experiments have to be conducted within a limited time window. Thus,   calibration or other issues that might degrade the utility of the experiments are often discovered only when reconstructions are computed after the experiment has been completed. A capability for real-time or progressive reconstructions would be of enormous value in addressing this difficulty~\cite[Sec. PRO 1]{BES2019Roundtable}.
    \end{itemize}

\begin{callout}
Inverse problems at DOE facilities produce very large volumes of data, for example, 10 GB/s at the current Linac Coherent Light Source, with 100 GB/s predicted for planned instruments.
\end{callout}

While careful design of massively parallel algorithms can provide close to real-time reconstructions of relatively large problems given a sufficient number of compute nodes~\cite{hidayetolu2019memxct}, not all inverse problems are amenable to this type of problem decomposition, and such large-scale computing facilities (256,000 cores in~\cite{hidayetolu2019memxct}) are not widely available.

Randomization methods offer a number of different approaches to address these challenges:
\begin{enumerate}
    \item Application of inherently randomized algorithms such as stochastic gradient descent, randomized Levenberg--Marquardt, and derived methods for solving the optimization problems associated with the inverse problems~\cite{bjarkason-randomized-2018, bergou-stochastic-2018, sun2019online, tang2020fast, sun2021asyncred}
    \item Use of randomized methods for solution of subproblems of non-randomized algorithms (e.g., use of sketching for solving the linear subproblem related to the data fidelity term within an alternating direction method of multipliers (ADMM) algorithm~\cite{boyd2011admm})
    \item Use of randomized algorithms for efficiently training machine learning  surrogates of physics models, with efficient use of queries of expensive physics models, and for efficiently using large, complex data sets
\end{enumerate}

\subsubsection{Quantifying Uncertainty in Inverse Problems through Randomized Algorithms}

\begin{figure}[htbp]
  \centering
  \includegraphics[width=0.5\textwidth]{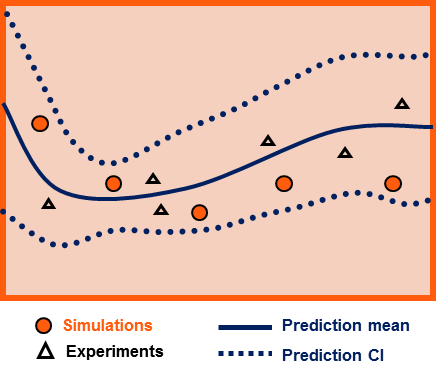}
  \caption{Illustration of the estimation of mean and confidence interval from both simulations and experiments. (diagram by ``Jiangzhenjerry,'' CC BY-SA 3.0 license, source \url{https://commons.wikimedia.org/w/index.php?curid=19797809})}
  \label{fig:bias-correction}
\end{figure}

Quantifying uncertainty in inverse problems and parameter estimation problems is an important step toward providing confidence in estimation results (see, e.g., Figure~\ref{fig:bias-correction}), but emerging sensors and novel applications pose significant challenges. For example, LIDAR sensors, for which the usage is rapidly increasing, provide  much higher resolution information about wind profiles. While the instrument measurement error is well understood,  little alternative information exists that can be used to assess the accuracy of a reconstructed 3D wind field at that resolution (tens of meters in space and seconds in time).  
The increased expectations of capabilities lead researchers to consider applications with parameters of high spatiotemporal heterogeneity. Thus,   uncertainty must be expressed  over parameter spaces of vastly increased dimensionality with respect to problems currently approachable via state-of-the-art uncertainty quantification techniques.
Moreover, this parameter heterogeneity, the complexity of the physics of these novel applications, and the advancement of sensing techniques result in experimental data sets composed of various data sources with vastly different data collection protocols in different regions of their state space and errors of different probabilistic characteristics.
Despite the significant advances in methods for Bayesian inference, efficiently leveraging the physical constraints and laws characterizing applications of interest in conjunction with data remains a significant computational challenge.
This is particularly true where the encoding of physical constraints and laws is in the form of  expensive computational simulations with their own complexity drivers; this challenge is in turn compounded by the heterogeneity of data sources described above.

\begin{callout}
Quantifying uncertainty in inverse problems and parameter estimation problems is an important step toward providing confidence in estimation results.
\end{callout}

Randomization methods offer an unprecedented promise for tackling the challenges in uncertainty quantification.
Specifically, randomization techniques may lead to gains in computational efficiency in various places along the probabilistic modeling pipeline  (e.g., accelerated solution of subproblems, the training of machine-learning-based surrogate models).
Furthermore, randomization can support the solution of stochastic programs associated with approximate Bayesian inference such as variational inference.
Other examples of research challenges and opportunities include the following:
\begin{enumerate}
\item Leveraging randomization methods for probabilistic inference with streaming data. %
  Integrating online data assimilation algorithms (e.g., particle filters) together with randomization techniques such as sketching, approximate factorizations, and randomized calculations with hierarchical matrices may lead to improvements in scalability and efficiency.
  Furthermore,  the impact of compression via randomization of streaming data on the inference process needs to be explored.
\item Analysis of the effect of randomization in sketching, data compression, and other techniques on the convergence and bias of probabilistic reconstructions.
  Such an analysis would distinguish, in the probabilistic setting, the uncertainty stemming from randomized methods and the uncertainty inherent in the inverse problem due to observation errors, for example.
\end{enumerate}

Carrying out these advances will result  not only in faster solution to the analysis problems but also in increased predictive capabilities of the resulting computational tools.

%
%
%
%
%

%
%
%
%
%

%
%
%
%

%
%
%
%
%

%
%
%
%
% --- End Inserted File ---
% ---- Inserted File ----
\subsection{Applications with Discrete Structure}
\label{sec:discrete-driver}
\sublead{J.~Restrepo}

We next consider problems with discrete structure, notably networks and graphs.
In the context of network science, specific application drivers are familiar:
critical infrastructures such as the Internet  and power grids, as well as  biological, social, and economic structures.

 Faster and better ways to
 analyze, sample, manage, and sort discrete events, graphs, and connected data streams will have dramatic impact on networked applications.
 In what follows, we highlight  two application areas that demonstrate the mathematical and algorithmic challenges.
 
 Community detection is arguably one of the most widely used network analysis procedures.
 The basic purpose is to detect and categorize aggregations of activity in a network. Many  algorithms have  been developed for this purpose. Practitioners often run these algorithms on their data, but their data is actually a snapshot of the whole (e.g., a Twitter snapshot from the entire stream). These methods yield no guarantee on the structure of the original, not the observed, network. 
 Current sampling methods include stratified sampling techniques along with topological/dimensionality reduction techniques (i.e., the Johnson--Lindenstrauss lemma \cite{joli}). Some algorithms  reduce quadratic time to near-linear time complexity for specific classes of problems, but the scope is frequently narrow.
 
 Researchers have little understanding of the mathematics in downsampling such a complex structure, and the current state-of-the-art approaches are based on unproven heuristics (see surveys~\cite{LF06, maiya2011benefits, ahmed2014network}). 
 Associated with sampling and searching is a general class of algorithms called streaming.
A rich history of streaming algorithms exists in the theoretical computer science and algorithms community. While some of these methods have had significant success in practice (e.g., the HyperLogLog sketch~\cite{FlMa07}), much of this field has remained purely mathematical. 
Advances in graph sampling would provide methods to subsample a graph so that community detection algorithms on the subsample would provide guarantees on the entire structure. 

New graph search challenges appear in the context of black-box optimization techniques, as a result  of its relevance to many machine learning and reinforcement learning contexts. For graphs that are larger to manage or store than the resources available on a single machine, the search requirements on the discrete space are impractical or expensive.

Randomized algorithms
may have an impact on streaming, and in general on sampling and searching, and thus an impact on community detection and other analysis needs in network science. Further, new uses for searches in connection with tasks associated with machine learning will also benefit, should randomization lead to efficiencies associated with informing neural nets with data associated with graph structures.

\begin{figure}
    \centering
    \includegraphics[scale=0.4,angle=-90]{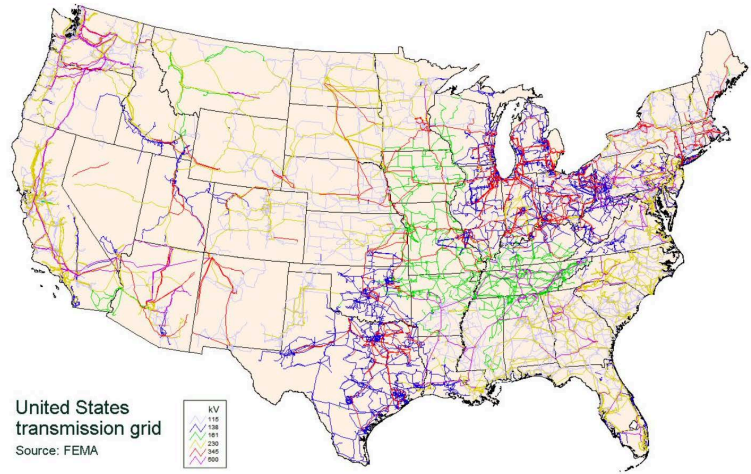}
    \caption{The U.S. electric power transmission grid is over 120,000 miles long. The figure suggests that complexities in the analysis and management of the network result from node heterogeneity, the federated nature of the management of the network, and the complexity of sources and users. Inherent stochasticity in this and other networks could be exploited by proposing ways to speed up sampling, searching, and  analysis of the network via randomized algorithms. (source: FEMA.}
    \label{fig:powergrid}
\end{figure}

\paragraph{Power Grid} A critical national security challenge is the maintenance of the integrity of national power grids (Figure \ref{fig:powergrid}) under adverse conditions caused by nature or humanity. 

With more renewable power sources on the grid, such as solar and wind,  uncertainty on the power supply side increases, caused by variations in weather; and potential disruptions become even more difficult to manage. 
Thus, grid planners and operators need to assess grid management and response strategies to best maintain the integrity of the national power grid under extremely complex and uncertain operating conditions.  
Currently, the Exascale Computing Project (ECP) subproject ExaSGD (Optimizing Stochastic Grid Dynamics at ExaScale) is developing methods to optimize the grid’s response to a large number of potential disruption events under different weather scenarios \cite{cindyph1}. 
For example, in the Eastern Interconnection grid one must consider thousands of contingencies  taking place under thousands of possible (and impactful) weather scenarios.  ExaSGD is developing massively parallel algorithms for security-constrained optimal power flow analysis that could involve simultaneous optimization of millions of power grid realizations. 
Future research will need to address other power grid analysis questions that require discrete optimization. 
For example, the unit commitment problem, selecting how much steady-generation power (e.g., coal-based or nuclear-based) to buy and from where, is a large mixed-integer program for given demand and generation estimates.  
Finding the worst-case placement of $k$ outages is a bilevel discrete optimization problem since the network can reoptimize to mitigate the damage.

\begin{callout}
Grid planners and operators need to assess grid management and response strategies to best maintain the integrity of the national power grid under extremely complex and uncertain operating conditions. 
\end{callout}

Randomized algorithms could help in a number of places  in the near term.
Progress in these will also have an impact on other networked infrastructure systems and beyond: 
\begin{enumerate}
\item
Randomized rounding to find feasible solutions for, for  example, unit commitment given a fractional solution to a relaxation. Finding a provably good solution, or even finding a feasible solution with reasonable probability, is valuable. Using approximations for the DC optimal power flow (DCOPF) may speed up interdiction problems. 
\item
Fast approximation of DCOPF.  Randomized approximation schemes for network flow exist \cite{cindyph3}.  Can these be made faster, if less accurate?  Can these network-flow approximation algorithms be extended to include the phase-angle constraints from DCOPF?
\item
Randomized selection of scenarios.  Is there a way to select a finite set of scenarios that are representative of the full set?  
There has been some experimental success on finding average damage estimates on stochastic versions of network-interdiction problems (e.g., \cite{cindyph2}). 
The largest current supercomputers tend to have GPUs on the nodes, and GPUs are particularly strong when used for randomized algorithms.  
One might even use metaheuristics, even without provable performance guarantees, to find better solutions in practice.
\end{enumerate}

\begin{figure}
     \centering
     \includegraphics[width=2.5in]{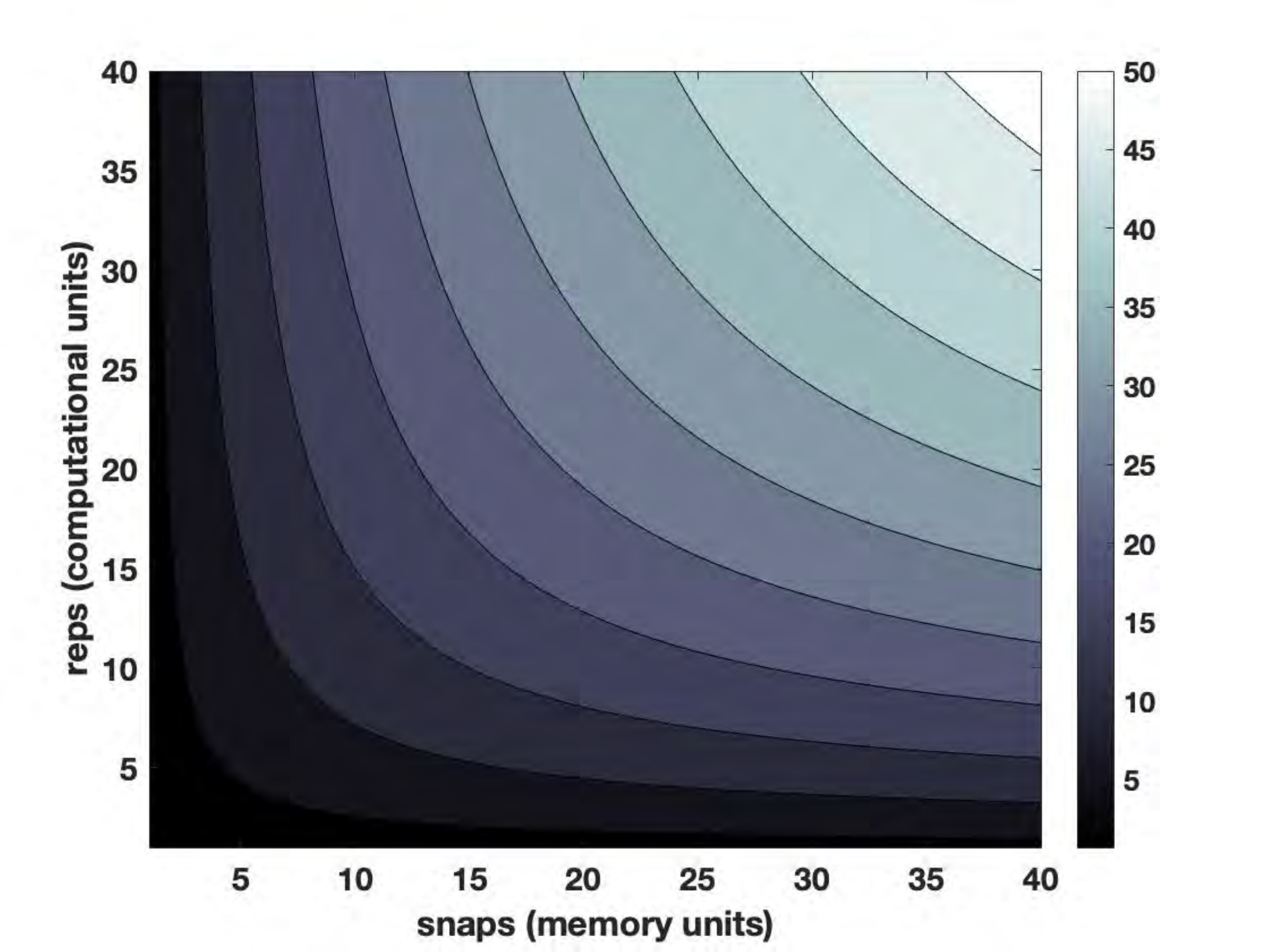}
     \caption{Logarithm of the total operations required to compute a derivative via checkpointing as a function of storage (snaps) and time or stage steps (reps); 
     see \cite{snaprep}. Randomization could bring down the computational costs when the differentiation products do not have to be exact in structure and/or in value.}
     \label{fig:snapsreps}
 \end{figure}

\paragraph{Automatic Differentiation}
Automatic differentiation (i.e., algorithmic differentiation) is a methodology for computing derivatives of functions defined by algorithms \cite{griewankwalther,autodif}. Automatic differentiation is the key technique underlying 
backpropagation for computing gradients of 
neural networks \cite{backprop,Schmidhuber2015}, but automatic differentiation can also
account for data-dependent control flow. 
Combinatorial problems abound in automatic differentiation. A fundamental
method is so-called checkpointing \cite{snaprep}, wherein derivatives are found with an awareness of finite storage and run-time resources on a given machine, trading one for the other, depending on the resource limitations. Figure~\ref{fig:snapsreps} shows curves of constant effort required in obtaining a derivative by exchanging storage resources and run times. Automatic differentiation often models computation using directed acyclic graphs, and many automatic differentiation algorithms
can be interpreted as graph transformations.  Sparsity structure in
Jacobians and Hessians is detected by using Bayesian 
probing~\cite{Bayesian-probing} or related techniques and exploited
by using graph coloring techniques \cite{Gebremedhin2005WCI}.

 Randomized algorithms have the potential to address many of the
combinatorial challenges in automatic differentiation.
For example, in checkpointing,
 randomization could be exploited to overcome computational resource challenges associated with storage and/or run time by  exchanging fidelity in obtaining derivatives.
  Such an exchange is an acceptable tradeoff  in the context of many optimization applications as well as in sensitivity analysis.
  \begin{callout}
  In many applications an approximate gradient is adequate. Randomized algorithms could lead to lower computational costs in  differentiation-related computations  by harnessing randomized linear algebra and randomized changes in the automatic differentiation algorithms.
  \end{callout}

%
%
%
%
% --- End Inserted File ---
% ---- Inserted File ----
\subsection{Experimental Designs}
\label{sec:experimental-designs}
\sublead{D.~Vrabie \& P.H.~Zwart}

Experimentation in real and computational environments forms the basis for hypothesis-driven scientific discovery. When planning laboratory experiments, changing control parameters in an accelerator or nuclear reactor, or performing any task associated with planning a new course of action to collect new data or in response to new information just collected, in combination with models for forecasting dynamic response and all other knowledge available, experimental design theory can aid in making optimal choices. 

Advances in technology, such as brightness improvements in light sources or robots for automated chemical synthesis, are rapidly pushing the complexity of scientific experiments to a level where human intuition can no longer keep up with the  high dimensionality of the decision landscape that needs to be explored in order to select the best possible next actions under uncertainty.

While experimental design approaches, such Gaussian process--based strategies, alleviate some of these  problems, we are rapidly encountering bottlenecks due to the computational cost associated with some of the underlying algorithms that, when implemented naively, can have computational overheads that exceed the time required to perform the experiment without advanced decision-making algorithms. Furthermore, the traditional scientist-guided approaches for selection of critical parameters that rely on human intuition could introduce bias in the experimental design and the end results. 

\subsubsection{State of the Art}
The vast majority of experimental design approaches require some sort of uncertainty quantification at their core, since this uncertainty or functions thereof are the driving force that guides experimental design choices  \cite{Rasmussen2004-yr}. Approaches such as Gaussian processes and Bayesian neural networks provide easy access to uncertainty quantification but can incur significant overhead depending on the problem. The underlying computational bottlenecks within experimental design or autonomous experiments are typically related to matrix inversion, often recast as solving a large linear system of equations, and the global optimization of some utility function that provides guidance on the next set of actions to take. Randomized algorithms will have a major impact on these types of problems, enabling a computationally efficient exploration of decision space by balancing utility and uncertainty reduction. 

Experimental design questions that are encountered can be roughly grouped in two categories: (1) \emph{one-shot designs}, in which a data collection strategy is determined up front and cannot be changed, or only at great cost, once data acquisition is initiated, and (2) \emph{adaptive designs}, in which new measurements have the ability to influence the data acquisition schemes.    

The theory of how to design one-shot experimental designs is well established \cite{Ryan2007-yd}, with examples ranging from the design of where to place wireless 5G transmitters, traffic monitoring sensors, temperature or pH sensors in reactors, to the design of clinical trials, or to the placement of fixed-position direct radiation monitoring systems. 

\begin{figure}
  \centering
  \includegraphics[width=4in]{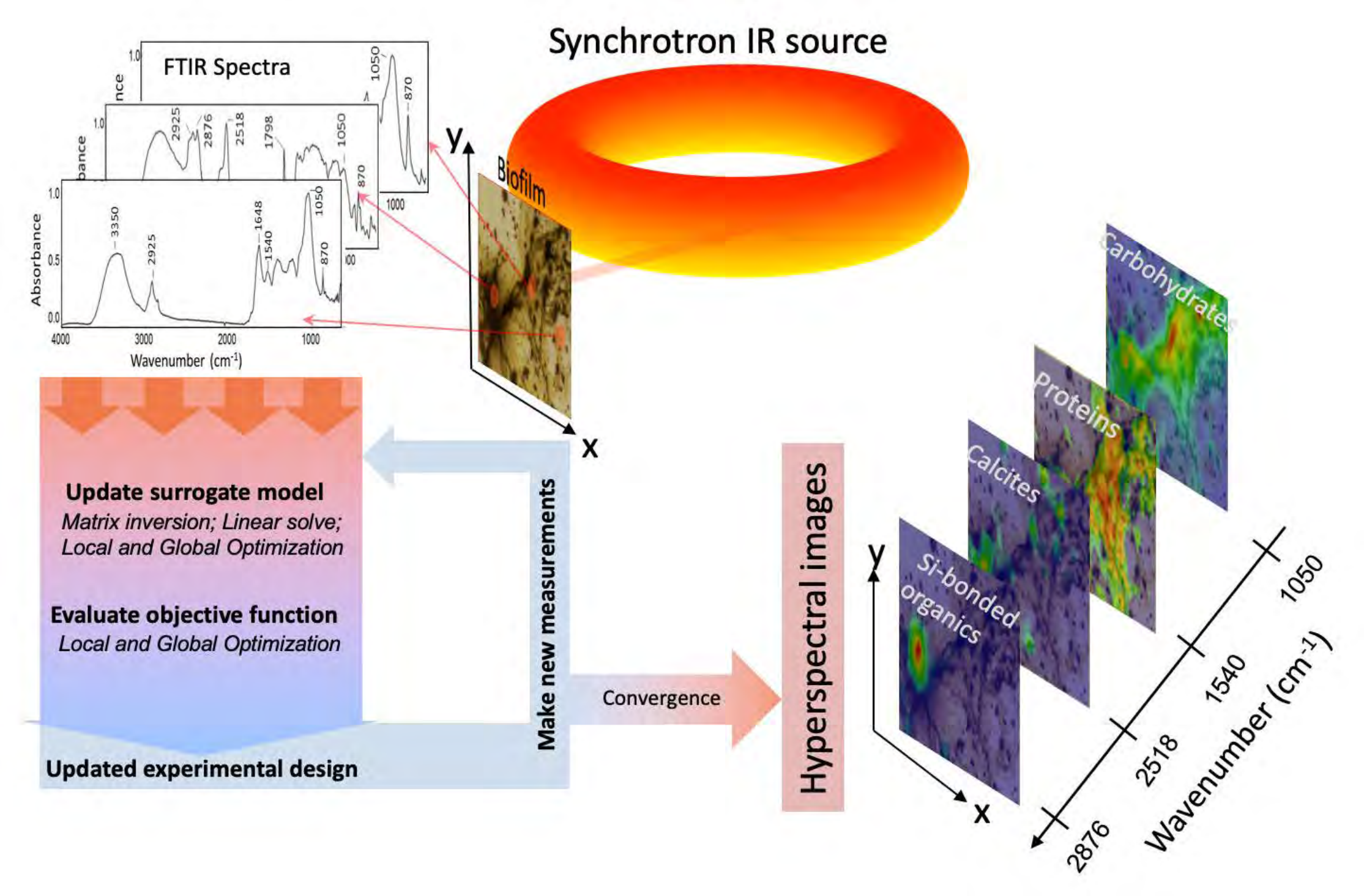}
  \caption{
  At the DOE large-scale scientific user facilities, real-time feedback loops provide continuous updates on experimental design, enabling extremely efficient autonomous data acquisition. Some subtasks in this workflow are expensive when using deterministic approaches and can cause  a bottleneck in decision-making. Randomized algorithms can alleviate these problems, providing rapid feedback at timescales compatible with experiments that require fast response.
  (figure courtesy of BSISB \& CAMERA, LBNL). }
  \label{fig:Experimental_Design}
\end{figure}

Autonomous decision-making systems are replacing the intuition of the scientist and can scan through the data and make smart decisions about how the experiment should proceed. This capability is critical in contexts characterized by high-complexity dynamics and high-dimensional decision spaces. In the experimental sciences, for instance, the use of adaptive experimental designs is rapidly becoming commonplace (Figure \ref{fig:Experimental_Design}). Beamlines at DOE large-scale scientific user facilities such as the National Synchrotron Light Source, Advanced Photon Source, and Advanced Light Source are utilizing adaptive design approaches to improve the throughput and usage of scientific instruments \cite{Noack2020-bb,Melton2020-yt}. These approaches build, interrogate, and update a surrogate while an experiment is running and provide rapid feedback on how to perform measurements. In so-called self-driving autonomous laboratories this notion is pushed even further, where artificial intelligence/machine learning approaches provide suggestions on which samples need to be synthesized \cite{Ren2018-vv,Pollice2021-mq}. A similar situation is encountered in running large-scale simulations, where choices of system parameters that can change the outcome of the results need to be tuned, for instance, to ensure reproducing related observational data. In all these approaches, the underlying computational complexity can rapidly escalate such that approximation methods are required to train the hyperparameters of surrogate models or efficiently interrogate these models in order to obtain new, optimal experimental design parameters.

\subsubsection{Opportunities for Randomized Algorithms}

While the majority of applications in experimental design have  focused on controlling and providing feedback on continuous state variables, challenges in materials design and bioengineering require the handling of discrete and combinatorial decision spaces as well \cite{Lewis2012-wm}. The integration of randomized algorithms approaches in these spaces will enable the deployment and integration of fast, scalable decision-making frameworks to a diverse application space. Randomized algorithms can potentially be used to construct data-driven surrogate models or to hybridize multifidelity data-driven and physical models for expensive/complex systems. 

When developing new randomized algorithm approaches, the availability of formal verification of stability and probabilistic performance characteristics will be of great importance because it will provide the end user of these algorithms with correct expectations and the ability to obtain the right tradeoff between precision and run time. A thorough understanding of the error properties of randomized algorithm approaches is especially important when the cost of making a mistake is very expensive, for instance in the control of accelerators or fusion reactors, such as tokamaks and stellarators, or when not exploring a parameter space can be costly, for instance in materials or molecular design.  

The use of randomized sketching algorithms, for instance, can simultaneously regularize and reduce computational complexity without impacting the accuracy of the overall procedure. Alternatively, randomized algorithms that provide well-understood tuning options that can balance accuracy and computational complexity in a predictable fashion could be the optimal approach for cases where medium- or even low-accuracy answers are useful, as long as these answers come with an associated uncertainty estimate.

Data is the new frontier for enhancing the predictive capabilities of global-scale models of the Earth and environmental systems, understanding water cycles, and predicting extreme events. Real-time feedback from dynamic systems provides the opportunity to explore the complex decision landscapes more agilely and more effectively. Computational models powered by the world’s fastest computers must guide the experimental data collection, aid in interpreting the data, ultimately inform follow-up actions such as the design of new experiments, or serve as a basis for public policy. Randomized algorithms play a critical role in advancing the application and use of autonomous experimental designs: as the range of applications and the availability of data increase, the need for general-purpose, high-performance, well-tested,  plug-and-playable algorithms and software is of paramount importance.

\begin{callout}
Randomized algorithms play a critical role in advancing the application and use of autonomous experimental designs: as the range of applications and availability of data increases, the need for general-purpose, high-performance, well-tested,  plug-and-playable algorithms and software is of paramount importance. 
\end{callout}

%

%

%

%

%

%

%

%
%
%
%
%

%
%

%

%
%
%
%
%
%

%
%
	
%

%

%

%
%

%

%
%
%
%
% --- End Inserted File ---
% ---- Inserted File ----
\subsection{Software and Libraries for Scientific Computing}
\label{sec:1f:libraries}
\sublead{S.~Wild}

Numerical software and libraries are a cornerstone of scientific computing and continue to transform science \cite{Perkel2021}. Such libraries can enhance productivity of computational scientists in many ways, including by reducing development time and enabling performance portability. 
  Libraries that include implementations of randomized algorithms would allow scientists to focus on the use and application of these algorithms for addressing grand-challenge domain-science problems. 
  Extending the reach, reliability, and understanding of randomized algorithms for DOE's complex software and system stacks would help realize performance gains on problems and architectures not yet imagined.

\begin{figure}
    \centering
    \includegraphics[width=0.95\linewidth]{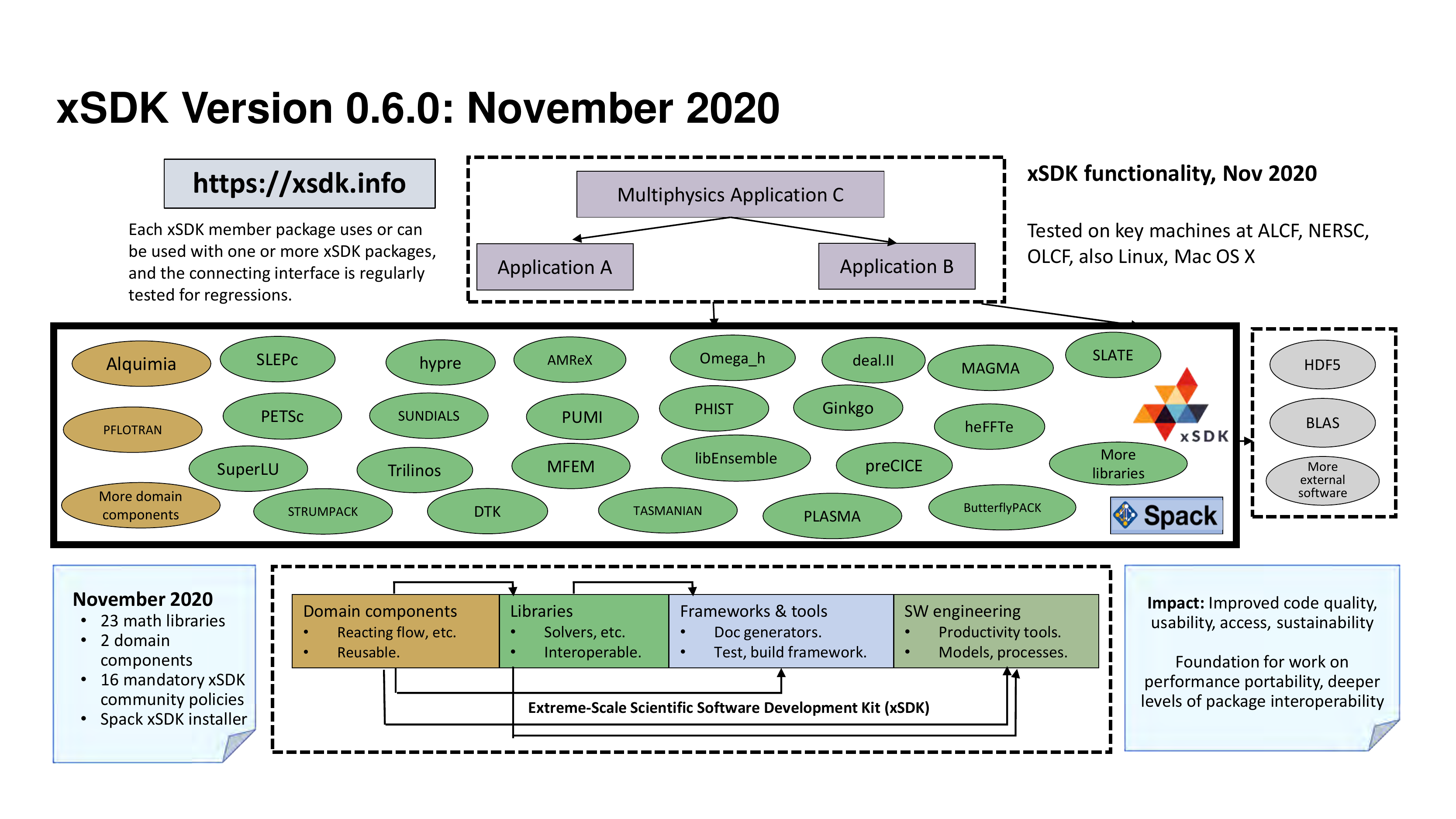}
    \caption{xSDK is an example of a component  of  the current DOE ecosystem of exascale math libraries \cite{bartlett2017xsdk}.
    The diverse DOE hardware and software stacks represent both consumers and providers of randomized algorithms for scientific computing.
    }
    \label{fig:xsdk}
\end{figure}

\subsubsection{Current Status}

In recent years, development of open-source numerical libraries for scientific computing has focused on addressing the challenges associated with emerging exascale computing architectures and enabling the solution of big data problems. Emerging hardware and special-purpose accelerators are also a driver and are discussed in  \cref{sec:emerginghardware}.

DOE's Exascale Computing Project \cite{ECPST2020}) and SciDAC programs have centralized much of the development of production software. Community-driven efforts such as the Extreme-scale Scientific Software Development Kit (xSDK \cite{bartlett2017xsdk}) have transformed the state of interoperability among math libraries used on DOE's leadership-class computing facilities (\cref{fig:xsdk}).

With few exceptions (e.g., Monte Carlo--based software), 
production libraries in use on DOE compute systems have addressed challenges distinct from those arising in randomized algorithms. 
Traditionally such libraries have focused on deterministic algorithms that produce highly accurate results with provable guarantees. One expects that the results are reproducible in the sense that the output is bitwise identical each time the algorithm runs under the same software and hardware conditions.
These guarantees are generally established based on a specified precision level for the underlying elementary operations, all of which are assumed deterministic. An enduring example of such a numerical setting is the LINPACK benchmark \cite{DonLusPet2003}, which has been used to measure performance of the top supercomputers in the world for nearly three decades.

At the same time, deterministic precision levels have received significant attention. For example, recent years have seen the introduction of a veritable zoo of floating-point conventions (e.g., bfloat16, TensorFloat, fp24, PXR24) beyond traditional IEEE standards. 
A significant driver of such developments has been data-intensive computing and special-purpose and commodity hardware and accelerators. 
Similarly, mixed- and variable-precision techniques are of increasing interest \cite{Abdelfattah2020survey,Haidar2020}. Exploration and adoption of these techniques are a recognition of performance gains realizable by allowing libraries and software to exploit multiple (deterministic) precision levels.

Randomized techniques have been used extensively for empirical performance optimization and software testing to find bugs \cite{Ali2010}. The basis for these approaches is a recognition of the tradeoffs among the applicability, accuracy, and expense relative to deterministic techniques such as formal verification or analytic performance optimization. 

\subsubsection{Challenges and Opportunities}
The development of libraries of randomized algorithms shares many of the challenges of producing production machine learning software frameworks for high-performance computing \cite{Berry2015ml} 
and using randomized testing and performance optimization techniques. 
For example, shared modeling challenges include considering metrics and solution characteristics beyond simplistic floating-point operation-based and machine-precision-based quantities.
For some problems, randomized algorithms offer the potential for accuracy levels beyond machine precision or attainable by classical deterministic methods. 
\begin{callout}
For some problems, randomized algorithms offer the potential for accuracy levels beyond machine precision or attainable by classical deterministic methods.
\end{callout}

Similarly, the trends driving hardware technology %
(\cite[Sec.~2]{vetter2018extreme})
have resulted in increasingly heterogeneous and nondeterministic computing paradigms in order to extend performance gains. 
For example, as math libraries and software seek to avoid synchronization and mitigate faults whenever possible, the costs of preserving bitwise reproducibility has begun to outweigh the benefits for many scientific use cases.
In other cases, the process of pushing the hardware and software layers to the computing fabric also results in a loss of such reproducibility. Hybrid classical and post-Moore computing workflows (\cref{sec:emerginghardware}) are further contributing to nondeterminism in computing environments.

The growing recognition of the tradeoffs between performance and achieving traditional notions of reproducibility are only a first step in leveraging  nondeterminism for scientific advances and efficiency. The subtle distinction between random data (e.g., from nondeterministic hardware) and intentionally randomized operations (from a randomized algorithm) poses challenges for software development and debugging 
as well as validation and verification.
Standardized benchmarks for randomized algorithms are lacking, despite the fact that they  have been repeatedly identified as a need for advancing progress and co-design for DOE scientific computing %
\cite[Sec.~16]{AI-for-Science}. The benchmarks would come with well-defined notions of convergence/correctness. Convergence for deterministic iterative solvers is typically described in terms of an iteration budget or residual tolerance.
Bringing randomized algorithms and their associated benchmarks to a similar status would be a breakthrough in terms of enabling software-hardware co-design and facilitating optimization for diverse architectures and computing environments. Since randomized algorithms offer an approach for addressing problems where data is too large to fit in memory, benchmarks would further understanding about which algorithms perform best for given input sizes on specific systems.

The use of AI-inspired and other automated techniques to improve programmer productivity is also recognized as a  DOE grand-challenge computing problem (\cite[Sec.~4.1]{vetter2018extreme},  \cite[Sec.~9]{AI-for-Science}). An example of such an approach is to automatically synthesize software programs based on a scientific user's intent \cite{ProgSynth2021}. Other uses include the automation of compilation, testing, and debugging of numerical software. The search spaces, both discrete and continuous, that arise in such problems are prohibitively large. Randomized algorithms offer a means to navigate such spaces for design goals within defined resource requirements.

%
%

%

%
%
%
%
% --- End Inserted File ---
% ---- Inserted File ----
\clearpage
\subsection{Emerging Hardware}\label{sec:emerginghardware}
\label{sec:emerging-hw}
\sublead{A.~DeGennaro}

Many of the problems of interest to applied mathematics are computationally intensive. Modern problems in optimization, uncertainty quantification, and engineering design involve evaluating the output of sophisticated computer models over high-dimensional parameter spaces. As maintenance of Moore’s law slows due to physical constraints in hardware manufacturing, it is imperative to pursue research that expands the capabilities of new computational paradigms and hardware. Quantum computing and neuromorphic computing are particular examples of such emerging paradigms. At the same time, randomization has already proven to be a powerful technique for computational acceleration, and its role in these computing schemes should be researched. \begin{callout}A central question for future progress will be how to co-design emerging hardware and randomized algorithms in a way that is optimized for particular tasks (e.g., speed, error-proneness).\end{callout}

\begin{figure}
    \centering
    \includegraphics[scale=0.25]{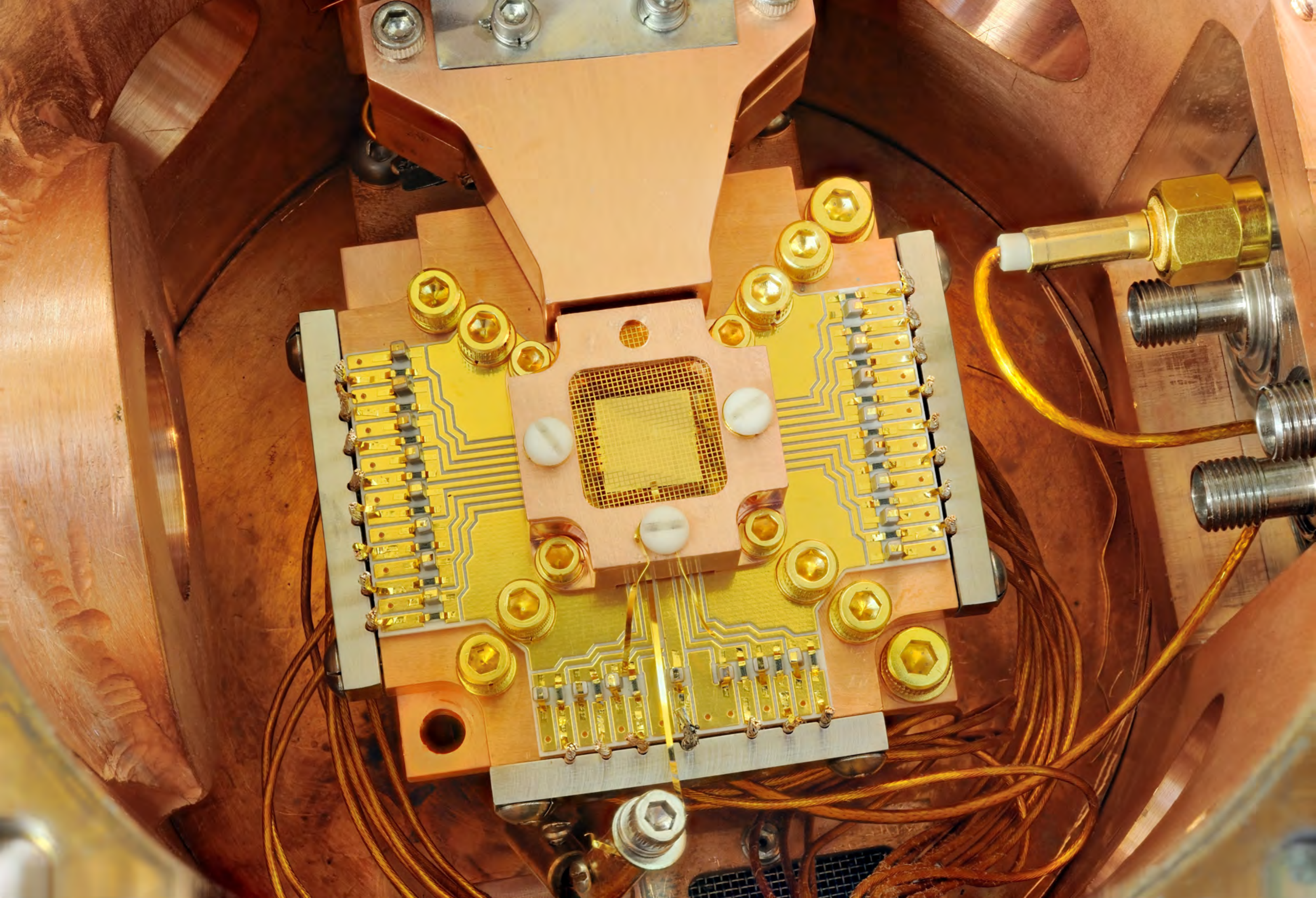}
    \caption{NIST device used for ion trapping in quantum computing (source: \url{https://commons.wikimedia.org/wiki/File:Quantum_Computing;_Ion_Trapping_(5941055642).jpg}, public domain CC-PD-Mark license).}
    \label{fig:quantum}
\end{figure}

Quantum computing \cite{bennett2000quantum, preskill2018quantum} is an emerging computational paradigm that could benefit from randomization. Quantum computers (Figure~\ref{fig:quantum}) currently can solve only  small problems, chiefly because  of noise in qubit states and quantum decoherence. Co-design of quantum hardware with randomized algorithms might help expand the size of problems that could be computed. Opportunities exist to discover and apply quantum-informed downsampling as an algorithm to load a small but statistically representative sample of a given data set onto a quantum computer. This hints at a more general motivation that quantum computing can inform classical algorithms. It also suggests that in the quantum realm, and in the classical realm as well, randomization should be used to find novel initialization schemes and more efficient methods for optimization and exploration. Randomization can help optimize quantum algorithms, such as quantum annealing and quantum Monte Carlo, and can help with the efficient solution of problems in optimization (e.g., quadratic unconstrained binary optimization) and linear algebra (e.g., eigenvalue decomposition). If successful, the integration of randomized algorithms with quantum hardware could result in significantly faster time to solution, as well as privacy preservation. Quantum supremacy could also potentially aid the solution of machine learning problems that require large data sets.

Neuromorphic computing \cite{merolla2014million, davies2018loihi} is another computational paradigm that could benefit from randomization. Graph algorithms as currently implemented on neuromorphic computers are deterministic in nature \cite{HaImHu17,ScHaMiAd19} and could benefit from the usual speed and efficiency of randomization. Neuromorphic sensors are plagued by a significant amount of noise and variations that must be reduced. A good noise model of these sensors is clearly needed. Further research could potentially reveal a better understanding of the tradeoffs of randomized algorithms with noise and provide an opportunity to balance computation robustness with error tolerance. Randomization might also help us understand the computational capabilities of neuromorphic computers. Co-design of randomized algorithms with neuromorphic computers could lead to a novel random programming model or help researchers understand and evaluate the parameter space for neuromorphics (e.g., spike thresholds, synaptic delays). If successful, neuromorphic computers could lead to better algorithms for emulating random processes and solving stochastic/partial differential equations. They could also lead to fast computing algorithms with low memory requirements associated with data.

%

%

%

%

%

%
%
%
%
% --- End Inserted File ---
% ---- Inserted File ----
\subsection{Scientific Machine Learning}
\label{sec:sciml}
\sublead{S.~Wright}

\begin{figure}[ht]
    \centering
    \includegraphics[width=5.5in]{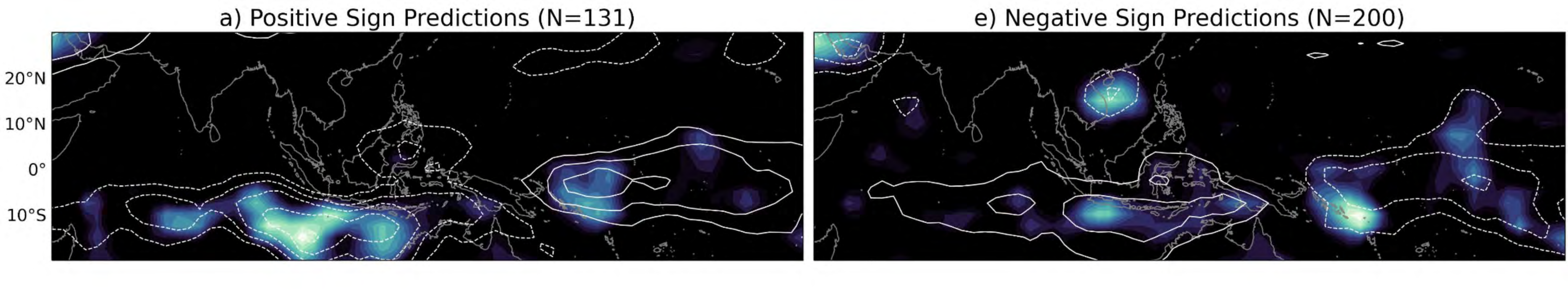}
    \caption{Interpreting a subseasonal forecast: Light areas show the geographical regions most relevant to the forecast. (from \cite[Figure~4]{10.1002/essoar.10505448.1}.) }
    \label{fig:sml1}
\end{figure}

The use of machine learning techniques in scientific computing has a long history dating back to the 1990s and earlier. 
The SIAM Conference on Data Mining, held every year since 2001, has always had a strong focus on science and engineering applications and on connections to high-performance computing \cite{SIDM01_url}.
 In January 2018, a DOE ASCR workshop and report \cite{Baker2019}
identified six Priority Research Directions for Scientific Machine Learning (SciML) that highlight
basic research challenges such as: 
\begin{enumerate}
    \item Science and engineering applications often make use of detailed models, based on laws of physics, chemistry, and biology, that enable detailed simulations to be performed and useful predictions to be made. The ``model-free'' ethos that pervades machine learning---the idea of ``letting the data speak for itself''---is not naturally compatible with the use of physical models. However, there is increasing interest in making current machine learning techniques ``play well'' with physical models---augmenting, enhancing, and complementing physical models in ways that potentially reduce computational requirements while maintaining adequate fidelity to scientific laws.
    \item Even in their most successful applications, including speech and image recognition, machine learning models are susceptible to perturbations in input data and parameters. That is, their predictions can be affected strongly by minute changes to the data. {\em Robustness} of these models reduces such sensitivity and is essential to scientific applications where model outputs that are obviously invalid would reduce the credibility of machine learning methodology.
    \item It is particularly important in scientific applications for models to be {\em interpretable}---for their simulations and predictions to accord with prior knowledge (see, e.g., Figure~\ref{fig:sml1}). Since the applications can be mission-critical, {\em trustworthiness} is another essential property; the outputs and predictions must be reliable. An example of the latter phenomenon is that when a machine learning model is presented with data that is outside the scope of its training data, it is able to flag that data as being ``out of distribution'' and issue a warning that the predictions may not be trustworthy.
\end{enumerate}

Machine learning enhancements to physical and biological models can be  useful in  ``plugging gaps'' in existing composite models, using data-driven machine learning models in those parts of the system for which the physics is not adequately known. But even in cases where the physics {\em is} known, machine learning  can still play a role in surrogate models that can be executed more cheaply as part of  optimization, control, and inversion processes. Such surrogates can be trained using data generated by high-resolution physical models---an expensive process, but one that can be done ``offline'' and in a way that makes use of massively parallel computing.
 Surrogates of this type already have been used  in  fields such as Earth science %
 \cite{Bauer2021}.
 Better understanding is needed at an abstract level of how physical and machine learning model components can be composed in ways that are efficient and serve the uses of the overall model. Randomness plays an important role in generating data to train machine learning surrogates from physical models, accounting for the uses to which the overall model will be put (for example, optimization, control, inversion). Potential benefits of this improved machine-learning-enhanced modeling methodology are vast and include accelerated scientific discovery in  transistor design, materials discovery, and aerospace engineering.

In scientific applications,  the outputs of machine learning models must be valid and trustworthy, with quantified sensitivity and uncertainty and with interpretable behavior. 
The scientific computing community includes generations of computational scientists with wide and deep experience in modeling important processes. 
Machine learning models whose outputs conflict with this experience are unlikely to be trusted by these scientists.
Techniques for improving interpretability and quantifying uncertainty are active areas of research in the machine learning community, but the existing community of scientific computing people must be engaged in order to ensure that the results of this work are meeting their standards of quality. 
The challenges include high dimensionality, nonlinearity, and nonconvexity of the models, all of which make it difficult to sample the uncertainty in ways that are both theoretically and practically valid.  
Randomization can drive sampling strategies and help reduce the effective problem dimension. 
Techniques for exploring high-dimensional parameter spaces (based, for example, on Bayesian neural nets) are under investigation.

Robustness of the outputs of models to perturbations is essential in many mission-critical applications  (e.g., reactor control).
Machine learning models are known to be sensitive to perturbations in their inputs and to their learned parameters.
An area of active investigation for the past decade has been on improving the robustness of machine learning models to such perturbations.
Various approaches are being investigated, including dropout in neural network training, bagging, bootstrapping, and adversarial training.
Randomization can help by generating augmented training sets, and also in the form of stochastic differential equation-based analysis  leading to model outcomes that are distributional rather than point estimates.

Randomness already plays an essential role in machine learning (the optimization algorithms used to train neural nets incorporate randomness, for example). It plays an important role, too, in resolving the issues cited above, in ways that bring the benefits of modern advances in machine learning to scientific computing.
\begin{callout}
Randomness will be essential to the design of
the  next generation of machine learning models, facilitating robustness and reliability with  respect to perturbations in model parameters and data.
\end{callout}
Potential research directions include the design of stable network architectures~\cite{haber2017stable,erichson2021lipschitz}, novel  noise injection methods for training robust machine learning models, randomness as a resource to introduce implicit regularization, randomness  for data augmentation and robust  training~\cite{cohen2019certified}, and randomness as a strategy for  computing distributional estimates rather than just point  estimates~\cite{tran2017deep}. 
Such innovations are key to enabling deployment of machine learning models in mission-critical scientific applications.

Better understanding of the randomized optimization algorithms that are at the heart of scientific machine learning (and in fact all of machine learning) will be vital to future progress. 
The basic analysis of such algorithms makes assumptions that do not hold true in practical situations. 
Scientific machine learning requires solution of nonconvex, nonsmooth problems in which the randomized gradients do not satisfy an ``independent identically distributed'' (i.i.d.) property. 
Although progress has been made in understanding the algorithms under these conditions and  in  analyzing scaled stochastic gradient approaches such as ADAM, much foundational work remains to be done.
%

%

%

%

%

%

%

%

%

%
%

%

%

%

%
%
%
%
%
%
%
%
%

%

%
%
%
%
%
%
%
%
%
%
%
%
%
%
%

%

%
%
%
%
%
%
%

%

%
%
%
%
% --- End Inserted File ---

%
%
%
%
% --- End Inserted File ---

\clearpage
% ---- Inserted File ----
\section{Current and Future Research Directions}
\label{sec:research}

Having reviewed the wide range of application drivers for randomization in scientific computing, we  now cover the research directions
that must be pursued in order to enable randomization as a first-class tool and a driver of progress in scientific computing. One of the recurrent themes throughout this section is increased emphasis on practicality, whether it is sharpening complexity bounds by reducing the constants and other lower-order terms, determining precise sketch size or sampling rates based on application needs, or integrating randomized algorithms into coupled workflows.  

In many cases, the existing theory beyond popular randomization techniques, such as sketching or sampling, is too general and does not provide tight enough bounds for scientific computing tasks. By restricting the problem domain or structure, greater statistical and computational efficiencies can be gained, which will greatly broaden the applicability of randomized methods to scientific computing.  

Several subsections reiterate the importance of making the randomization itself computationally efficient in practice, beyond the big-$O$ notation. Since most scientific computing tasks are truly large scale and require massively parallel computing, emphasis is also placed on the coupling of randomization and parallelization. Moreover, the community suggests that these computational aspects of randomization be captured through software abstractions for wide adoption, availability, portability, and high performance.

% ---- Inserted File ----
\subsection{Opportunities for Random Sampling Schemes} 
\label{sec:sampling}
\sublead{K.~Myers}

As we see throughout this report, random sampling undergirds and enables many other types of randomized algorithms. From the Monte Carlo methods used to generate random samples in many forward models (\Cref{sec:forward-models}) to stratified sampling and graph sampling approaches for discrete processes (\Cref{sec:discrete-driver}) to the desire for data reduction or compression in the context of massive  data generators and quantum computers (\Cref{sec:facilities,sec:emerging-hw}), random sampling is the key component of many scientific computing advances.

Likewise, this report showcases several cutting-edge scientific areas that are faced with higher volumes or rates of data than ever before (see \cref{sec:facilities}). Experts need answers more quickly than is possible with the current state of the art. Random sampling offers the promise of tractable analysis ``downstream'' from data-generating mechanisms, whether they be exascale simulations, experimental data from high-throughput user facilities, or opportunistic measurements from sensors with ever-increasing data rates. At the same time, we must have assurance that the random sample will retain the relevant characteristics of the original data sets. Without this assurance, we cannot trust the conclusions. 

\begin{callout}
Random sampling offers the promise of tractable analysis ``downstream'' from data-generating mechanisms, whether they be exascale simulations, experimental data from high-throughput user facilities, or opportunistic measurements from sensors with ever-increasing data rates.
\end{callout}

Furthermore, we need sampling schemes that are themselves computationally tractable in the presence of large and/or streaming data. Ideally we want efficient sampling methodologies that ensure accuracy in the solution with minimal computational effort. An added challenge in the context of complex simulations is that we may need to control the computational cost of the sampling scheme before we know the available computational resources, which could change as the simulation evolves.

\begin{figure}[h]
  \centering
  \begin{tabular}{cc}
  \includegraphics[width=0.45\textwidth]{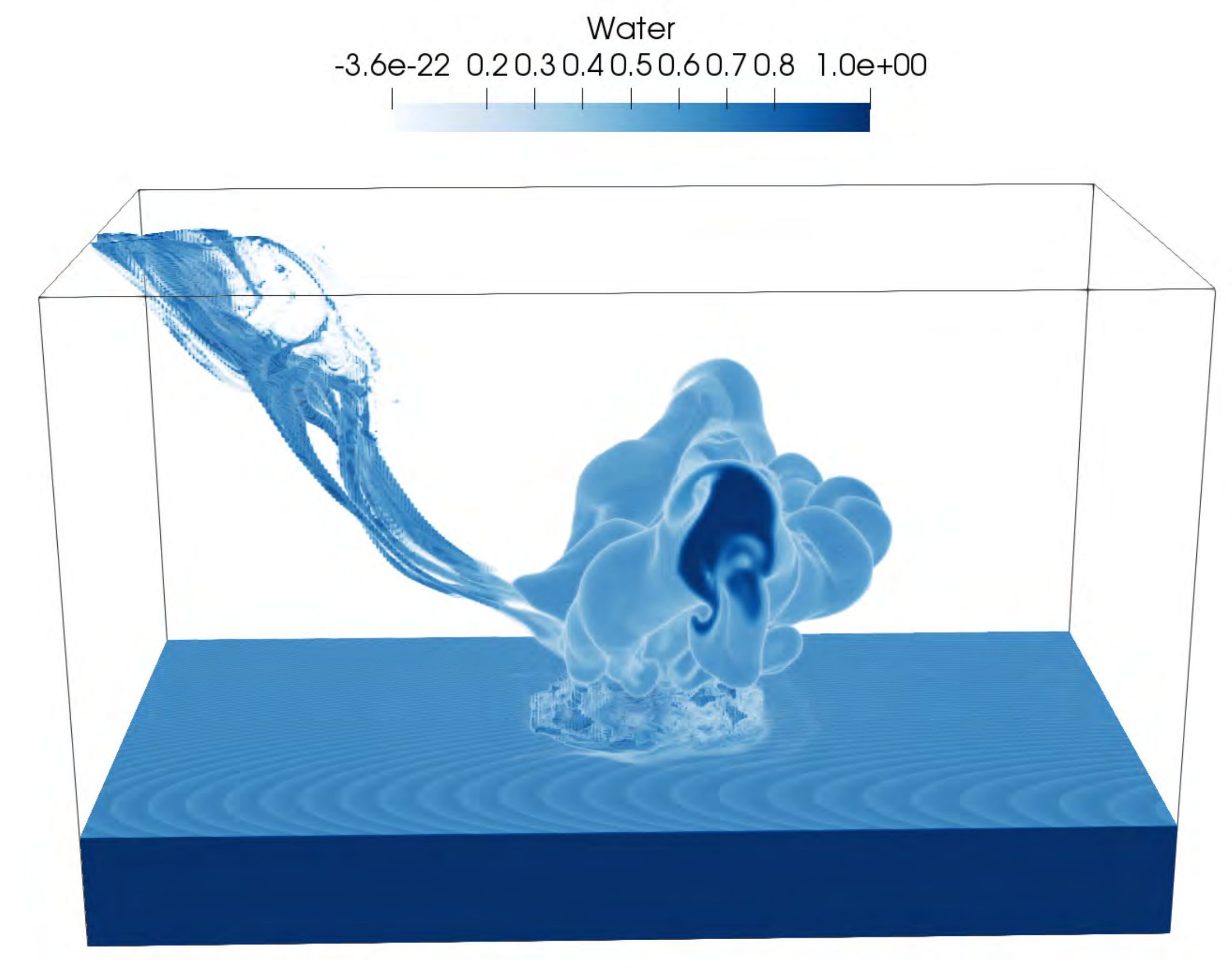} & \includegraphics[width=0.45\textwidth]{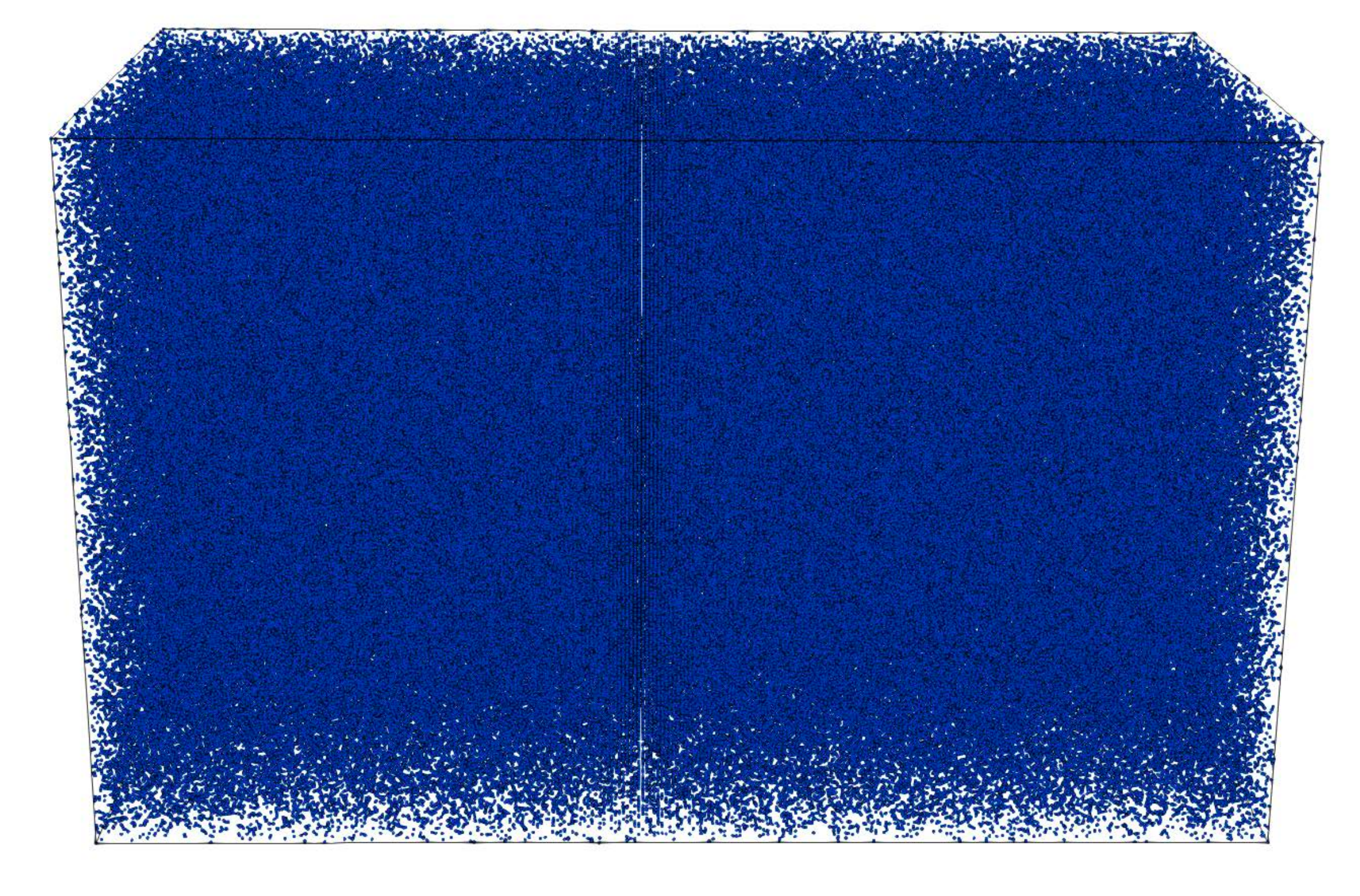} \\
  (a) & (b) \\
  \includegraphics[width=0.45\textwidth]{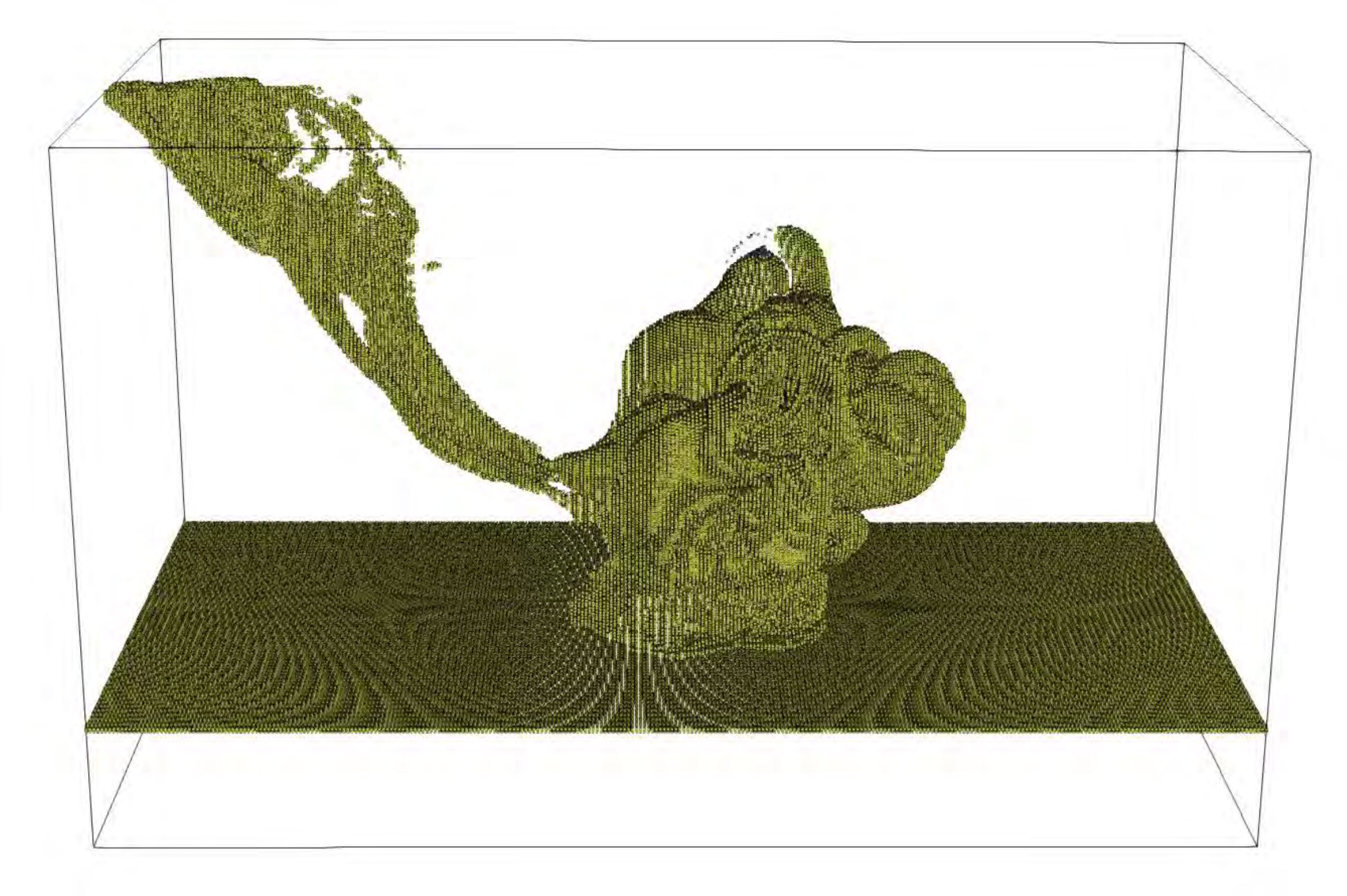} & \includegraphics[width=0.45\textwidth]{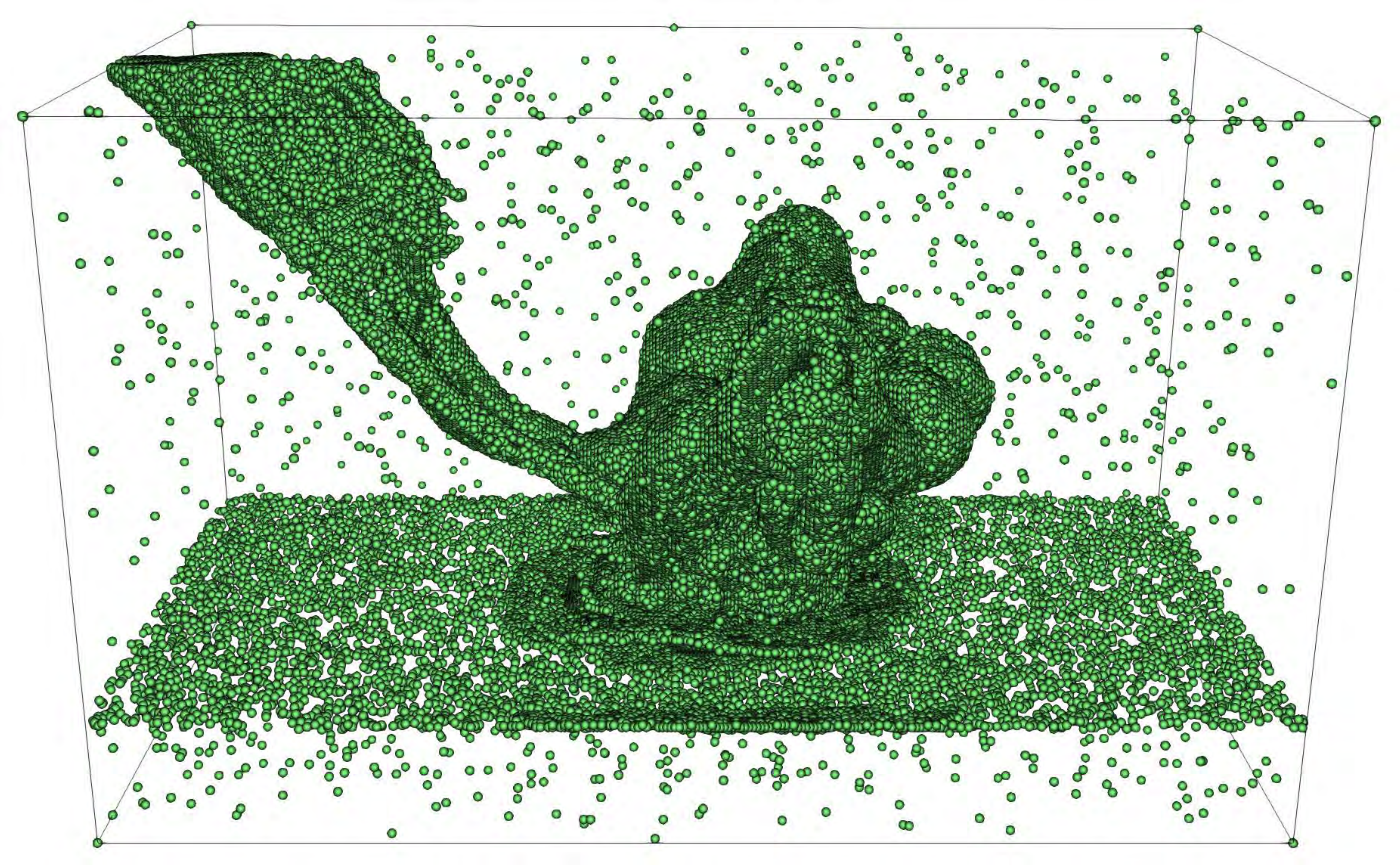}  \\
   (c) & (d)
 \end{tabular}
  \caption{  Sampling schemes applied to an ensemble of simulations run on a regular grid to explore asteroid-generated tsunami. (a)~Volume-rendered visualization of the simulation's water fraction variable, illustrating the plume of water generated after an
asteroid has hit the surface of the ocean. (b)~Simple random sampling scheme in which every simulation grid point has an equal probability of being included in the sample regardless of the underlying scientific content. (c)~Local-smoothness sampling.  (d)~Joint multicriteria sampling. Both (c) and (d) are data-driven sampling approaches that reveal the features of scientific interest. All three sampling schemes use a sampling ratio of 2\%. The color schemes in each panel are used to indicate that each panel presents a different sampling method. (adapted from \cite{biswasSampling2020}).}
  \label{fig:asteroid_sampling}
\end{figure}

To illustrate some of these concepts in the context of a scientific data set, 
\cref{fig:asteroid_sampling} presents sampling schemes explored and developed under the ECP \cite{ECPST2020}. Here the focus was the Deep Water Impact Ensemble data set \cite{PatchettAhrens2018}, a set of simulations produced on a regular grid and used to study asteroid-generated tsunamis.  Panel (a) shows a volume-rendered visualization of the simulation's water fraction variable, showing the plume of water generated after an
asteroid has hit the surface of the ocean.

Suppose our storage budget  allows us to save only 2\% of the grid points of the simulation. We need to generate a sample of the simulation that will support post hoc analysis. Panel (b) of Figure~\ref{fig:asteroid_sampling} shows an example of a simple random sampling scheme, where each simulation grid point has the same probability of being included in the sample regardless of what underlying scientific content may be conveyed by that grid point. This is easy to compute but yields a uniformly distributed collection of points that loses the scientific features of interest. In contrast, panels (c) and (d) show two different data-driven schemes for selecting 2\% of the original grid points that were explicitly designed to capture salient features of a specific scientific data set in a computationally tractable framework \cite{biswasSampling2020}.

More developments are needed in these directions, especially as more data-intensive applications come online and data rates continue to accelerate. Here we discuss several promising research directions.

\FloatBarrier

\paragraph{Computationally efficient sampling} 
Efficient sampling of data offers many important research opportunities, particularly in the context of massive and/or streaming data sets. These include how to adapt to the geometry of data, achieve robust nonlinear dimension reduction, identify sparse representations (e.g., intrinsic low-dimensional structures), filter noise, and identify anomalies.
An exciting research direction along these lines is adaptive sampling. In the context of user facilities and other experiments, this methodscan address the question of where to sample next in order to gain the most information, allowing scientists to find the ``needle in the haystack'' under highly dynamic conditions. In optimization, some adaptive sampling approaches use varying sample sizes to gradually reduce the variance in the stochastic gradient. These enable optimal balance between the computational burden and the accuracy of approximated information.

\paragraph{Stratified and topologically aware sampling} 
A concern is that naive sampling techniques can miss small but important subsets of data.
  Stratified sampling---strategically partitioning data into classes to which we assign a sampling distribution---can address this.
  However, many partitioning techniques assume that data points that are geometrically close to one another are similar, a situation that is not always true.
  Topological data analysis tools allow us to construct graphs from data where relationships are informed by more than distance.
  
\paragraph{Scientifically informed sampling} For assuring that the sampled data set retains the salient characteristics of the original data set, an interesting challenge is that the salient characteristics could differ depending on the scientific questions of interest. 
For instance, if the interest is in recognizing the occurrence of rare events in a massive data set, a Monte Carlo sampling scheme might produce some samples with no instances of that event, leading to an underestimate of their occurrence, and other samples with one or more instances, leading to an overestimate. 
On average the Monte Carlo samples will have the correct proportion, but any given sample could be far from the truth. 
In that situation, importance sampling \cite{owen2000safe} may be more appropriate because of the particular interest in rare events. 
This sort of situational responsiveness demands the development of sampling methods that are informed by the scientists. 

\paragraph{Reproducibility} The idea that different samples could have different characteristics even when generated from the same sampling scheme leads to important questions about reproducibility. Will the scientific results be different if a different set of samples is used? Scientists are unlikely to use analysis algorithms that give widely different answers for different random samples. To address this concern and to evaluate whether data samples are useful to the scientist could require research in information theory or theoretical computer science. Success here would lead to greater acceptance of randomized algorithms,
more confidence in science results, and
fewer false discoveries.

%
%

%

%

%
%
%
%

%

%

%
%

%

%

%
%
%
%
% --- End Inserted File ---
% ---- Inserted File ----
\clearpage
\subsection{Sketching for Linear and Nonlinear Problems}
\label{sec:sketching}
\subleads{T.~Kolda \& P.G.~Martinsson}

Sketching is a mathematical technique wherein a large problem or data set is replaced by a much smaller ``sketch" that retains essential properties. 
Counterintuitively, the size of the sketch can be
\emph{independent} of the size of the original problem,
meaning that the cost savings can be better than exponential.
Linear sketching has been applied successfully 
in many scenarios, including regression (\cref{fig:rand-least-squares}) and low-rank factorization
\cite{Halko2011,Mahoney11,Woodruff14,martinsson_tropp_2020}.
As a relatively new technique, the most convincing DOE applications of sketching have been
in numerical linear algebra.
More broadly speaking, sketching is widely used in industry for counting unique elements, estimating quantiles, or detecting frequent items in massive data sets or data streams (e.g., \cite{datasketches}).
Looking forward, sketching promises to be a key tool in areas including
solution of large-scale nonlinear inverse problems, PDE-constrained optimization, solution of linear and semidefinite programmer solvers, and quantum chemistry.

\begin{callout}
  Counterintuitively, the size of the sketch can be
\emph{independent} of the size of the original problem,
meaning that the cost savings can be better than exponential.
\end{callout}

\subsubsection{State of the Art}

\begin{figure}
  \centering
  \includegraphics[width=4in]{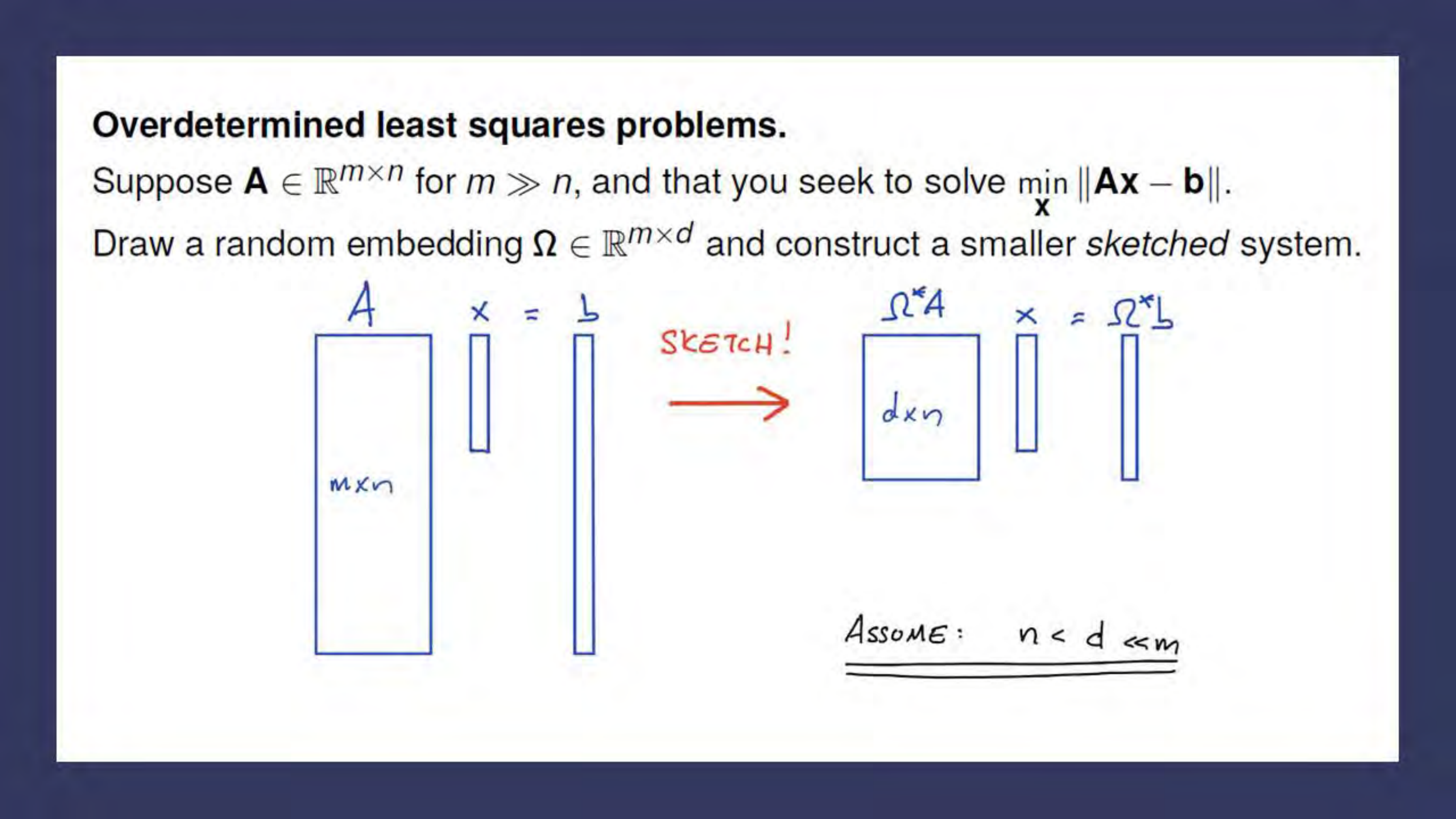}  
  \caption{Matrix sketching in the context of regression, using a random embedding to create a smaller problem. (image from
  the \href{http://users.oden.utexas.edu/~pgm/Talks/2020_12_DOE.pdf}{RASC workshop presentation} of Per-Gunnar Martinsson, December 2, 2020.)}
  \label{fig:rand-least-squares}
\end{figure}

A prototypical problem is linear regression, fitting an $n$-dimensional linear model to a set of $m$ observations
where the number of observations is orders of magnitude larger than the number of dimensions ($m \gg n$).
We let $\mathbf{A} \in \mathbb{R}^{m \times n}$ denote the matrix of $m$ observations, $\mathbf{b}$ be the
corresponding right-hand side and $\mathbf{x}$ be the solution. Since the problem is
overdetermined, we cannot in general find a solution that solves the problem exactly. Instead, we seek a \emph{least-squares} solution
that yields the  minimum square error, namely,  $\min_{\mathbf{x}} \| \mathbf{Ax} - \mathbf{b} \|^2$.
When the coefficient matrix $\mathbf{A}$ is large, the problem of finding the minimizer can be accelerated
by forming a much smaller ``sketch'' of the full system,
produced via a sketching matrix $\mathbf{\Omega} \in \mathbb{R}^{m \times d}$, so that the sketched system 
$\min_{\mathbf{x}} \| \mathbf{\Omega^T Ax} - \mathbf{\Omega^T b} \|^2$ has only $d \ll m$ rows and is much more efficient to solve (\cref{fig:rand-least-squares}).

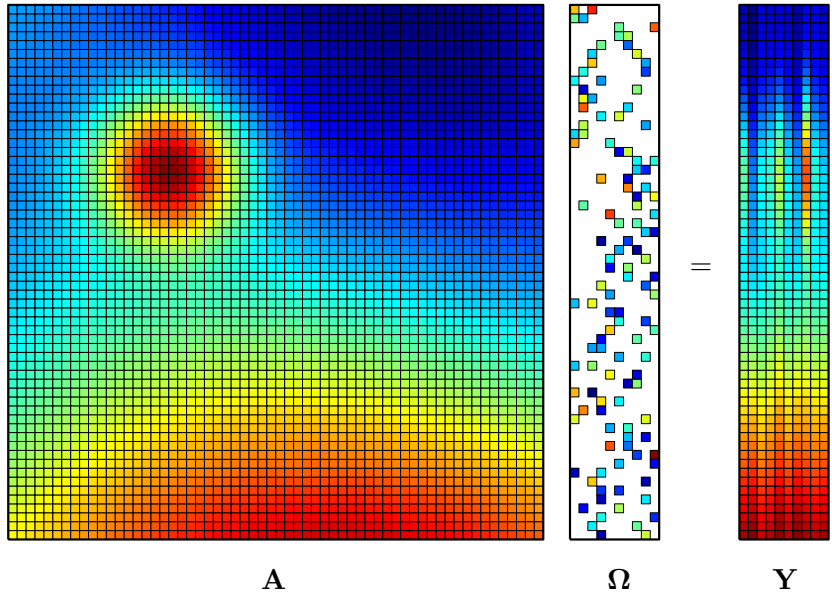
\begin{figure}
     \centering
     \setlength{\unitlength}{1mm}
     \begin{picture}(110,80)
     \put(000,06){\includegraphics[width=110mm]{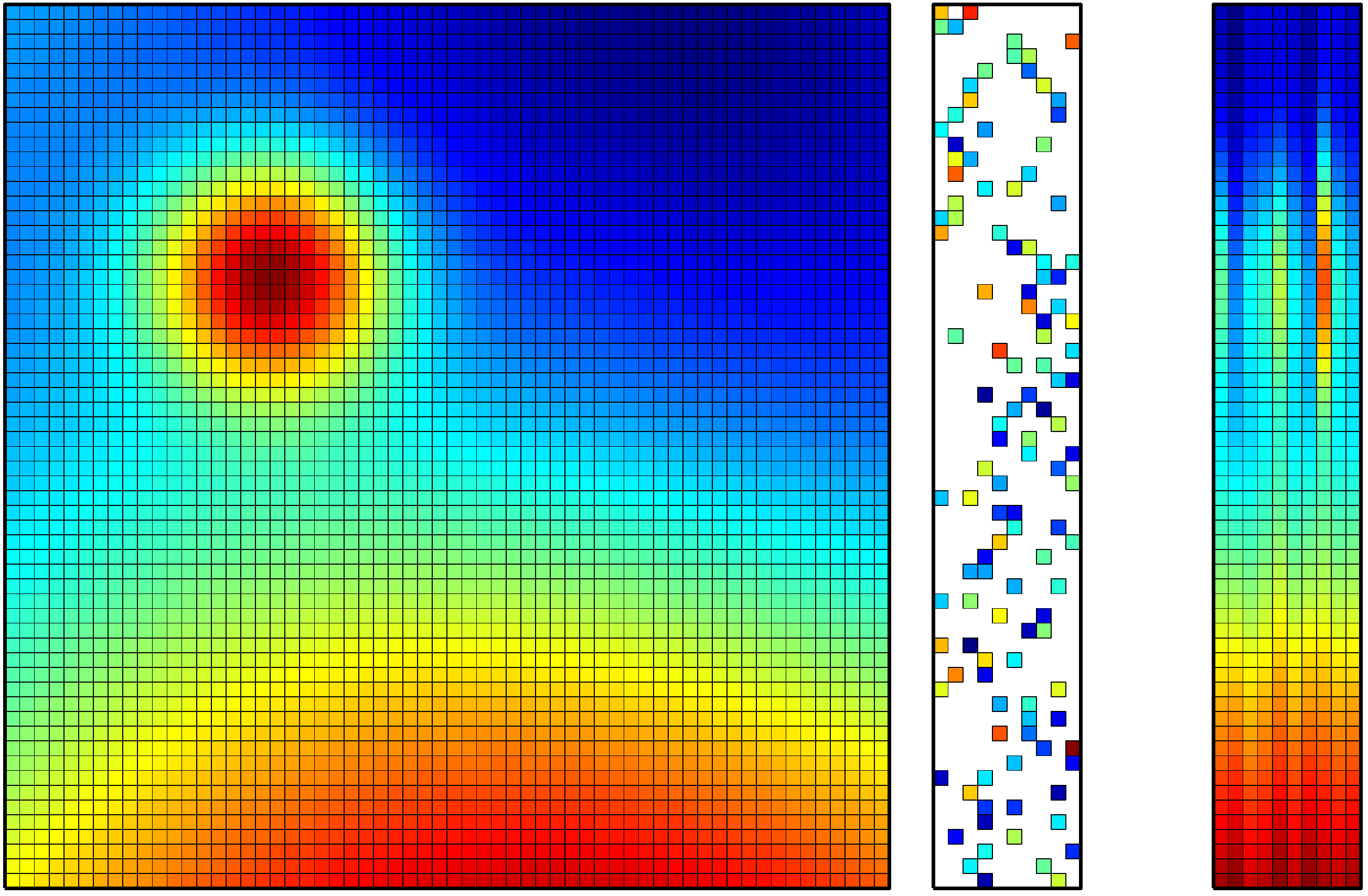}}
     \put(034,00){$\mathbf{A}$}
     \put(080,00){$\mathbf{\Omega}$}
     \put(102,00){$\mathbf{Y}$}
     \put(091,42){$=$}
     \end{picture}
     \caption{Randomized sketching in linear algebra: Given a matrix $\mathbf{A}$, a compressed sketch $\mathbf{Y}$ is formed by applying $\mathbf{A}$ to a tall, thin random matrix $\mathbf{\Omega}$. When $\mathbf{\Omega}$ is drawn from a ``good'' distribution, the sample matrix $\mathbf{Y}$ contains all the information required to compute an approximate basis for the column space, to find a set of rows of the matrix that approximately spans its row space, and to accomplish many other tasks.}
     \label{fig:randomsketch}
 \end{figure}

Many ways can be used to create the linear sketch $\mathbf{\Omega}$.
A common approach is to choose $\mathbf{\Omega}$ to have random entries drawn from a
standard normal distribution. Such Gaussian sketches are fast and reliable and
yield the smallest possible sketch:  the size of $d$ is as small as theoretically possible.
For the price of a slightly larger sketch, further acceleration is possible by using random maps that are sparse (\cref{fig:randomsketch})
or have other internal structure that enables their application using highly
efficient algorithms such as fast Fourier transforms~\cite{10.1145/3019134,Halko2011}.
For instance, fast Johnson--Lindenstrass transforms~\cite{ailon2009fast,liberty2007randomized}  employ this strategy.
Specialized sketches have been developed for the case that $\mathbf{A}$ is sparse,
using specialized sampling strategies, such as leverage-score sampling, that interact with only a subset of the data~\cite{10.1145/3019134,Mahoney11}.

In some environments, the minimizer of the sketched system can serve as a good approximation to the
minimizer of the original problem, referred to as the ``sketch-to-solve'' regime.
Using the solution to the sketched system directly
can lead to dramatic acceleration, but the error can be bounded only when the
properties of the original system are a priori well understood.
Alternatively, the sketched system can be used as a preconditioner
that ensures rapid convergence in an iterative solver
for the original problem.
Such a ``sketch-to-precondition'' approach has proven to be powerful in accelerating practical
computations and has a particular advantage in that this solver is 100\% reliable, since the computed solution is \textit{guaranteed} to fit the data well
\cite{rokhlin2008fast,avron2010blendenpik}.

Another successful application of matrix sketching concerns
low-rank approximations of matrices. The idea is to use linear sketches to 
compute approximate bases for the row and/or column spaces, (cf.~\cref{fig:randomsketch}). 
Once these have
been constructed, all further computations can be executed on the small sketches.
Algorithms of this type have proven to be highly communication and storage
efficient and excel in severely communication-constrained environments such
as GPU computing or when  data is stored out of core \cite{woolfe2008fast,martinsson_tropp_2020}. 

Other recent examples of sketching include
a two-stage Gauss--Seidel preconditioner with a randomized and asynchronous version of the Gauss--Seidel preconditioner developed by Avron et al.\@  using a graph Laplacian problem as a probe,
randomized pivot selection in QR \cite{duersch2017randomized,duersch2020randomized,martinsson2017householder},
accelerated tensor decomposition \cite{sun2020low,BaBaKo18},
and even very recent and potentially groundbreaking work in semidefinite program solvers \cite{YuTrFeUd21}.

Training of large-scale machine learning methods has led to many new and  popular randomized methods for optimization, such as
AdaGrad \cite{duchi2011adaptive} (with approximately 9,000 citations per Google Scholar as of this writing)
and ADAM \cite{KiBa15} (with over 60,000 citations).
Despite its popularity, the convergence of ADAM is not yet well understood. 

\subsubsection{Future Challenges}

Many open research problems remain. Here we mention a few exemplars.
In each case the problem requires domain expertise to understand the accuracy requirements and special structure.
Theoretical and statistical analyses are needed to, for instance, determine the required size to obtain the required accuracy
or to develop appropriate sampling strategies.
Implementations need to be adapted or rewritten, especially to realize reduced communication costs.

\paragraph{Incorporating sketching for solving subproblems}
Most large-scale simulations solve a sequence of linear systems to find a solution, and these linear solvers
are the primary bottleneck.
Consider applications in
high-frequency electromagnetic scattering and strongly advective
advection-diffusion-reaction systems.
Sketching is a promising tool, but foundational questions need to be answered.
What is the size of the sketch that is required in order to guarantee the needed accuracy in the overall simulation?
Can smaller sketches be used in earlier iterations where less accuracy is needed?
Alternatively, consider a problem with multiple subsystems as in multiscale problems. Can
smart sketching yield improved approximate solutions at some scales? 
In most cases, the answers will be application specific and perhaps even problem specific.
In the context of ill-conditioned inverse problems,
the modes associated with small singular values are important because these are actually large when viewed from the perspective of the inverse.
Thus, some randomized algorithm ideas associated with ignoring small singular values are not directly applicable.
We do know, however, that many subblocks of the matrix inverse can often be approximated by low-rank operators.
Some hierarchical basis methods can already exploit this property, but further research into these types of algorithms should be expanded.
An interesting and challenging question is how  one can detect subblocks where it is appropriate to employ low-rank approximations via randomized algorithms.
This problem has been solved in some specific cases, but for more general matrices this is significantly less understood.

\paragraph{Specialized randomization for structured problems}
Greater efficiencies (i.e., in the form of smaller or easier-to-compute sketches) can be realized by
exploiting problem structure. For instance, can we exploit the dependency grid structure of PDE solvers
to come up with randomized variants of multilevel preconditioners? 
Can we exploit Kronecker structure in quantum structure calculations?
How does doing so impact efficiency and robustness?
This could potentially speed up the setup phase of an algebraic multigrid solver
considerably in the context of extreme parallelism.

\paragraph{Overcoming parallel computational bottlenecks with probabilistic estimates}
In parallel computing, the cost of floating-point operations is negligible  compared with communication costs.
Sketching can greatly reduce communication costs, and it should be feasible for certain applications
to ameliorate the loss in accuracy with  inexpensive extra iterations. 
As another application, global reduction operations are bottlenecks.
However, one may be able to distribute the data such that responses from only
a subset of the compute nodes are adequate to guarantee sufficient accuracy.

\paragraph{Randomized optimization for DOE applications}
In machine learning, stochastic optimization is standard practice.
Such methods compute an inexpensive stochastic gradient by using only partial information.
Can such methods be employed in the context of DOE applications based on large-scale simulations?
For instance, perhaps the stochastic gradient employs only a subset of the grid points.
This situation is not dissimilar to multigrid methods, except that those are deterministic and used only  in the context of linear solves. Can the computational burden of large-scale partial differential equation optimization be reduced while providing the same kinds of guarantees on accuracy or uncertainty quantification?
A potential advantage of randomized approaches is removing dependencies on outliers in data integration optimization tasks.
While randomized algorithms are  powerful, many popular ones (e.g., ADAM) are not well understood: they work, but it is not always clear why, how, or under what circumstances.
As computing resources and applications compel more use of such algorithms in high-performance scientific computing, it is vital that these algorithms are understood from a theoretical and quantitative standpoint.
Significant efforts in the theoretical foundations are necessary in order to characterize, measure, and understand those computational outputs, which are distributional in nature.

%
%
%
%
%

%

%
%

%

%
%
%
%
%
%
%
%
%
%
%
%
%
%
%
%
%
%
%
%
%
%
%
%
%
%
%
%
%

%
%
%
%
%
%
%
%
%
%
%
%
%
%
%
%
%
%
%
%
%

%
%
%
%
%
%
%
%
%
%
%
%
%
%
%
%
%
%
%
%
%
%
%
%
%
%
%
%
%
%
%

%
%
%
%
%

%
%
%
%
%
%
%
%
%

%
%
%
%
% --- End Inserted File ---
% ---- Inserted File ----
\subsection{Algorithms for Discrete and Combinatorial Problems} \label{sec:discretealgorithms}
\label{sec:discrete-research}
\sublead{A.~Bulu\c{c}}

We organize the major research themes in randomized algorithms for discrete problems in five distinct themes. The overarching goal of randomization here is finding scalable ways to sample, organize, search, or analyze very large data streams and discrete structures  on finite resource machinery. 
In this section, all connected structures such as graphs and their high-level counterparts (e.g., hypergraphs and simplicial complexes) are collectively referred to as ``networks.''

\paragraph{Randomized algorithms for discrete problems that cannot be modeled as networks}
Many important discrete problems cannot, or need not, be represented as graphs or their generalizations. For example, randomized techniques have been successfully used for routing in modern supercomputers~\cite{kim2008technology} and load balancing in various settings~\cite{mitzenmacher2001power}. As the concurrency increases to extreme scales, these methods will find more and more use in scientific computations. Another area where discrete non-graph problems arise is the analysis of sequencing data. For example, randomized algorithms are used to find compact index structures~\cite{ekim2020randomized}, which are crucial for efficiently comparing large sequencing (DNA, RNA, or protein) data sets. Exponential rise in sequencing data that has been outpacing Moore's law is a pressing reason to adopt these randomized algorithms widely in practice. 

\paragraph{Randomized algorithms for solving well-defined problems on networks} 

Randomized algorithms are used to provide approximate solutions to many subgraph counting problems with errors diminishing with sample size~\cite{jha2015path}. Various modifications to the celebrated color-coding technique of Alon et al.~\cite{alon1995color} have been used for this purpose. 
Some randomized graph algorithms are known for problems that require exact optimality, such as the min-cut~\cite{karger1996new} and minimum spanning tree~\cite{karger1995randomized} problems. 
However, several open issues impede the adoption of these clever techniques. While randomized algorithms often match or exceed the complexity bounds of the best deterministic problems in the worst case, they sometimes fail to match the performance of the best deterministic problems on real inputs in practice. 
A common reason is that existing randomized algorithms are designed to perform the same number of operations regardless of the input, making the common case as slow as the worst case. This situation is exemplified in Karger's algorithm for minimum cuts, whose ``primary misfortune is that it always runs in its worst-case $O(n^2 \lg{n})$ time bound''~\cite{chekuri1997experimental}.

Adoption of more realistic complexity measures by the community (\cref{sec:complexity}) will help close the gap between theory and practice. 
An obstacle to adoption of randomized methods is scalability to large concurrencies, especially on distributed-memory architectures where most big science computations are performed. 
New research demonstrating scalability of parallel randomized algorithms for important discrete problems will ignite the interest of domain scientists on randomized methods. 
Furthermore, the ability to run on a streaming setting is crucial for processing data coming from experimental facilities.

\begin{figure}[th]
  \centering
  \includegraphics[height=4cm,width=6cm]{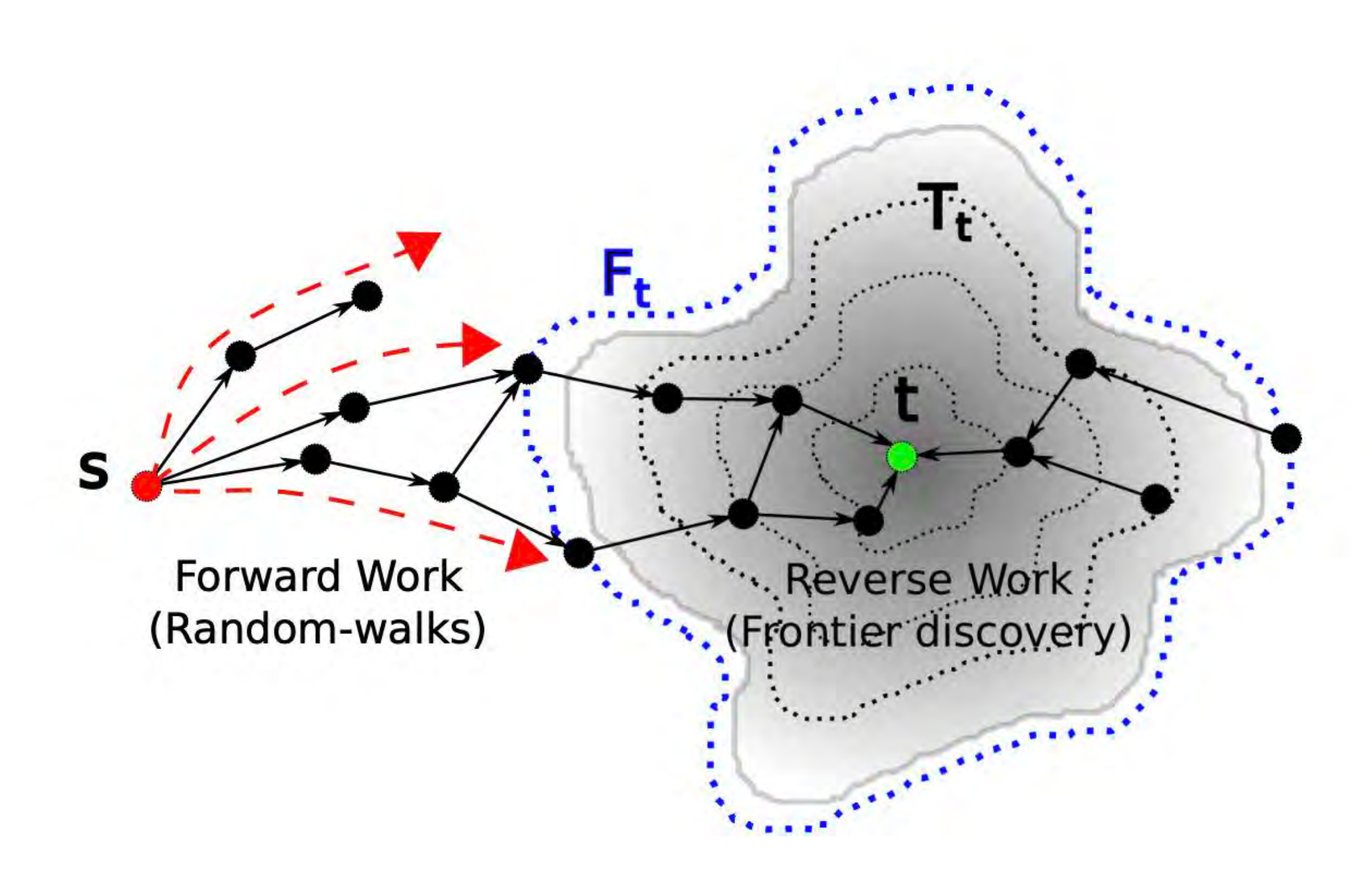}
  \caption{Example of randomization on graph traversal. The FAST-PPR algorithm, a fast personalized PageRank algorithm, uses careful random sampling to find relevant vertices in a massive network~\cite{LoBa+14}.}
    \label{fig:randomwalk}
\end{figure}

\paragraph{Universal sketching and sampling on discrete data} The technique of ``graph sketching'' refers to working on a subset of either nodes or edges from a much larger graph to draw a conclusion. Sketching can be achieved by using various forms of graph sampling methods. A popular graph sampling strategy is based on random walks, as shown in \cref{fig:randomwalk}. The theory of sketching and sampling, especially in the streaming setting, is often phrased in terms of specific algorithms that solve prespecified questions, such as frequent elements. In practice, the questions are often determined \emph{after} generating a sketch of the data or data stream. A theory of universal sketches, where streaming and data analysis algorithms can answer a large variety of questions, needs to be developed.

With the explosion of high-volume and high-velocity data, the extraction of information has become a serious challenge. 
Ample research opportunities exist for exploring, determining, and optimizing how randomization may prove fruitful in achieving scalability and high performance in network and topological data analysis.
Furthermore, the practicality of these algorithms is of paramount importance for their adoption in scientific computing. 
The theory of graph sampling, particularly as it concerns statistically nonstationary and time-dependent graphs, is rich with questions that have implications on the practical side of proposing randomized algorithms to search, sample, explore, and reduce networks.

Much of the existing literature provides bounds of the following form: Given accuracy and confidence parameters, necessary mathematical bounds exist on the sketch size and memory footprint. In practice, the situation is inverted. There is a fixed memory budget, and users need to get the ``best possible" answer. A research opportunity exists to develop bounds of the latter form. Moreover, existing theory focuses on asymptotic results ignoring constant factors. For practical applications of the theory we need a more precise theory that tackles the constant factors involved. The purview of challenges in streaming algorithms extends to edge computing as well as distributed computing.

\begin{callout}
Various forms of subgraph sampling are key to performing efficient training for graph representation learning. 
The methods  currently used  often lack generality, and their computational complexity is poorly understood.
\end{callout}

\paragraph{Randomized algorithms for combinatorial and discrete optimization}

A significant portion of problems in combinatorial optimization suffer from exponential (or worse) complexity using traditional methods~\cite{wolsey1998integer}. In practice, the situation is often exacerbated by the presence of nonlinear and nonconvex constraints, resulting in a lack of algorithmic scalability that even the most advanced high-performance computing platforms fail to overcome. Randomized algorithms can help overcome the scalability challenges in combinatorial optimization, especially if provable approximation guarantees are provided. A specific technique is randomized rounding for stylized combinatorial optimization problems. Randomization can also help solve PDE-constrained optimization problems arising, for example, in the control of additive manufacturing processes for a given control trajectory (combinatorial component).

\paragraph{Randomized algorithms for machine learning on networks}
Geometric deep learning~\cite{bronstein2017geometric} is often the umbrella term for various machine learning techniques on unstructured connected data, with the prime example being graph neural networks. Various forms of subgraph sampling are key to performing efficient training for graph representation learning~\cite{hamilton2017inductive}. 
The methods  currently used for this purpose often lack generality, and their computational complexity is poorly understood. 
Developing a robust theory of sampling graphs, hypergraphs, and other discrete structures is a key research direction. 
Furthermore, we need to understand how scientific goals relate to existing discrete and graph sampling techniques that are being employed and develop methods to quantify the effect of these sampling techniques on the scientific goal. 
Going beyond simple graphs that are characterized by pairwise interactions and generalizing these methods to higher-order structures~\cite{battiston2020networks} are another key research direction.

\FloatBarrier

%
%
%

%

%
%

%

%
%

%
%
%
%

%
%

%
%
%
%
%

%
%
%
%
%
%
%
%
%
%
%

%
%
%
%
%
%

%

%
%
%
%
%
%
%
%
%
%
%
%

%
%

%
%
%
%
%

%
%
%
%

%
%
%
%
%
%
%

%
 
%
%
%
%
%
%

%
%
%
%

%
%

%
%
%
%
%
%

%
%
%
%
%
%
%
%
%
%
%

%
%
%
%
%
%

%
%
%
%
%
%
%
%
%
%
%
%
%
%
%
%
%
%
%

%
%

%

%
%
%
%
% --- End Inserted File ---
% ---- Inserted File ----
\subsection{Streaming Algorithms and Data}
\label{sec:streaming}
\sublead{J.~Nelson}

A sketch of a data set $\mathcal D$ is simply a low-memory data structure to support answering any query from some given family of queries (see \cref{sec:sketching}). A primary goal is to achieve a sketch size that is sublinear in  $|\mathcal D|$, the size of $\mathcal D$. A {\it streaming algorithm} is simply a sketch that supports dynamic data; in other words,  the data structure should be able to process a stream of updates to $\mathcal D$, during which the sketch should be updated on the fly.

The earliest and perhaps simplest  streaming algorithm is the probabilistic counter of Morris \cite{Morris78} to count the number of events in a data stream using very few bits.
This can be useful for sensors or edge computing where there are extremely limited hardware or energy resources on device.
The Morris algorithm maintains a counter of up to $N$ values subject to a single operation: {\it increment} (\cref{fig:morris}). Whereas a counter that is exact must use $\Omega(\log N)$ bits of memory, Morris leverages randomization to develop his approximate counter (which reports the counter value $N$ up to approximately 1\% multiplicative error with at most 1\% failure probability) using only $O(\log\log N)$ bits of memory---an exponential improvement. Indeed, for many streaming problems both randomization and approximation are necessary in  order to obtain sublinear memory \cite{AlonMS99}.

\begin{figure}
  \centering
  \footnotesize  
  \renewcommand{\arraystretch}{2}
  \begin{tabularx}{\textwidth}{|rY|}
    \hline
    \bf Goal: & Count up to $N=100000$ (streaming) events using an 8-bit counter \\
    \bf Method: & Initialize $\eta=0$ to be the 8-bit value. Increment $\eta$ probabilistically according to the following procedure:\\[0mm]
    & \multicolumn{1}{c|}{
      \begin{minipage}{3in}
        \begin{itemize}
        \item Let $\xi$ be a uniform random value in $(0,1)$
        \item If $\xi<(a/(a+1))^\eta$, set $\eta = \eta +1$
        \end{itemize}
      \end{minipage}
    }\\[3mm]
    \bf Result: & $\hat n \equiv a((1+1/a)^{\eta}-1) \approx n$ with error $\sigma^2=n(n-1)/2a$ \\
    \hline
  \end{tabularx}
  \caption{Morris's Probabilistic Counter~\cite{Morris78}}
  \label{fig:morris}
\end{figure}

In early literature on streaming algorithms in the late 1970s to the mid-1990s, motivations ranged from wanting to study a crisp algorithmic model out of intellectual curiosity, without regard to practice, to  wanting to use low-memory data analytics in applications such as network traffic monitoring and databases \cite{Morris78,MunroP80,MisraG82,FlajoletM85,AlonMS99,AlonGMS02}. More recently, streaming algorithms have found their way into computational linear algebra \cite{Woodruff14}, machine learning \cite{GhaziPW19}, and state-of-the-art optimization algorithms for problems as fundamental as linear programming \cite{BrandLSS20}. 

A remarkable illustration of how randomization enables streaming algorithms has been the discovery of techniques
for computing an approximate low-rank factorization of a given matrix or tensor in a single pass over its entries \cite{martinsson_tropp_2020}. Traditional techniques for computing such a factorization, for example, Krylov methods or Gram--Schmidt orthogonalization, require multiple interactions with the matrix and cannot be deployed to matrices that are too large to be stored. In contrast, randomized sketches (as illustrated in \cref{fig:randomsketch}) of the row and column spaces of a matrix can be extracted in a single pass, and one can reconstruct the matrix using only the information contained in these sketches.
These new algorithms have the potential to dramatically enhance our ability to store and analyze gigantic data sets arising in applications such as turbulence modeling and molecular dynamics.

More specific to current DOE interests, with rapid increases in computing power, modern scientific simulations generate high-fidelity data that is outpacing our ability to write this data to disk for later analysis.  Similarly, rapid advancements in sensor technologies create storage and analysis bottlenecks for DOE scientific facilities.  Thus, a high-priority research area for the DOE Advanced Scientific Computing Research program is the development of robust, efficient, and scalable algorithms for dimensionality reduction and/or data compression of streaming scientific data. Challenges that must be overcome include accurately quantifying uncertainties in such representations arising from both the data and stochastic approximations inherent in randomized algorithms, making the algorithms robust to hyperparameter tuning to ensure their effectiveness for real DOE scientific problems, making the algorithms efficient enough in terms of computational complexity and software implementation for in situ and/or online application, and porting the algorithms to emerging computing architectures that emphasize high-bandwidth streaming computations over random accesses that are common in many randomized algorithms.  If successful, such techniques would dramatically increase the throughput of DOE's simulation and data acquisition workflows, thereby quickening the pace of scientific breakthroughs.

Also of interest is the analysis of large structured data such as graphs or matrices, which are a key component of data science workflows that becomes expensive in distributed memory, usually because of memory and especially communication overhead. \begin{callout}
Randomized streaming and especially sketching algorithms can approximately summarize, e.g., vertex neighborhoods \cite{McGregor16} or matrix rows \cite{ClarksonW09} in small memory for subsequent communication and analysis on other compute nodes or client hardware---affording approximation of quantities of interest with reduced latency by dramatically limiting communication overhead.
\end{callout}
High-performance computing algorithmic pipelines can use sketches to perform numerical linear algebra, local structure approximations in graphs, dimensionality reduction, and nearest-neighbor computations, among other key data science tasks. Such approximations and latency improvements, along with performant and user-friendly software, are necessary in order to make high-performance computing resources accessible and useful to nonexpert data scientists across many scientific domains of interest.

\paragraph{Mergeable summaries} One useful technique for distributed processing of streaming data is that of using {\it mergeable summaries} \cite{AgarwalCHPWY13}. A  fully mergeable streaming algorithm is one that allows several different streams to be processed separately so that the resulting sketches can be merged in an arbitrary merge tree with no degradation in accuracy or confidence to obtain a sketch for the union of all datasets. Such algorithms are important for minimizing communication (only sketches need to be communicated) when data is naturally distributed across a network. They are  useful even when data is not distributed because they allow for a divide-and-conquer approach to obtain parallel algorithms.
\paragraph{In situ and real-time data analysis} As discussed in \cref{sec:facilities}, in situ and real-time data analysis are necessary in order to keep pace with increasing rate, size, and complexity of streaming data in national science user facilities. Streaming algorithms are also needed to train or update models in an online fashion as more data is seen.

\paragraph{Privacy} In several applications, data is streamed in from multiple sources that would like to maintain privacy against the central server processing the data. For example, consider Apple wanting to automatically learn words for its spellchecker dictionary by monitoring words typed by iPhone users, while guaranteeing user privacy so that Apple itself cannot determine which users typed which texts \cite{ThakurtaVV+17}. A formal definition of privacy preservation is given by {\it differential privacy} \cite{DworkMNS06} in which a database is preprocessed into a randomized output that is statistically nearly indistinguishable from what would be output had any single user's data been removed. In traditional differential privacy this data randomization is performed by a trusted central server, but the example given here shows the necessity of development of solutions in the so-called  local model, where data is distributed and the central processing server is untrusted. Several recent works have given solutions to specific tasks in this local model, as well as a newer  shuffle model \cite{CheuSUZZ19}, but this direction is still in an early stage of development.

%
%
%
%
%
%
%
%

%
%
%
%
% --- End Inserted File ---
% ---- Inserted File ----
\subsection{Complexity Analysis} 
\label{sec:complexity}
\sublead{M.~Anitescu}

\begin{figure}
  \centering
  \includegraphics[width=4in]{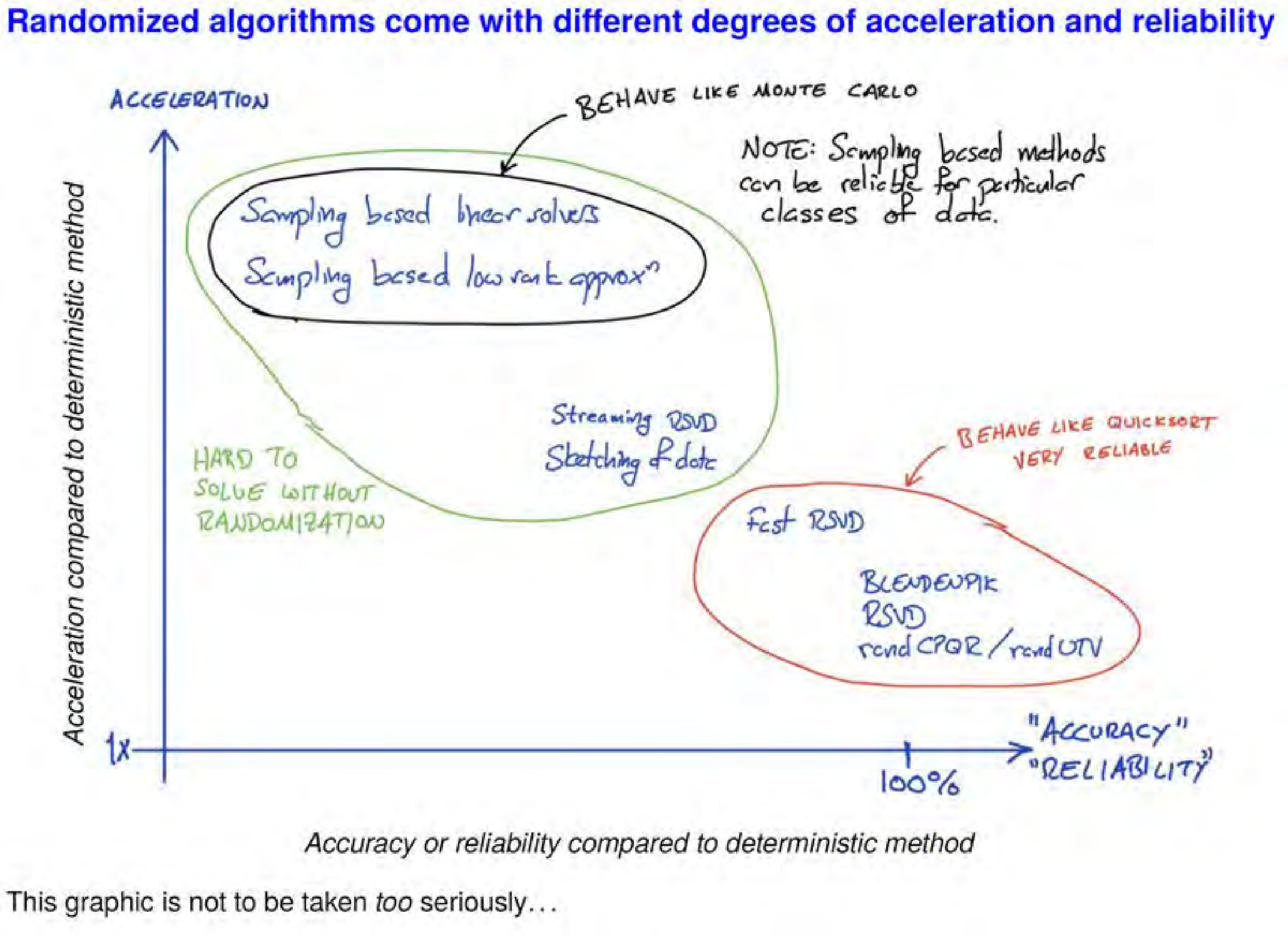}  
  \caption{Very rough sketch of the landscape of accuracy and acceleration for existing randomized algorithms for linear algebraic problems.
    Some algorithms can be 100\% accurate and essentially risk-free, even improving robustness.
    For massive problems, small sacrifices in accuracy may yield orders of magnitude acceleration.
    (image from
  the \href{http://users.oden.utexas.edu/~pgm/Talks/2020_12_DOE.pdf}{RASC workshop presentation} of Per-Gunnar Martinsson, December 2, 2020.) }
  \label{fig:acceleration-vs-reliability}
\end{figure}

One of the main drivers of this document is the strikingly reduced complexity (e.g., dependent primarily on the rank rather than the dimension, in the case of singular value decomposition) that randomized algorithms achieve in some notable circumstances. Nevertheless, many advances are still required in order to understand when and how to use such algorithms and to sharply quantify their performance and limitations. For instance, even for the fairly basic cases of linear systems and matrix approximations, a spectrum of randomized algorithms exists, some that are very accurate but only 2--10 times faster than deterministic methods and others that are less accurate but many orders of magnitude faster than deterministic methods.  As a  rough sketch,  \cref{fig:acceleration-vs-reliability} shows that the boundaries of the respective domains are not sharply understood in practice. A persistent challenge in randomized algorithms is identifying the constants in the big-$O$ promises of theory for randomized algorithms and thus obtaining a sharp characterization of the problem size at which randomized approaches start to be competitive with, or better than,  deterministic ones. 

Alternatively, understanding the boundary of such regimes sometimes offers the opportunity for hybridizing discrete and randomized algorithms to obtain an even better complexity/accuracy boundary. An example of the ``best of both worlds" is provided by randomized quasi--Monte Carlo approaches \cite{l2016randomized}. Such approaches are deterministic approaches to high-dimensional integration that reduce integration error from Monte Carlo's $O(n^{-1/2})$ to "almost" $O(n^{-1})$. Two problems with quasi--Monte Carlo exist:  it is deterministic, with no computable a posteriori error bounds, and the ``almost'' qualification  about error reduction hides worst-case powers of log(n) that are not negligible. The randomized version of quasi--Monte Carlo supports error estimation by replication, has finite sample variance no worse than a constant multiple of Monte Carlo's, and effectively circumvents the worst case. Moreover, for some sufficiently smooth integrands, the error is much better than either Monte Carlo's or quasi--Monte Carlo's. Composing or combining deterministic and randomized methods and understanding the properties of the resulting algorithms would bring about both novel mathematics and improved capabilities for DOE's applications.

In addition to such general challenges and opportunities concerning the complexity of randomized algorithms, this workshop identified two notable priority research directions related to the complexity of randomized algorithms.

\subsubsection{Analysis of Randomized Algorithms for Production Conditions}
The promise of reduced complexity of randomized algorithms is an exceptional opportunity to advance the state of the art in mathematics and computer science while addressing critical questions facing the applications in the space of  DOE. Randomization has recently been demonstrated to vastly improve both the theoretical and practical complexity of ubiquitous computational kernels, 
and it is a key enabler for approaching complex tasks that are deterministically intractable. 

\begin{callout}
Randomization has recently been demonstrated to vastly improve both the theoretical and practical complexity of ubiquitous computational kernels, and it is a key enabler for approaching complex tasks that are deterministically intractable.
\end{callout}

Achieving such lofty goals faces a set of challenges, such as quantifying and predicting performance of randomized algorithms under realistic production conditions. Moreover, we need to develop analytical tools and frameworks for different classes of randomized methods for paradigms beyond linear algebra. Furthermore, accelerating scientific adoption requires the ability to crisply communicate the algorithmic options and their properties to help users choose the best algorithm for any given task.

To these ends, the community will build on existing tools and results from theoretical computer science,
for example by developing rigorous and practical metrics to monitor convergence of  randomized algorithms. Novel efforts are needed not only to achieve sharper a priori complexity and run-time estimates but also practical a posteriori error estimates (which are typically significantly less conservative than a priori estimates \cite{martinsson_tropp_2020}) for realistic usage environments. A sustained effort is required in the analysis of tradeoffs between cost and accuracy, with a particular focus on characterizing environments and algorithmic templates where less accuracy is sufficient. In relation to hardware models, an interesting opportunity occurs in randomized techniques to avoid worst-case aggregation of errors (randomized rounding or truncation) and, more generally, in the integration of probabilistic error estimates with floating-point error estimates. Furthermore, since the optimal hardware world is likely to be hybrid, an important research direction is the analysis of coupled (hybrid) deterministic/randomized algorithms that combine the best of both worlds.

\subsubsection{Randomized Algorithms' Cost and Error Models for Emerging Hardware}

The performance of algorithms is affected not only by their mathematical formulation but also by the nature of the hardware system, which can exhibit radically different costs for different primitives.
A recent striking example concerns the vastly different computational speeds of integer programming in the classical versus quantum model. This can be seen in Shor's algorithm \cite{shor1994algorithms} that achieves polynomial time factorization of an integer in the quantum model, whereas the running times of the best-known deterministic algorithms is exponential (all statements with respect to the number of bits required to represent the number). In this regard, urgent needs have emerged caused by the ending of Moore's law and data federation in combination with novel hardware contexts that include an increased emphasis on streaming application and heterogeneous architectures. Important conceptual challenges include the following: How do algorithms need to change to mirror the hardware evolution? How does error analysis incorporate not only algorithmic errors due to randomization but also hardware originated errors?

\begin{callout}
The performance of algorithms is affected not only by their mathematical formulation but also by the nature of the hardware system, which can exhibit radically different costs for different primitives.
\end{callout}

Addressing such challenges requires progress on multiple research fronts at the intersection of mathematics and computer science. An important first step is a proper abstraction of the problem, which needs interaction between mathematicians, computer scientists, statisticians, and hardware architects to develop concise cost models for the underlying hardware to aid algorithm design. Such descriptions should lead to novel error models and estimators for randomized algorithms  for heterogeneous architectures. Moreover, such a focus would broaden the scope of algorithmic innovation and error analysis itself by creating opportunities for new optimized randomized algorithms for evolving hardware cost models that now include, for example, bandwidth and latency limitations.

Such a holistic approach to analysis and algorithmic design will not only increase confidence in randomized algorithms but will produce better randomized algorithm infrastructure for underlying software that benefits many applications; see  \Cref{sec:software}.

%

%
%
%
%
%
%
%

%
%
%
%
% --- End Inserted File ---
% ---- Inserted File ----
\subsection{Verification and Validation} 
\label{sec:vnv}
\sublead{J.~Jakeman}

Verification and validation are processes for checking the accuracy and reliability of algorithms or models. Broadly speaking, verification and validation involve the use of systematic tools to study how well computational results agree with known solutions or reference data.
 DOE has a strong history of supporting verification and validation research for {\it deterministic} physics-based computations; and  verification and validation standards are well established for many science drivers~\cite{DOD_V&V_report, AIAA_V&V, ASME_V&V}. 
In the context of \emph{randomized algorithms}, however, the processes for verification and validation have not yet reached the same level of development. Accordingly, future efforts in verification and validation are needed to ensure that new technologies based on randomized algorithms can be used safely and with high confidence.

In order to develop tools and systems for verification and validation  of randomized algorithms, challenges and opportunities need to be addressed in several directions.

\paragraph{Going beyond worst-case error analysis} 
Traditionally, error analysis of deterministic algorithms is studied from a \emph{worst-case perspective}. In other words, the goal of this type of analysis is to measure the largest error that may arise among all possible inputs.
A common limitation of this approach is that the error bounds tend to be overly conservative for typical inputs, and consequently they may not provide a realistic guidance about an algorithm's performance in practice. As a way of overcoming this issue, randomized algorithms naturally lend themselves to other types of error analysis. In particular, randomized algorithms are suited to probabilistic analyses that are flexible enough to handle average-case error, as well as a posteriori error estimates to be discussed below. 
For instance, statements of the form ``the error is no worse than $\epsilon$ with probability exceeding $(1{-}\delta)$" are 
common in theory for randomized algorithms, and such \emph{probabilistic bounds} could help in going beyond worst-case analyses for verification and validation.

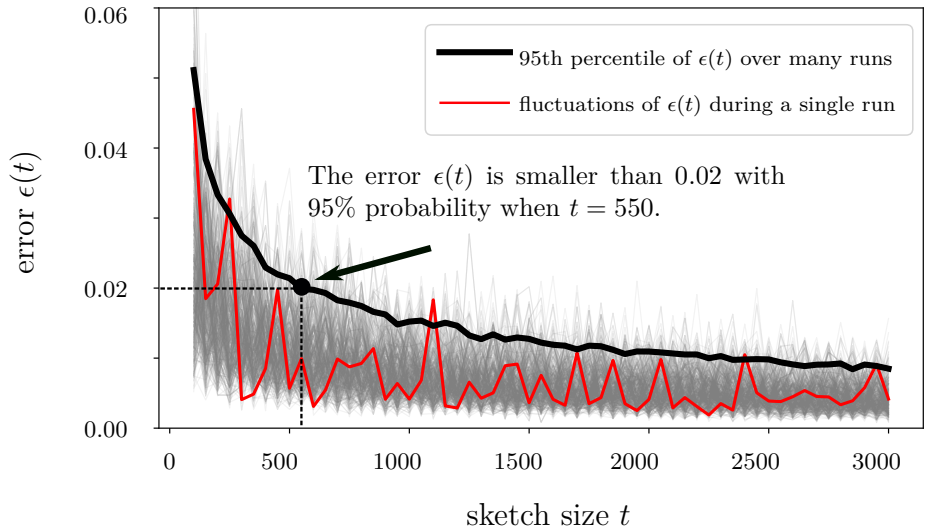
\begin{figure}[htb]
	\centering
	\DeclareGraphicsExtensions{.pdf}
	\begin{overpic}[width=0.75\textwidth]{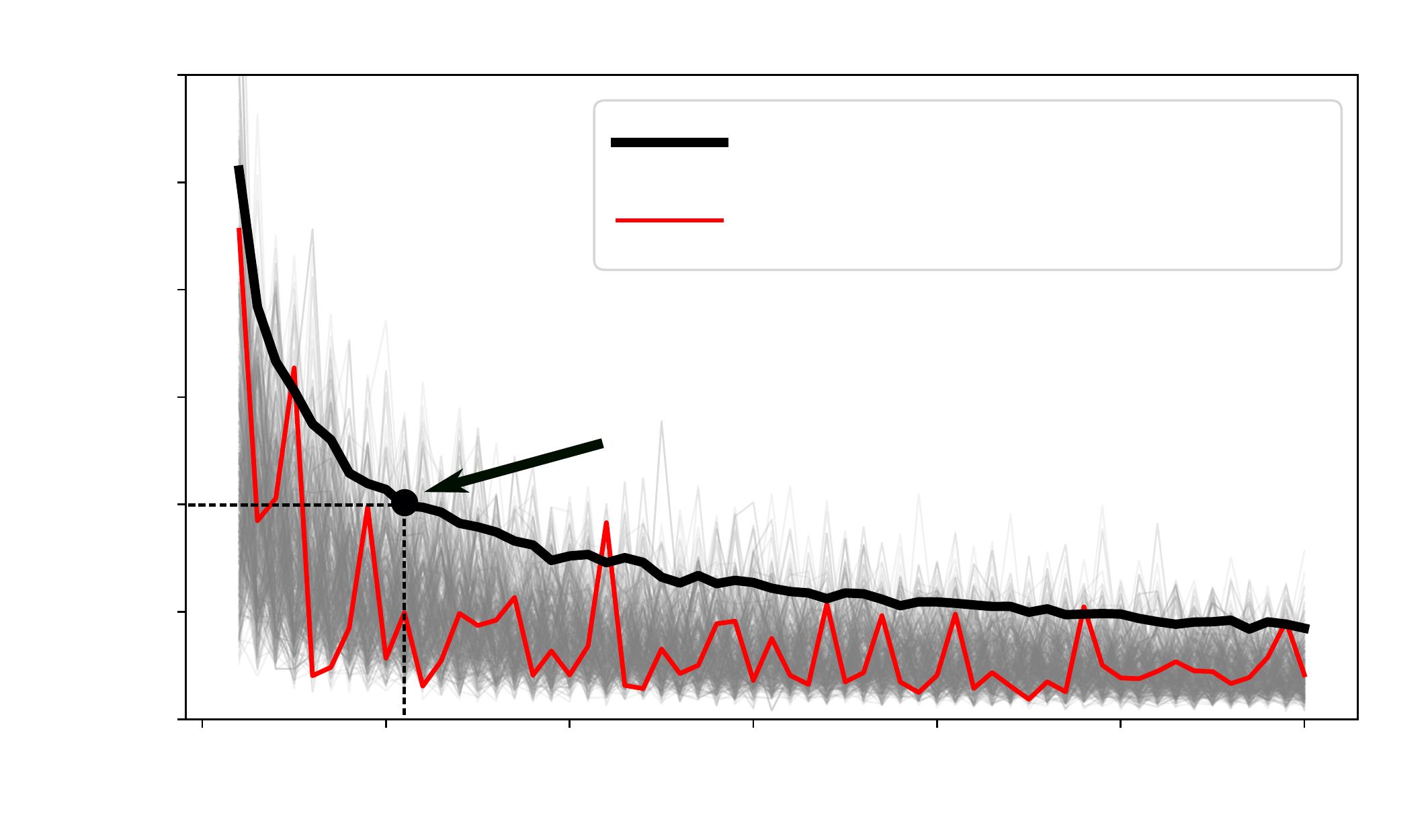} 
		\put(-3,24){\rotatebox{90}{ \large error $\epsilon(t)$ }}
		\put(45,-2){\color{black}{ \large sketch size $t$}} 	
		\put(29,33){\color{black}{\small \parbox{2.5in}{The error $\epsilon(t)$ is smaller than $0.02$ with 95\% probability when $t=550$.}}}	
		
		\put(51.7,47.5){\color{black}{\scriptsize 95th percentile of $\epsilon(t)$ over many runs}} 	
		\put(51.7,42.5){\color{black}{\scriptsize fluctuations of $\epsilon(t)$ during a single run}} 	
		
		\put(13.5,4){\color{black}{\footnotesize 0}} 	
		\put(24,4){\color{black}{\footnotesize 500}} 	
		\put(36,4){\color{black}{\footnotesize 1000}} 						
		\put(49,4){\color{black}{\footnotesize 1500}} 						
		\put(61.5,4){\color{black}{\footnotesize 2000}} 						
		\put(74.5,4){\color{black}{\footnotesize 2500}} 						
		\put(87.5,4){\color{black}{\footnotesize 3000}}

		\put(5,7.5){\color{black}{\footnotesize 0.00}} 	
		\put(5,22.5){\color{black}{\footnotesize 0.02}} 	
		\put(5,37.5){\color{black}{\footnotesize 0.04}} 						
		\put(5,52){\color{black}{\footnotesize 0.06}} 						
	\end{overpic}\hspace{-0.3cm}
	\vspace{+.10in}	
	\caption{Understanding how the errors of a randomized sketching algorithm fluctuate, with a view toward verification and validation. The light gray curves correspond to different runs of the randomized algorithm as the sketch size $t$ is increased. (Larger $t$ corresponds to more computation.) One such curve for a single run is highlighted in red. The black curve represents the 95th percentile of the error $\epsilon(t)$ over many runs. (That is, for any fixed sketch size $t$, the error $\epsilon(t)$ will fall below the black curve with 95\% probability.) Although the black curve is unknown in practice, it can be approximated by using \emph{bootstrap methods} for error estimation. Once an approximation to the black curve is available, it can guide the user in selecting an appropriate sketch size for a desired error tolerance. (adapted from the article~\cite{Lopes2020} with permission from the authors.)} 
	\label{fig:motivation}
	\vspace{-0.3cm}
\end{figure}

\paragraph{Bridging computational and statistical perspectives}
A long-term challenge for verification and validation is bridging the perspectives of computational and statistical research. 
From the viewpoint of statistics, the output of a randomized algorithm can be considered an ``estimate,'' while an exact solution can be considered an ``unknown parameter.''  When randomized algorithms are viewed from this standpoint, the  potential exists to apply a variety of classical statistical methods in the service of error estimation for verification and validation (\cref{fig:motivation}). 
Some of the most well-established  methods in this class are bootstrap, jackknife, and cross-validation. 
Although these methods have been applied in statistics for decades, their potential uses for verification and validation of randomized algorithms has yet to be fully realized;  a recent overview of these connections may be found in~\cite[Secs.~4.5--4.6 and 12.1--12.2]{martinsson_tropp_2020}. 
From the viewpoint of computer science or applied mathematics, these tools often are referred to as methods for \emph{a posteriori error estimation}, because they are designed to quantify error \emph{after} a randomized solution has been computed. 
Furthermore, this approach to error estimation offers an interesting contrast to the worst-case error analysis  because the a posteriori approach tends to be less pessimistic and more adaptive to a given input. 
Also important  is  that these connections with statistical methods are generally not available for deterministic algorithms and they represent a unique opportunity that is directly enabled by the use of randomized algorithms.

\begin{callout}
From the viewpoint of statistics, the output of a randomized algorithm can be regarded as an ``estimate,'' while an exact solution can be regarded as an ``unknown parameter.''  When randomized algorithms are viewed from this standpoint, there is a potential to apply a variety of classical statistical methods in the service of error estimation for verification and validation.
\end{callout}

\paragraph{Integrating randomized algorithms into coupled workflows}%
Growing evidence suggests that randomized algorithms can dramatically reduce the cost of solving problems that require only  moderate accuracy. 
Research is needed to develop error estimates for a wide range of accuracy requirements, from modest accuracy to machine precision. 
Often, randomized algorithms are part of a larger workflow; for example,   a randomized linear algebra solver is used within a finite element model. 
Little attention has been given, however, to quantifying the effects of randomization for predictions in coupled workflows. 
Algorithms that can delineate between the role and effects of noise, errors, data-set distribution, and randomness on overall performance or accuracy would be of great value. 
Such information could be used to identify the largest sources of error and efficiently focus resources to reduce error in downstream prediction goals.

Successful efforts in this area will be transformative. 
Verification and validation of randomized algorithms can impact numerous tasks, from  distinguishing failure of algorithms versus failure of code (e.g.,~an insufficient number of iterations versus a bug), to assessing the accuracy of quantum simulations, to providing error estimates needed to estimate risk in decision-making in natural sciences, engineering, and public policy.

%
%
%
%
% --- End Inserted File ---
% ---- Inserted File ----
\subsection{Software Abstractions} \label{sec:software}
\label{sec:2g:software}
\sublead{R.~Kannan}

\begin{figure}
    \centering
    \includegraphics[width=\columnwidth]{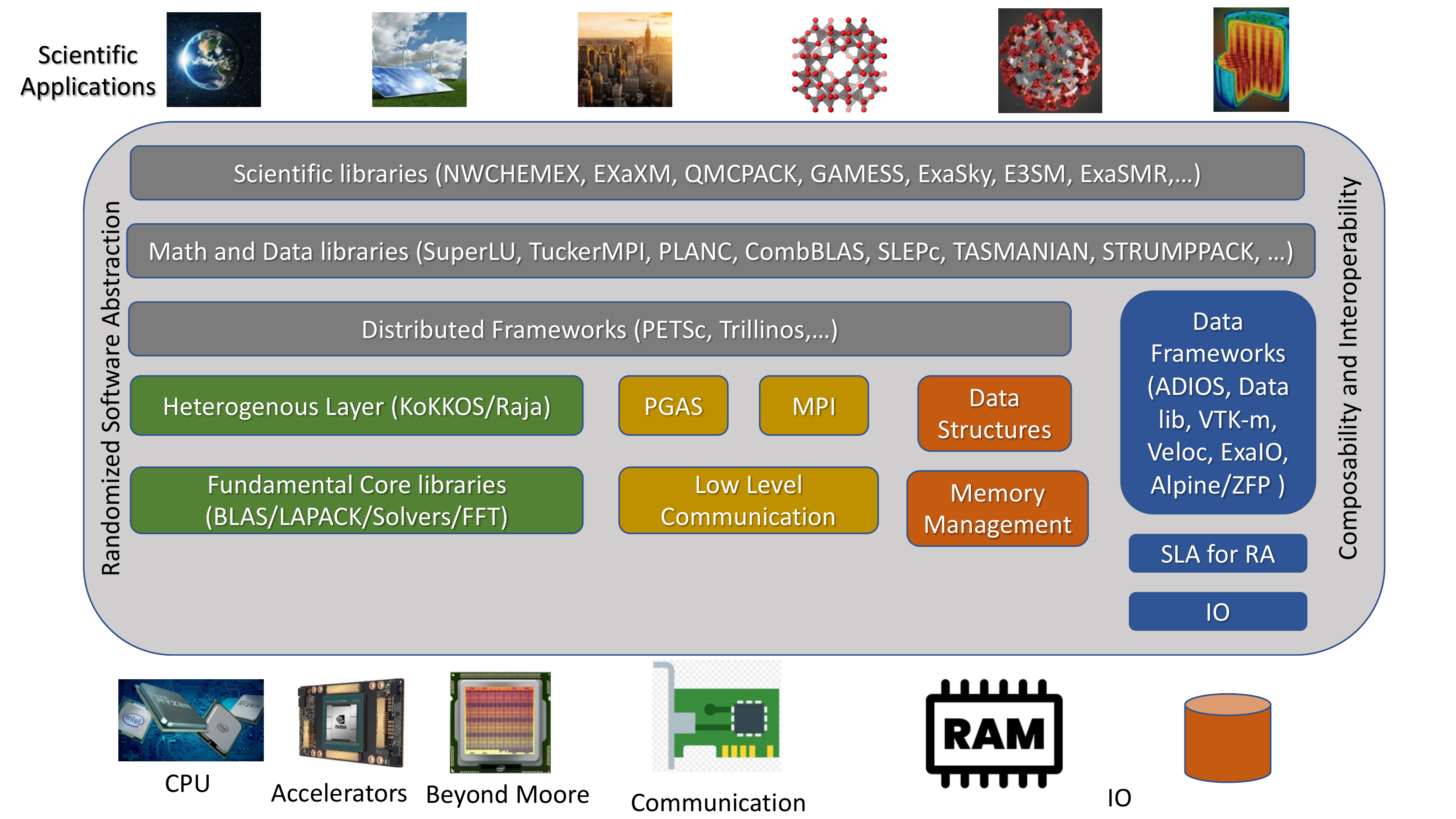}
    \caption{Interoperable randomized abstractions for the DOE software ecosystem.}
    \label{fig:softwareabstraction}
\end{figure}

Scientific data is growing exponentially. Scientists are using faster hardware without any changes on their existing computational algorithm to process the growing data. Because  of increasing challenges of miniaturization of chip design, Moore’s law is becoming obsolete, and we can no longer expect computing power to scale as it has in the past.  Beyond exascale computing, fundamentally novel techniques are  required, such as randomized algorithms to accelerate scientific discoveries on ever-growing data.
In this section we consider software abstractions for randomized algorithms of three important components in von Neumann architectures:
computation, communication, and input/output. 
These randomized abstractions for software should enable increased productivity for developers and make it easy to port existing applications. 
While any specification or standardization would be a community effort, we
lay out  key principles and challenges for designing randomized software abstraction. 

\subsubsection{Composable and Interoperable Randomized Abstractions for Computation} 
Most  randomized algorithms are based on sketching and sampling.
Applying these  techniques to existing numerical and scientific libraries (e.g., BLAS, LAPACK, FFT,
GraphBLAS) and parameterizing them appropriately are  challenging. 
In the near future, most scientific libraries should have some randomization support, such as  the sampled dense-dense matrix multiplication operation in the BLAS.
Generally speaking,
randomized solvers and algorithms are heavily parameterized. 
Determining the externalization of these parameters
and the default values for broader application classes will be an area of further investigation. Many solvers and
scientific libraries leverage structures. 
Randomization techniques must be designed to retain  local properties such as symmetric, hierarchical, block, and neighborhood relations, while also preserving global properties of
the data (e.g., norm, density). 
We envision different software layers, from
fundamental core operations, solvers, and scientific libraries, to applications that will support various modes of 
randomization. Real-world applications should deal with the  combinatorial search problem of composing the varied randomized layers of the software
stack for ideal performance. That is, 
once the different layers of software (Figure~\ref{fig:softwareabstraction}) start supporting various randomized abstractions, each of the component parts will
be optimized, but a workflow with multiple randomized parts will suffer from differences in abstractions. 
\begin{callout}
Interoperable randomized software abstractions across multiple libraries from different entities
(academia, labs, and industry) will require a coordinated effort. 
\end{callout}

\subsubsection{Use Cases for Randomized Communication} 
Communication is important when data is too large to fit in local memory. 
Many distributed
software environments have evolved over different communication libraries, such as MapReduce, MPI, and SHMEM.
Sparsification of data thas long been used to reduce communication. Recently, researchers in deep learning algorithms have considered sparsification of gradients to minimize $All\_Reduce$ time. These randomization strategies for communication will impact the realization of a communication operation. The higher-level
communication layers, such as  MPI and PGAS, and the lower levels, such as Mellanox SHARP, will support different
randomization of communication.  In theory, similar to computation, algorithms designed for randomized
communication will scale to a larger number of processors and data.  The existing partitioners that consider
balancing load versus minimizing the communication volume must also include randomization as an additional constraint. A potential use case for randomization in communication is addressing missing information from faulty nodes. 
That is, instead of communication randomization schemes that are local to every node, randomized communication
for collective calls can give some rigorous approximations.

\subsubsection{Broker Abstractions for Randomized Input/Output} 
There is no free lunch for input/output in randomized algorithms. In
von Neumann--based computer architectures, the memory access is block by block; and in the case of slower storage,
it is also sequential. Hence, low compute-intensive randomized algorithms that involve memory and storage
access, such as graph applications, cannot show significant advantages for overall time to solution. 

A potential
approach is to have a broker-based input/output abstraction for expressing the randomization requirements as service-level
agreements to address memory access issues. 
 Randomization algorithms also require investigation of
fundamental data structures to offer near-real-time random access to the data.  

Apart from these three important directions,  another important topic is education outreach of randomized thinking and programming. Most existing
programming abstractions are based on sequential and deterministic approaches. The computing world faced a
significant challenge when educating programmers on distributed and parallel techniques, and history will repeat
itself with regard to randomized programming. Some of the hurdles that must be faced include educating researchers
on randomized equivalents of existing traditional deterministic solutions using well-defined techniques and novel
algorithm design when ground truths are unavailable.  Aside from the critical aspects of computation,
communication, and input/output, other areas that require new abstractions include
reproducibility, debugging, fault tolerance, journals and logs for reversible computation, instrumentation, and performance evaluation, including metrics and measurements for randomized algorithms.

%

%

%

%
%
%
%
%
%
%

%
%
%
%
%
%
%
%
%
%
%
%
%
%
%
%
%

%

%
%
%
%
%
%
%
%

%

%
%
%
%
%
%

%
%
%
%
%
%
%
%
%
%
%
%
%
%
%
%
%
%
%
%
%
%
%
%
%

%
%
%
%
%
%
%
%
%
%
%

%
%
%
%
%
%
%
%
%
%
%

%
%
%
%
%
%
%
%

%
%
%
%
%

%

%
%
%
%
% --- End Inserted File ---
% --- End Inserted File ---

\clearpage
% ---- Inserted File ----

\section{Themes and Recommendations}
\label{sec:them-recomm}

We surveyed the driving application needs in \cref{sec:drivers} and proposed research directions in \cref{sec:research}.
We close with  identification of the overarching research themes in randomized algorithms
and recommendations for moving forward.

\subsection{Themes}
\label{sec:themes}

Increased computational capacity is required on multiple fronts, including 
ever-higher resolution in simulations for designing more efficient batteries,
processing of massive data from ITER's nuclear fusion experiment,
real-time control of scanning transmission electron microscopy, and
integrating ever more data for numerical weather prediction and long-term climate modeling.
Traditional algorithms, software, and hardware can
no longer keep pace, and randomized
algorithms offer the potential for exponential
increases in computational efficiency,
leading to Theme 1.

\usebox{\thmCapacity}

As we begin to take a fresh look at long-standing problems from
a new perspective, novel approaches emerge, as has taken place in every revolution in science. Consider
the algorithmic advances that have occurred in \emph{serial}
algorithms simply from revisiting them to enable more parallelism.
Just considering how to incorporate randomized algorithms
has already inspired a fresh look at verification and validation
and the hope of incorporating ideas such as bootstrapping from
statistics. This inspires Theme 2.

\usebox{\thmNovel}

In signal processing, the rate at which a continuous signal is sampled is usually selected according to the Nyquist–-Shannon sampling theorem, depending on the highest frequency present in the signal to be sampled. In the 2000s, a great deal of excitement was generated by the concept of \emph{compressed sensing}, which allows sampling at a potentially much lower rate by exploiting randomized sampling and the common signal properties of \emph{sparsity} or \emph{compressibility}.
For magnetic resonance imaging in health care, this
has had significant real-world impacts
such as scan times reduced by 50\% and
increased accuracies for the same scan times.%
\footnote{See, e.g., \href{https://www.usa.philips.com/healthcare/education-resources/publications/fieldstrength/how-compressed-sense-makes-mri-faster}
  {How Compressed SENSE makes MRI Faster} from Philips.}
In computing, we know that we can realize huge gains in efficiency if we allow for the occasional numerical error.
In emerging computing regimes such as quantum and neuromorphic  computing, imprecision is inherent.
This brings us to Theme 3. 
\usebox{\thmHW}

Indiscriminate use of randomness in algorithms is not what is proposed.
For instance, sampling has long been popular in handling large-scale graphs,
but the errors cannot be bounded when naive approaches are employed.
Instead, methods have been devised that use more sophisticated sampling strategies
and provide probabilistic error guarantees (e.g., \cite{SePiKo13a}).
Rather, we propose
the integration of domain-informed sampling techniques, sketching with theoretical
guarantees, and online computations achieving almost the same accuracy as
static computations. In turn, this approach will require substantial efforts
in overcoming both seen and unseen technical hurdles in  order to achieve the
deployment of randomized algorithms to key DOE science and national security applications,
as in Theme 4.

\usebox{\thmBarrier}

Even though inputs to our calculations have a degree of uncertainty, skepticism about randomized algorithms remains. Various discretizations are accepted as the cost of doing business, 
and faulty computations due to errant cosmic particles are all but certain in exascale computers.
Indeed, poorly crafted randomized algorithms
can be terribly inaccurate, but the same is true of any numerical method.
The answer is not only to develop better algorithms
but also to work on educating our colleagues and users
about the advantages and even necessity of randomized algorithms per Theme 5.
\usebox{\thmPsych}

DOE has funded decades of world-class research in computational science at the national labs and at universities.
Nevertheless, barriers to randomized algorithms persist because naive strategies are rarely competitive
with well-understood and optimized deterministic approaches. 
We need \emph{new expertise} to ably surmount the technical hurdles outlined above,
which means outreach to a broader constituency of researchers and is
 the motivation for Theme 6.
One could  argue that this is analogous to the
integration of mathematics, computer science,
and domain expertise in the founding of
computational science and engineering.

\usebox{\thmExpertise}

\subsection{Recommendations}
\label{sec:recommendations}

A concerted research program in randomized algorithms will require a mixture of efforts for success. 
Pursuing research programs in one area at the expense of other areas will slow progress along all fronts.
Here we  present six recommended priorities for research efforts. %

The importance of basic research, promoted in Recommendation 1, cannot be overstated.
Such research may be in smaller stand-alone projects or part of joint efforts.
Regardless of how it takes place, the researchers engaged in foundational research
will need to commit to engaging with algorithmic researchers to bring
the theory into practice.

\usebox{\recfoundations}

Development of randomized algorithms is the cornerstone of the proposed effort,
per  Recommendation 2.
The role of algorithm researchers is to unravel the theory into
working prototypes of methods, tested on idealized problems
that reflect real-world applications.
Algorithmic researchers will need to engage with 
applications and possibly emerging hardware.

\usebox{\recalgs}

While one may imagine application-agnostic randomized algorithms,
the reality is that most applications will need tailoring
of approaches to the domain.
This might be in the form of sampling strategies,
such as appropriate stratified sampling,
or it might go so far as requiring new specific theory.
The goal of  Recommendation 3 is to focus on customizing
solutions to specific applications and their individual needs.

\usebox{\recapps}

Randomly distributed data, as in sparse matrices, has
always bedeviled computational efficiency. One might conclude 
that introducing more randomness could be detrimental
to computational efficiency. However,
the next-generation hardware will have inherent randomness.
The goal of  Recommendation 4 is to develop and co-design
randomized algorithms that are scalable.

\usebox{\rechw}

The shift to randomized algorithms represents a fundamentally new direction for DOE;
thus, it requires new specializations that are not currently represented in its
research program. 
 Recommendation 5 has to do with broadening the teams
of researchers that are engaged with DOE via this new effort.
Pursuing this effort will accelerate the success
of deploying randomized algorithms over the next decade
and  bring a broader perspective to DOE's problems overall.
Without a specific push in the direction of diversification, it will be too easy to fall back
on the known and trusted personnel in the current program.

\usebox{\recoutreach}

In concert with efforts in computer science and elsewhere,
we also need to consider the standardization of randomized algorithms,
which is the focus of  Recommendation 6.
It is difficult to think of standardization in a topic that is still so new
in computational science, yet the next decade should bring a wealth of
advances. For these advances to have the greatest impact, they will
need to be incorporated into software frameworks, which will require
standards in how to do so.

\usebox{\recstandardization}

%
%
%
%
% --- End Inserted File ---

\clearpage
\phantomsection
\addcontentsline{toc}{section}{\refname}
\bibliographystyle{abbrvnatmod} 

%%\\n

%
\clearpage
\appendix
\phantomsection
\addcontentsline{toc}{section}{Appendices}

% ---- Inserted File ----
\section{Workshop Information}
\label{app:workshop-agenda}

\subsection*{Workshop Agenda}

\newcommand{\emti}[1]{\emph{#1}} %
\emph{All times Eastern}

\subsubsection*{Part 1 Bootcamp, Randomized Algorithms Tutorials: December 2, 2020 }

\begin{longtable}{l p{5in}}
11:00 AM--11:30 AM & Introduction and Charge \newline
	Steven Lee, US Department of Energy, ASCR \\*
11:30 AM--11:55 AM & \emti{Probability for Randomized Algorithms}
    \newline
	Miles Lopes, University of California, Davis \\*
11:55 AM--12:15 PM & Discussion \\*
12:15 PM--12:40 PM & \emti{(Practical) Randomized Algorithms for Graph Problems} \newline
	C. Seshadhri, University of California, Santa Cruz \\*
12:40 PM--1:00 PM & Discussion \\*
1:00 PM--1:15 PM & Break \\*
1:15 PM--1:40 PM & \emti{Randomized Algorithms in Linear Algebra and Scientific Computing} \newline
	Per Gunnar Martinsson, University of Texas, Austin \\*
1:40 PM--2:00 PM & Discussion \\*
2:00 PM--2:25 PM & \emti{A Whirlwind Tour of Stochastic Optimization} \newline
	John Duchi, Stanford University \\*
2:25 PM--2:45 PM & Discussion \\*
2:45 PM--3:00 PM & Wrap-up Part 1, Day 1 \\*
\end{longtable}

\subsubsection*{Part 1 Bootcamp, Science and Engineering Opportunities: December 3, 2020 }

\begin{longtable}{l p{5in}}
11:00 AM--11:15 AM & Agenda and Randomized Breakout Structure\\*
11:15 AM--11:30 AM & \emti{Random Thoughts on Randomized Algorithms} \newline
	Julia Ling, 
	Citrine Informatics\\*
11:30 AM--11:45 AM & \emti{Random Thoughts on Randomized Algorithms} \newline
	Salman Habib, 
	Computational Science Division, 
	Argonne National Laboratory \\*
11:45 AM--12:00 PM & \emti{The Intersection of Discrete Math and Chemistry: Opportunities for Randomized Algorithms} \newline
	Aurora Clark, 
	Department of Chemistry,
	Washington State University \\*
12:00 PM--12:30 PM & Application Drivers Panel \#1 Discussion \\*
12:30 PM--12:45 PM & Break \\*
12:45 PM--1:00 PM & \emti{Thoughts on
Applications of RASC
for Inverse Problems in Imaging} \newline
	Charles Bouman, 
	School of Electrical and Computer Engineering,
	Purdue University \\*
1:00 PM--1:15 PM & \emti{Imaging Biology}
 \newline
	Peter Zwart, 
	MBIB / BSISB / CAMERA,
    Lawrence Berkeley National Laboratory \\*
1:15 PM--1:35 PM & Application Drivers Panel \#2 Discussion \\*
1:35 PM--1:45 PM & Breakout Group Instructions \\*
1:45 PM--2:45 PM & Breakout Groups on Priority Research Opportunities \\*
2:45 PM--3:00 PM & Wrap-up Part 1 \\*
\end{longtable}

\newpage

\subsubsection*{Part 2: January 6, 2021}

\begin{longtable}{l p{5in}}
11:00 AM--11:15 AM & Agenda and Part 2 Aims\\*
11:15 AM--12:15 PM & \emti{Federal Agency Panel} \newline
	Fariba Fahroo, AFOSR \newline
	Steven Lee, DOE  \newline
	Reza Malek-Madani, ONR \newline
	Grace Peng,  NIH\\*
12:15 PM--12:30 PM & Instructions for Day 1 Breakouts  \\*
12:30 PM--1:30 PM & Breakouts: Application Needs and Drivers \\*
1:30 PM--1:45 PM & Break \\*
1:45 PM--2:30 PM & Application Needs and Drivers Breakout Report Backs \\*
2:30 PM--2:45 PM & Wrap-up Part 2, Day 1 \\*
2:45 PM--3:00 PM & Writing Committee Meeting \\*
\end{longtable}

\subsubsection*{Part 2: January 7, 2021}

\begin{longtable}{l p{5in}}
11:00 AM--11:15 AM & Welcome to Day 2 and Summary of Day 1\\*
11:15 AM--11:30 AM & Instructions for Day 2 Breakouts\\*
11:30 AM--12:45 PM & Breakouts: Foundational/Computational Research  \\*
12:45 PM--1:00 PM & Break \\*
1:00 PM--1:45 PM & Foundational/Computational Research Breakout Report Backs \\*
1:45 PM--2:00 PM & Workshop Wrap-up \\*
2:00 PM--3:00 PM & Writing Committee Meeting \\*
\end{longtable}

\subsection*{Workshop Format}
Part 1 of the workshop took advantage of expanded capacity enabled by a virtual format. 
Attendees pre-registered for the workshop and participated in discussion and breakouts. Additional attendees were able to follow the bootcamp (all of Part 1 except for the breakouts) in real time through a youtube stream.   

Registration to Part 2 required submission of a 200-word thesis statement on long-term research needs. These statements were used to structure the breakouts in Part 2 and seed the report content.

% --- End Inserted File ---

\clearpage
% ---- Inserted File ----
\section{Workshop Participants}
\label{app:participants}

\tablefirsthead{\bf Name & \bf Institution\\ \hline}
\tablehead{\bf Name & \bf Institution\\ \hline}
\tabletail{\hline \multicolumn{2}{r}{\emph{Continued on next page}}\\}
\tablelasttail{\hline}
\begin{supertabular}{ll}
Karthik Aadithya & Sandia National Laboratories\\
Erin Acquesta & Sandia National Laboratories\\
Nagesh Adluru & University of Wisconsin-Madison\\
Bulbul Ahmmed & Los Alamos National Laboratory\\
James Ahrens & Los Alamos National Laboratory\\
Sinan Aksoy & Pacific Northwest National Laboratory\\
H. Metin Aktulga & Michigan State University\\
Srinivas Aluru & Georgia Institute of Technology\\
Vinay Amatya & Pacific Northwest National Laboratory\\
Oluwamayowa Amusat & Lawrence Berkeley National Laboratory\\
Marian Anghel & Los Alamos National Laboratory\\
Mihai Anitescu & Argonne National Laboratory\\
Rick Archibald & Oak Ridge National Laboratory\\
Yeva F. Ashari & Institut Teknologi Bandung (ITB)\\
Selin Aslan & Argonne National Laboratory\\
Ahmed Attia & Argonne National Laboratory\\
Alan Ayala & University of Tennessee, Knoxville\\
Jacob Badger & University of Texas at Austin\\
Siwar Badreddine & INRIA\\
Zhe Bai & Lawrence Berkeley National Laboratory\\
Craig Bakker & Pacific Northwest National Laboratory\\
Prasanna Balaprakash & Argonne National Laboratory\\
Grey Ballard & Wake Forest University\\
David Barajas-Solano & Pacific Northwest National Laboratory\\
Andrew Barker & Lawrence Livermore National Laboratory\\
Andreas B\char0228\@rtschi & Los Alamos National Laboratory\\
William Beckner & University of Texas at Austin\\
Getachew Befekadu & Morgan State University\\
Julie Bessac & Argonne National Laboratory\\
Vivek Bharadwaj & University of California, Berkeley\\
Noah Birge & Los Alamos National Laboratory\\
Simon Bolding & Los Alamos National Laboratory\\
Raghu Bollapragada & University of Texas at Austin\\
Erik Boman & Sandia National Laboratories\\
Brian Borchers & New Mexico Institute of Mining \& Technology\\
Tyler Borgwardt & Los Alamos National Laboratory\\
Kerry Bossler & Sandia National Laboratories\\
Nicolas Boumal & Swiss Federal Institute of Technology Lausanne\\
Charles Bouman & Purdue University\\
Robert Bridges & Oak Ridge National Laboratory\\
Christopher Brislawn & Los Alamos National Laboratory\\
David Brown & Lawrence Berkeley National Laboratory\\
Johannes Brust & Argonne National Laboratory\\
Ayd\i n Bulu\c{c} & Lawrence Berkeley National Laboratory\\
Dmitry Burov & California Institute of Technology\\
Geunyeong Byeon & Argonne National Laboratory\\
Daan Camps & Lawrence Berkeley National Laboratory\\
Suma Cardwell & Sandia National Laboratories\\
Erin Carson & Charles University\\
Umit Catalyurek & Georgia Institute of Technology\\
Luis Chacon & Los Alamos National Laboratory\\
Moses Charikar & Stanford University\\
Sridhar Chellappa & Max Planck Institute, Magdeburg, Germany\\
Chao Chen & University of Texas at Austin\\
Xiao Chen & Lawrence Livermore National Laboratory\\
Changqing Cheng & Binghamton University\\
Hari Chhetri & Oak Ridge National Laboratory\\
Eric Chi & North Carolina State University\\
Jocelyn Chi & North Carolina State University\\
Jong Choi & Oak Ridge National Laboratory\\
Youngsoo Choi & Lawrence Livermore National Laboratory\\
Pinghan Chu & Los Alamos National Laboratory\\
Julianne Chung & Virginia Tech\\
Matthias Chung & Virginia Tech\\
Richard Clancy & University of Colorado\\
Aurora Clark & Washington State University\\
Lisa Claus & Lawrence Berkeley National Laboratory\\
Mark Coletti & Oak Ridge National Laboratory\\
Emil Constantinescu & Argonne National Laboratory\\
Alice Cortinovis & Ecole Polytechnique Federale de Lausanne\\
Martin Crawford & Sandia National Laboratories\\
Jody Crisp & Oak Ridge Institute for Science and Education\\
Eric Cyr & Sandia National Laboratories\\
Prasanna Date & Oak Ridge National Laboratory\\
Ieva Dauzickaite & University of Reading\\
Mike Davis & Cray/HPE\\
Eduardo D'Azevedo & Oak Ridge National Laboratory\\
Saibal De & University of Michigan, Ann Arbor\\
Eric de Sturler & Virginia Tech\\
Nathan DeBardeleben & Los Alamos National Laboratory\\
Suman Debnath & Oak Ridge National Laboratory\\
Victor DeCaria & Oak Ridge National Laboratory\\
Anthony DeGennaro & Brookhaven National Laboratory\\
Marta D'Elia & Sandia National Laboratories\\
James Demmel & University of California, Berkeley\\
Aditya Devarakonda & Johns Hopkins University\\
Som Dhulipala & Idaho National Laboratory\\
Zichao Wendy Di & Argonne National Laboratory\\
Wen Ding & Columbia University\\
Zhiyan Ding & University of Wisconsin, Madison\\
Edgar Dobriban & University of Pennsylvania\\
Jin Dong & Oak Ridge National Laboratory\\
Jack Dongarra & University of Tennessee\\
J\char0225\@n Drgo\v{n}a & Pacific Northwest National Laboratory\\
John Duchi & Stanford University\\
Jed Duersch & Sandia National Laboratories\\
Alan Edelman & Massachusetts Institute of Technology\\
Karl Elbakian & Los Alamos National Laboratory\\
Amr El-Bakry & ExxonMobil Upstream Integrated Solutions\\
Eirik Endeve & Oak Ridge National Laboratory\\
Ethan Epperly & California Institute of Technology\\
N. Benjamin Erichson & University of California, Berkeley\\
Laureano Escudero & Universidad Rey Juan Carlos\\
Jenniffer Estrada Lupianez & Los Alamos National Laboratory\\
Barry Fadness & Lawrence Livermore National Laboratory\\
Fariba Fahroo & AFOSR\\
Doreen Fan & Lawrence Berkeley National Laboratory\\
Yilin Fang & Pacific Northwest National Laboratory\\
Ionut Farcas & University of Texas at Austin\\
Ali Farghadan & University of Michigan\\
Maryam Fazel & University of Washington\\
Adrian Febre & MacLeod Ale Brewing Company\\
S. M. Ferdous & Purdue University\\
Gregory Fiechtner & DOE-BES\\
Thomas Flynn & Brookhaven National Laboratory\\
Chris Frazier & Sandia National Laboratories\\
Kasimir Gabert & Sandia National Laboratories\\
Jim Gaffney & Lawrence Livermore National Laboratory\\
Shreyas Gaikwad & University of Texas at Austin\\
Sayan Ghosh & Pacific Northwest National Laboratory\\
Sujit Ghosh & North Carolina State University\\
Pieter Ghysels & Lawrence Berkeley National Laboratory\\
Jay Gibble & Unaffiliated\\
Anna Gilbert & Yale University\\
Alex Gittens & Rensselaer Polytechnic Institute\\
Andrew Glaws & National Renewable Energy Laboratory\\
Abeynaya Gnanasekaran & Stanford University\\
Johnny Goett & Los Alamos National Laboratory\\
Emily Gottry & Azusa Pacific University\\
Carlo Graziani & Argonne National Laboratory\\
Andre Green & Los Alamos National Laboratory\\
Jared Greenwald & DOE-NNSA\\
Laura Grigori & INRIA\\
Sidharth GS & Los Alamos National Laboratory\\
Qiang Guan & Kent State University\\
Dan Gunter & Lawrence Berkeley National Laboratory\\
Castro Guzm\char0225\@n & University of S\char0227\@o Paulo\\
Salman Habib & Argonne National Laboratory\\
Hamed Haddadi & Corning Incorporated\\
Aric Hagberg & Los Alamos National Laboratory\\
Mahantesh Halappanavar & Pacific Northwest National Laboratory\\
Kathleen Hamilton & Oak Ridge National Laboratory\\
Wesley Hamilton & University of North Carolina at Chapel Hill\\
David Han & University of Texas at San Antonio\\
Songfeng Han & Corning Incorporated\\
Joseph Hart & Sandia National Laboratories\\
Tucker Hartland & University of California, Merced\\
Rebecca Hartman-Baker & Lawrence Berkeley National Laboratory\\
Tamir Hasan & Los Alamos National Laboratory\\
Cory Hauck & Oak Ridge National Laboratory\\
Koby Hayashi & Georgia Institute of Technology\\
Barbara Helland & DOE-ASCR\\
Bruce Hendrickson & Lawrence Livermore National Laboratory\\
Stefan Henneking & University of Texas at Austin\\
Benjamin Hernandez & Oak Ridge National Laboratory\\
Felix Herrmann & Georgia Institute of Technology\\
Judy Hill & Oak Ridge National Laboratory\\
Jeffrey Hittinger & Lawrence Livermore National Laboratory\\
Yang Ho & Sandia National Laboratories\\
Thuc Hoang & DOE-NNSA\\
Aaron Holder & DOE-BES\\
David Hong & University of Pennsylvania\\
Shahadat Hossain & University of Lethbridge\\
Paul Hovland & Argonne National Laboratory\\
Yicheng Hu & University of Wisconsin, Madison\\
Zixi Hu & Lawrence Berkeley National Laboratory\\
Kuang Huang & Columbia University\\
Yehan Huang & Stanford University\\
Ryan Humble & Stanford University\\
Travis Humble & Oak Ridge National Laboratory\\
Xiaoming Huo & Georgia Institute of Technology\\
Hoon Hwangbo & University of Tennessee, Knoxville\\
Jeffery Hyman & Los Alamos National Laboratory\\
Mac Hyman & Tulane University\\
Mahdi Imani & George Washington University\\
Alexander Infanger & Stanford University\\
Vighnesh Iyer & University of California, Berkeley\\
Elchin Jafarov & Los Alamos National Laboratory\\
John Jakeman & Sandia National Laboratories\\
Jacek Jakowski & Oak Ridge National Laboratory\\
Carter Jameson & North Carolina State University\\
Ruhui Jin & University of Texas at Austin\\
Hans Johansen & Lawrence Berkeley National Laboratory\\
Erik Johnson & University of Colorado, Boulder\\
Patrick Johnstone & Brookhaven National Laboratory\\
Erika Jones & Federal Contractor\\
Piet Jones & Oak Ridge National Laboratory\\
Caleb Ju & Georgia Institute of Technology\\
Vesa Kaarnioja & LUT University, Finland\\
Gokberk Kabacaoglu & Flatiron Institute, Simons Foundation\\
Chandrika Kamath & Lawrence Livermore National Laboratory\\
Ulugbek Kamilov & Washington University in St. Louis\\
Raghavendra Kanakagiri & Indian Institute Of Technology Tirupati\\
Lulu Kang & Illinois Institute of Technology\\
Shinhoo Kang & Argonne National Laboratory\\
Ramakrishnan Kannan & Oak Ridge National Laboratory\\
Nikita Kapur & University of Kansas\\
Kolja Kauder & Brookhaven National Laboratory\\
William Kay & Oak Ridge National Laboratory\\
Jeffrey Keithley & Los Alamos National Laboratory\\
Carl Kelley & North Carolina State University\\
Luke Kersting & Sandia National Laboratories\\
Arif Khan & Pacific Northwest National Laboratory\\
Ratna Khatri & George Mason University\\
Joe Kileel & University of Texas at Austin\\
Aleksandra Kim & ETH Zurich\\
Kibaek Kim & Argonne National Laboratory\\
Kyungjoo Kim & Sandia National Laboratories\\
Tamara Kolda & Sandia National Laboratories\\
Tzanio Kolev & Lawrence Livermore National Laboratory\\
Hemanth Kolla & Sandia National Laboratories\\
Olivera Kotevska & Oak Ridge National Laboratory\\
Aditi Krishnapriyan & Lawrence Berkeley National Laboratory\\
Simge Kucukyavuz & Northwestern University\\
Paul Laiu & Oak Ridge National Laboratory\\
Kevin Lamb & Los Alamos National Laboratory\\
Brett Larsen & Stanford University\\
Jeffrey Larson & Argonne National Laboratory\\
Randall Laviolette & DOE-ASCR\\
Earl Lawrence & Los Alamos National Laboratory\\
Joel LeBlanc & Michigan Tech Research Institute\\
Steven Lee & DOE-ASCR\\
Rich Lehoucq & Sandia National Laboratories\\
Ryan Lekivetz & JMP\\
Nathan Lemons & Los Alamos National Laboratory\\
Andrew Leong & Los Alamos National Laboratory\\
Eitan Levin & California Institute of Technology\\
Matthew Levine & California Institute of Technology\\
Cannada Lewis & Sandia National Laboratories\\
Sven Leyffer & Argonne National Laboratory\\
Sherry Li & Lawrence Berkeley National Laboratory\\
Wenting Li & Los Alamos National Laboratory\\
Ying Wai Li & Los Alamos National Laboratory\\
Florence Lin & Retired\\
Meifeng Lin & Brookhaven National Laboratory\\
Neil Lindquist & University of Tennessee\\
Julia Ling & Citrine Informatics\\
Bowen Liu & North Carolina State University\\
Frank Liu & Oak Ridge National Laboratory\\
Jia Liu & University of West Florida\\
Jinsong Liu & University of Texas at Austin\\
Xinchao Liu & University of Arkansas\\
Yan Liu & Oak Ridge National Laboratory\\
Yang Liu & Lawrence Berkeley National Laboratory\\
Li-Ta Lo & Los Alamos National Laboratory\\
Lena Lopatina & Los Alamos National Laboratory\\
Miles Lopes & University of California, Davis\\
Vanessa Lopez-Marrero & Brookhaven National Laboratory\\
Vincenzo Lordi & Lawrence Livermore National Laboratory\\
Kathryn Lund & Charles University\\
Erik Lundgren & Los Alamos National Laboratory\\
Dalton Lunga & Oak Ridge National Laboratory\\
Hengrui Luo & Lawrence Berkeley National Laboratory\\
Xihaier Luo & Brookhaven National Laboratory \\
Massimiliano Lupo Pasini & Oak Ridge National Laboratory\\
Piotr Luszczek & University of Tennessee\\
Jonathan Maack & National Renewable Energy Laboratory\\
Arthur Maccabe & Oak Ridge National Laboratory\\
Sarah Mackay & Lawrence Livermore National Laboratory\\
Sandeep Madireddy & Argonne National Laboratory\\
Michael Mahoney & University of California, Berkeley\\
Dhairya Malhotra & Flatiron Institute, Simons Foundation\\
Osman Asif Malik & University of Colorado, Boulder\\
Indu Manickam & Sandia National Laboratories\\
Aniruddha Marathe & Lawrence Livermore National Laboratory\\
Oana Marin & Argonne National Laboratory\\
Osni Marques & Lawrence Berkeley National Laboratory\\
Zach Marshall & Lawrence Berkeley National Laboratory\\
Per-Gunnar Martinsson & University of Texas at Austin\\
David Dennis Mascarenas & Los Alamos National Laboratory\\
Estelle Massart & University of Oxford\\
Anna Matsekh & Los Alamos National Laboratory\\
Yury Maximov & Los Alamos National Laboratory\\
Lennon McCartney & Corning Incorporated\\
Ryan McClarren & University of Notre Dame\\
Hugh Medal & University of Tennessee\\
Maksim Melnichenko & University of Tennessee\\
Dean Menezes & University of California, Los Angeles\\
Matt Menickelly & Argonne National Laboratory\\
Agnieszka Miedlar & University of Kansas\\
Benjamin Miller & University of Texas at Austin\\
Richard Mills & Argonne National Laboratory\\
Michael Minion & Lawrence Berkeley National Laboratory\\
Narasinga Rao Miniskar & Oak Ridge National Laboratory\\
Raul Miranda & DOE\\
Laura Monroe & Los Alamos National Laboratory\\
Leslie Moore & Los Alamos National Laboratory\\
Jose Morales & University of Texas at San Antonio\\
Dmitriy Morozov & Lawrence Berkeley National Laboratory\\
Juliane Mueller & Lawrence Berkeley National Laboratory\\
Aliasgar Musani & Indian Institute of Technology Tirupati\\
Jeremy Myers & William \& Mary, Sandia National Laboratories\\
Kary Myers & Los Alamos National Laboratory\\
Adan Myers y Gutierrez & Los Alamos National Laboratory\\
Samuel Myren & Los Alamos National Laboratory\\
Nicholas Nelsen & California Institute of Technology\\
Jelani Nelson & University of California, Berkeley\\
Esmond Ng & Lawrence Berkeley National Laboratory\\
Kam Ng & Corning Incorporated\\
Arnur Nigmetov & Lawrence Berkeley National Laboratory\\
Israt Nisa & Lawrence Berkeley National Laboratory\\
Marcus Noack & Lawrence Berkeley National Laboratory\\
Sewoong Oh & University of Washington\\
Thomas O'Leary-Roseberry & University of Texas at Austin\\
Vladyslav Oles & Oak Ridge National Laboratory\\
Eliza O'Reilly & California Institute of Technology\\
George Ostrouchov & Oak Ridge National Laboratory\\
Art Owen & Stanford University\\
Jonathan Ozik & Argonne National Laboratory\\
Mirjeta Pasha & Arizona State University\\
Vivak Patel & University of Wisconsin, Madison\\
Abani Patra & Tufts University\\
Jagabandhu Paul & California Institute of Technology\\
Grace Peng & NIH\\
Richard Peng & Georgia Institute of Technology\\
Talita Perciano & Lawrence Berkeley National Laboratory\\
Mauro Perego & Sandia National Laboratories\\
Ronan Perry & Johns Hopkins University\\
Elliott Perryman & Lawrence Berkeley National Laboratory\\
Kalyan Perumalla & Oak Ridge National Laboratory\\
Matt Peterson & Sandia National Laboratories\\
Russell Philley & University of Texas at Austin\\
Cynthia Phillips & Sandia National Laboratories\\
Eric Phipps & Sandia National Laboratories\\
Ali Pinar & Sandia National Laboratories\\
Maria Pinilla-Orjuela & Los Alamos National Laboratory\\
Robinson Pino & DOE-ASCR\\
Harun Pirim & Mississippi State University\\
Todd Plantenga & FireEye, Inc.\\
Nick Polydorides & University of Edinburgh\\
Ravi Ponmalai & University of California, Irvine\\
Gabriel Popoola & Sandia National Laboratories\\
Alex Pothen & Purdue University\\
Kumaraguru Prabakar & National Renewable Energy Laboratory\\
Benjamin Priest & Lawrence Livermore National Laboratory\\
Andrey Prokopenko & Oak Ridge National Laboratory\\
Constantin Puiu & Oxford University\\
Mihaela Quirk & DOE-NNSA\\
Robert Rallo & Pacific Northwest National Laboratory\\
Teresa Ranadive & University of Maryland\\
Vishwas Rao & Argonne National Laboratory\\
Navamita Ray & Los Alamos National Laboratory\\
Thomas Reeves & Cornell University\\
Lothar Reichel & Kent  State University\\
Yihui Ren & Brookhaven National Laboratory\\
Rosemary renaut & Arizona State University\\
Ashwin Renganathan & Argonne National Laboratory\\
Viktor Reshniak & Oak Ridge National Laboratory\\
Juan Restrepo & Oak Ridge National Laboratory\\
Elina Robeva & University of British Columbia\\
Heinrich Roder & Biodesix\\
Theron Rodgers & Sandia National Laboratories\\
Anna R\char0246\@rich & University of Stuttgart\\
Cl\char0233\@ment Royer & Universit\char0233\@ Paris Dauphine-PSL\\
Johann Rudi & Argonne National Laboratory\\
Minseok Ryu & Argonne National Laboratory\\
Anna Sabin & Pacific Northwest National Laboratory\\
Sonia Sachs & DOE-ASCR\\
Ilya Safro & University of Delaware\\
Cosmin Safta & Sandia National Laboratories\\
Roshni Sahoo & Stanford University\\
Arvind Saibaba & North Carolina State University\\
Jacob Salazar Solano & University of Texas at Austin\\
Zain Saleem & Argonne National Laboratory\\
Samitha Samaranayake & Cornell University\\
Arun Sathanur & Pacific Northwest National Laboratory\\
Florian Schaefer & California Institute of Technology\\
Drew Schmidt & Oak Ridge National Laboratory\\
Sudip Seal & Oak Ridge National Laboratory\\
Tom Seidl & Sandia National Laboratories\\
Oguz Selvitopi & Lawrence Berkeley National Laboratory\\
C. (Sesh) Seshadhri & University of California, Santa Cruz\\
John Shalf & Lawrence Berkeley National Laboratory\\
Nek Sharan & Los Alamos National Laboratory\\
Ruslan Shaydulin & Argonne National Laboratory\\
Deborah Shutt & Los Alamos National Laboratory\\
Erotokritos Skordilis & National Renewable Energy Laboratory\\
George Slota & Rensselaer Polytechnic Institute\\
Emily Smith & DOE-BES\\
J. Darby Smith & Sandia National Laboratories\\
Youngseok Song & Colorado State University\\
Braden Soper & Lawrence Livermore National Laboratory\\
Matthew Sottile & Lawrence Livermore National Laboratory\\
Raymond Spiteri & University of Saskatchewan\\
Prateek Srivastava & Rochester Institute of Technology\\
Andreas Stathopoulos & William \& Mary\\
Gilbert Strang & Massachusetts Institute of Technology\\
Roy Streit & Metron\\
Nicholas Stull & Los Alamos National Laboratory\\
Omer Subasi & Pacific Northwest National Laboratory\\
Anirudh Subramanyam & Argonne National Laboratory\\
Aditya Sundararajan & Oak Ridge National Laboratory\\
Kelly Swanson & Lawrence Livermore National Laboratory\\
Katarzyna Swirydowicz & Pacific Northwest National Laboratory\\
Peter Tait & McMaster University\\
Yu-Hang Tang & Lawrence Berkeley National Laboratory\\
Valerie Taylor & Argonne National Laboratory\\
Keita Teranishi & Sandia National Laboratories\\
Satya Tetala & Colorado State University\\
James Theiler & Los Alamos National Laboratory\\
Yifeng Tian & Los Alamos National Laboratory\\
David Tralli & The Aerospace Corporation\\
Anh Tran & Sandia National Laboratories\\
Hoang Tran & Oak Ridge National Laboratory\\
Benjamin Treweek & Sandia National Laboratories\\
Joel Tropp & California Institute of Technology\\
Antonino Tumeo & Pacific Northwest National Laboratory\\
Raymond Tuminaro & Sandia National Laboratories\\
Bora Ucar & CNRS and LIP ENS de Lyon\\
Madeleine Udell & Cornell University\\
Juliette Ugirumurera & National Renewable Energy Laboratory\\
Ugochukwu Ugwu & Kent State University\\
Jiblal Upadhya & Middle Tennessee State University\\
Nathan Urban & Brookhaven National Laboratory\\
Roel Van Beeumen & Lawrence Berkeley National Laboratory\\
Nick Vannieuwenhoven & KU Leuven\\
Heidy Vega & University of California, San Diego\\
Nicolas Venkovic & Cerfacs\\
Jeff Victor & General Electric\\
Draguna Vrabie & Pacific Northwest National Laboratory\\
Richard Vuduc & Georgia Institute of Technology\\
Andreas Waechter & Northwestern University\\
Ji Wang & University of California, Davis\\
Zhehui (Jeph) Wang & Los Alamos National Laboratory\\
Ziteng Wang & Northern Illinois University\\
Bei Wang Phillips & University of Utah\\
Ben Whitney & Oak Ridge National Laboratory\\
Stefan Wild & Argonne National Laboratory\\
Don Willcox & Lawrence Berkeley National Laboratory\\
David Williams-Young & Lawrence Berkeley National Laboratory\\
Vince Winstead & Minnesota State University, Mankato\\
Jonathan Wittmer & University of Texas at Austin\\
Brendt Wohlberg & Los Alamos National Laboratory\\
David Womble & Oak Ridge National Laboratory\\
Stephen Wright & University of Wisconsin, Madison\\
John Wu & Lawrence Berkeley National Laboratory\\
Penghao Xiao & Lawrence Livermore National Laboratory\\
Yu Xiao & Corning Incorporated\\
Miaolan Xie & Cornell University\\
Weijun Xie & Virginia Tech\\
Yuege Xie & University of Texas at Austin\\
Ricky Yan & Unaffiliated\\
Chao Yang & Lawrence Berkeley National Laboratory\\
Yunan Yang & New York University\\
Jia Yin & Lawrence Berkeley National Laboratory\\
Srikanth Yoginath & Oak Ridge National Laboratory\\
Stephen Young & Pacific Northwest National Laboratory\\
Roozbeh Yousefzadeh & Yale University \\
Kwangmin Yu & Brookhaven National Laboratory\\
Alfi Zakiyyah & Bina Nusantara University\\
Rohit Zambre & University of California, Irvine\\
Boya Zhang & Lawrence Livermore National Laboratory\\
Guannan Zhang & Oak Ridge National Laboratory\\
Jiaxin Zhang & Oak Ridge National Laboratory\\
John Zhang & University of California, Los Angeles\\
Pei Zhang & Oak Ridge National Laboratory\\
Xinyu Zhang & North Carolina State University\\
Yifan Zhang & University of Texas at Austin\\
Ziyun Zhang & California Institute of Technology\\
Lili Zheng & University of Wisconsin, Madison\\
Dekun Zhou & University of Wisconsin, Madison\\
Yanhui Zhu & University of West Florida\\
Yue Zhu & Pacific Northwest National Laboratory\\
Roger Zoh & Indiana University\\
Zhihui Zou & University of Texas at Austin\\
Petrus Zwart & Lawrence Berkeley National Laboratory\\
\end{supertabular}

%
%
%
%
% --- End Inserted File ---

\clearpage
% ---- Inserted File ----
\section{Acknowledgments}
\label{app:acks}

We are grateful to many contributors to the process that led to the composition of this report on an ambitious timeline. 

We are grateful to the staff at ORISE (Jody Crisp, Andrew Fowler, and Paul Hudson, in particular) for managing the logistics of a large, multipart virtual workshop. 

Special thanks are due to the presenters at
the bootcamp. 
Miles Lopes, C. (Sesh) Seshadhri, Per Gunnar Martinsson, and John Duchi coordinated to provide a cohesive set of tutorials. Julia Ling,  Salman Habib, Aurora Clark, Charles Bouman, and Peter Zwart shared valuable perspectives on challenges and opportunities in their respective scientific areas.

We are grateful to the breakout leads for the bootcamp, who were invaluable in facilitating discussion among randomly assigned breakout participants. These leads included 
Jim Ahrens,
Rick Archibald,
Eric Cyr,
Anthony DeGennaro,
Jed Duersch,
Aric Hagberg,
Joey Hart,
Rebecca Hartman-Baker,
Jeff Hittinger,
Paul Hovland,
John Jakeman,
Chandrika Kamath,
Ramki Kannan,
Hemanth Kolla,
Rich Lehoucq,
Sven Leyffer,
Sherry Li,
Osni Marques,
Kary Myers,
Jelani Nelson,
Esmond Ng,
Cindy Phillips,
Eric Phipps,
Ali Pinar,
Vishwas Rao,
Juan Restrepo,
Ray Tuminaro,
Draguna Vrabie,
David Womble,
Steve Wright,
Chao Yang, 
and
Stephen Young.

We thank the federal agency panelists
Fariba Fahroo (AFOSR),
Steven Lee (DOE),
Reza Malek-Madani (ONR), and 
Grace Peng (NIH)
for sharing insight on the status and opportunities for randomized algorithms in each of their agencies.

Breakout leads on both days of Part 2 distilled significant attendee input. We are grateful to 
Mihai Anitescu,
Charlie Bouman,
Aurora Clark,
Anthony DeGennaro,
Salman Habib,
Chandrika Kamath,
Ramki Kannan,
Miles Lopes,
Per Gunnar Martinsson,
Kary Myers,
Jelani Nelson,
Cindy Phillips,
Juan Restrepo,
C. Seshadhri,
Joel Tropp,
Draguna Vrabie,
Brendt Wohlberg,
Steve Wright,
Chao Yang, and
Peter Zwart
for serving in this role.

We thank Tiffani Conner and Gail Pieper for their technical editing of the report.

We thank Steven Lee for the charge to identify research needs in randomized algorithms for scientific computing 
to advance the mission of DOE's Office of Science.
% --- End Inserted File ---

\end{document}